\documentclass{article}

 \PassOptionsToPackage{nonatbib}{neurips_2020}
 \usepackage[preprint]{neurips_2020}

\usepackage[utf8]{inputenc} % allow utf-8 input
\usepackage[T1]{fontenc}    % use 8-bit T1 fonts
\usepackage{hyperref}       % hyperlinks
\usepackage{url}            % simple URL typesetting
\usepackage{booktabs}       % professional-quality tables
\usepackage{amsfonts}       % blackboard math symbols
\usepackage{nicefrac}       % compact symbols for 1/2, etc.
\usepackage{microtype}      % microtypography

\usepackage{graphicx}
\usepackage{subcaption, floatrow}

\usepackage{xcolor}
\usepackage{amsmath}
\usepackage{hyperref}
\usepackage{amssymb}
\usepackage{cleveref}

\usepackage{algorithmic}

\usepackage{algorithm}

\usepackage{xcolor}

\usepackage{makecell}

\newlength{\minuslength}
\renewcommand{\algorithmiccomment}[1]{{\color{blue}\texttt{\bgroup\hfill//~#1\egroup}}}

\DeclareMathOperator*{\argmin}{arg\,min}

\newcommand*{\tran}{^{\mkern-1.5mu\mathsf{T}}}

\title{Regularisation Can Mitigate Poisoning Attacks: \\ A Novel Analysis Based on \\Multiobjective Bilevel Optimisation}

\author{%
  Javier Carnerero-Cano$^1$, Luis Mu\~noz-Gonz\'alez$^1$, Phillippa Spencer$^2$, Emil C. Lupu$^1$ \\
  $^1$Department of Computing, Imperial College London\\
  $^2$Defence Science and Technology Laboratory (Dstl)\\
  \texttt{\{j.cano, l.munoz, e.c.lupu\}@imperial.ac.uk} \\
}

\begin{document}

\maketitle

\begin{abstract}
 Machine Learning (ML) algorithms are vulnerable to poisoning attacks, where a fraction of the training data is manipulated to deliberately degrade the algorithms' performance. Optimal poisoning attacks, which can be formulated as bilevel optimisation problems, help to assess the robustness of learning algorithms in worst-case scenarios. However, current attacks against algorithms with hyperparameters typically assume that these hyperparameters remain constant ignoring the effect the attack has on them. We show that this approach leads to an overly pessimistic view of the robustness of the algorithms. We propose a novel optimal attack formulation that considers the effect of the attack on the hyperparameters by modelling the attack as a \emph{multiobjective bilevel optimisation problem}. We apply this novel attack formulation to ML classifiers using $L_2$ regularisation and show that, in contrast to results previously reported, $L_2$ regularisation enhances the \emph{stability} of the learning algorithms and helps to mitigate the attacks. Our empirical evaluation on different datasets confirms the limitations of previous strategies, evidences the benefits of using $L_2$ regularisation to dampen the effect of poisoning attacks and shows how the regularisation hyperparameter increases with the fraction of poisoning points.
 
\end{abstract}

\section{Introduction}

\label{sec:intro}
In many applications, Machine Learning (ML) systems rely on data collected from untrusted data sources, such as humans, machines, sensors, or IoT devices that can be compromised and manipulated. Malicious data from these compromised sources can then be used to poison the learning algorithms themselves. In other applications, the labelling of the training datasets is done manually and \emph{crowdsourcing} techniques are used to aggregate the labelling information from a set of human annotators. In these cases, \emph{crowdturfing} attacks are possible where malicious annotators collude to deceive the crowdsourcing algorithm by manipulating some of the labels of the annotated examples \cite{yao2017automated}. All these scenarios expose ML algorithms to poisoning attacks, where adversaries manipulate a fraction of the training data to subvert the learning process, either to decrease its overall performance or to produce a particular kind of error in the system \cite{barreno2010security, huang2011adversarial, munoz2019security}. Poisoning attacks can also facilitate subsequent evasion attacks or produce \emph{backdoor} (or \emph{Trojan}) attacks \cite{gu2019badnets, liu2017trojaning}.

Several systematic poisoning attacks have already been proposed to analyse different families of ML algorithms under worst-case scenarios, including Support Vector Machines (SVMs) \cite{biggio2012poisoning}, other linear classifiers \cite{xiao2015feature, mei2015using, koh2018stronger}, and neural networks \cite{koh2017understanding, munoz2017towards}. These attack strategies are formulated as a bilevel optimisation problem, i.e. an optimisation problem that \emph{depends} on another optimisation problem. In these cases, the attacker typically aims to maximise some arbitrary malicious objective (e.g. to maximise the error for a set of target points) by manipulating a fraction of the training data. At the same time, the defender aims to optimise a different objective function to learn the model's parameters, typically by minimising some loss function evaluated on the poisoned training set. 

Some of the previous attacks target algorithms that have hyperparameters, but consider them constant regardless of the fraction of poisoning points injected in the training dataset. This can provide a misleading analysis of the robustness of the algorithms against such attacks, as the value of the hyperparameters can change depending on the type and strength of the attack. For example, Xiao et al. \cite{xiao2015feature} presented a poisoning attack against embedded feature selection methods, including $L_1$, $L_2$ and \emph{elastic-net} regularisation. Their experimental results show that the attacker can completely control the selection of the features to significantly increase the overall test error of linear classifiers. However, they assume a constant regularisation hyperparameter regardless of the attack scenario considered. As shown in this paper, this approach can provide overly pessimistic results on the ML algorithms' robustness to poisoning attacks, as the value of the regularisation hyperparameters significantly changes (if they are optimised) when malicious points are injected in the training set. 

In this paper we propose a novel and more general poisoning attack formulation to test worst-case scenarios against ML algorithms that contain hyperparameters. For this, we model the attack as a \emph{multiobjective bilevel optimisation problem}, where the outer objective includes both the learning of the poisoning points and the hyperparameters of the model, whereas the inner problem involves the learning of the model's parameters. In a worst-case scenario, where the attacker is aware of the dataset used to learn the model's hyperparameters and aims to maximise the overall error, the outer objective can be modelled as a \emph{minimax} problem. We applied our proposed attack formulation to test the robustness of ML classifiers that use $L_2$ regularisation. We used \emph{hypergradient} (i.e. the gradient in the outer problem \cite{maclaurin2015gradient, franceschi2017forward, franceschi2018bilevel}) descent/ascent to solve the proposed multiobjective bilevel optimisation problem. As the computation of the exact hypergradients can be very expensive, especially for neural networks, we used Reverse-Mode Differentiation (RMD) \cite{domke2012generic,maclaurin2015gradient,munoz2017towards,franceschi2018bilevel} to approximate the aforementioned hypergradients. Our experimental evaluation shows that, contrary to the results reported in \cite{xiao2015feature}, $L_2$ regularisation helps as a mechanism to partially mitigate the effect of poisoning attacks. This is not surprising, as $L_2$ regularisation is known to increase the stability of the learning algorithm, and thus, can hinder the ability of the attacker to poison the target algorithm. We show that the value of the regularisation hyperparameter increases with the strength of the attack. In other words, the algorithm automatically tries to compensate the negative effect of the poisoning points by increasing the strength of the regularisation term. Our proposed attack formulation allows us to appropriately evaluate the robustness of $L_2$ regularisation in worst-case scenarios.

\section{Related Work} \label{sec:related}
The first poisoning attacks reported in the literature targeted specific applications, such as spam filtering \cite{nelson2008exploiting,barreno2010security} or anomaly detection \cite{barreno2006can,kloft2012security}. A more systematic approach was introduced in \cite{biggio2012poisoning} to poison SVMs, modelling the attack as a bilevel optimisation problem. Subsequent works extended this approach to other families of ML algorithms, including linear and other convex classifiers \cite{mei2015using} or embedded feature selection methods \cite{xiao2015feature}. A more general approach was introduced in \cite{munoz2017towards} formulating different optimal attack strategies for targeting multiclass classifiers. The authors also proposed an algorithm to estimate the hypergradients in the corresponding bilevel optimisation problem through RMD, which significantly improves the scalability of optimal attacks, allowing to poison a broader range of learning algorithms, including neural networks. The authors in \cite{koh2018stronger} proposed an algorithm for solving bilevel optimisation problems with detectability constraints, allowing to craft poisoning points that can bypass outlier detectors. However, the algorithm is computationally demanding, which limits its applicability in many practical scenarios. 

Other approaches have also been proposed for crafting poisoning attacks: Koh et al. \cite{koh2017understanding} created adversarial training examples by exploiting influence functions. This approach allows to craft successful targeted attacks by injecting small perturbations to genuine data points in the training set. Shafahi et al.  \cite{shafahi2018poison} proposed a targeted attack for situations where the adversary does not control the labels for the poisoning points. A Generative Adversarial Net-based model to craft poisoning attacks at scale against deep networks was proposed in \cite{munoz2019poisoning}. This approach allows to model naturally detectability constraints for the attacker, enabling attacks with different levels of aggressiveness. 

On the defender's side, it is possible to mitigate poisoning attacks by analysing the samples that can have a negative impact on the target algorithms \cite{nelson2008exploiting}. However, this approach can be impractical in many applications, as it offers a poor scalability. Similarly, in \cite{koh2017understanding}, the use of influence functions is proposed as a mechanism to detect poisoning points. Different outlier detection schemes have proved to be effective to mitigate poisoning attacks in cases where the attacker does not model appropriate detectability constraints \cite{steinhardt2017certified,paudice2018detection}. Label sanitisation has also been proposed as a mechanism to identify and relabel suspicious training points \cite{paudice2018label, zhang2018training}. However, this strategy can fail in attacks where the poisoning points collude \cite{munoz2019poisoning}. Finally, Diakonikolas et al. \cite{diakonikolas2019sever} proposed a robust meta-algorithm, based on Singular Value Decomposition, capable of mitigating some poisoning attacks.

\section{General Optimal Poisoning Attacks} \label{sec:generalAttacks}
In data poisoning attacks the attacker can tamper with a fraction of the training data points to manipulate the behaviour of the learning algorithm \cite{barreno2006can,barreno2010security}. We assume that the attacker can manipulate all the features and the label of the injected poisoning points, provided that the resulting points are within a feasible domain of valid data points. We consider white-box attacks with perfect knowledge, i.e. the attacker knows everything about the target system, including the training data, the feature representation, the loss function, the ML model, and the defence (if applicable) used by the victim. Although unrealistic in most cases, these assumptions for the adversary allow us to analyse the robustness of the ML algorithms in worst-case scenarios for attacks with different strength.  

\subsection{Problem Formulation}
In line with most literature on poisoning attacks we consider ML classifiers. Then, in a classification task, given the input space $\mathcal{X}\in \mathbb{R}^m$ and the label space, $\mathcal{Y}$, the learner aims to estimate the mapping $f: \mathcal{X} \rightarrow \mathcal{Y}$. Given a training set $\mathcal{D}_\text{tr} = \{(\textbf{x}_{\text{tr}_i} , y_{\text{tr}_i})\}^{n_\text{tr}}_{i=1}$ with $n_\text{tr}$ IID samples drawn from the underlying
probability distribution $p(\mathcal{X}, \mathcal{Y})$, we can estimate $f$ with a model $\mathcal{M}$ trained by minimising an objective function $\mathcal{L}(\mathcal{D}_\text{tr}, \boldsymbol{\Lambda}, {\textbf w})$ w.r.t. its parameters\footnote{As in \cite{maclaurin2015gradient}, we use parameters
to denote ``parameters that are just parameters and not hyperparameters''.}, $\textbf{w}\in \mathbb{R}^d$, given a set of hyperparameters $\boldsymbol{\Lambda}\in \mathbb{R}^c$. 

We assume that the defender has access to a small validation dataset $\mathcal{D}_\text{val} = \{(\textbf{x}_{\text{val}_j} , y_{\text{val}_j})\}^{n_\text{val}}_{j=1}$ with $n_\text{val}$ trusted data points. In practical settings it is not uncommon to have access to a limited clean dataset, e.g. because the integrity of a small set of data sources can be ascertained or because the system is re-calibrated at specific points in time. This small clean dataset is held out for hyperparameter optimisation. Then, as proposed in \cite{foo2008efficient}, the model's hyperparameters can be learned by solving the following bilevel optimisation problem: 

\begin{equation}
\begin{aligned}
\min_{\boldsymbol{\Lambda} \in \Phi(\boldsymbol{\Lambda})}  \quad & \mathcal{L}(\mathcal{D}_\text{val},  \textbf{w}^\star)  \\
\text{s.t.} \quad & \textbf{w}^\star\in\argmin_{\textbf{w} \in \mathcal{W}}  & \mathcal{L}\left(\mathcal{D}_\text{tr}, \boldsymbol{\Lambda}, \textbf{w}\right),\\
\end{aligned}
\label{eqHyperparams}
\end{equation}

 where $\Phi(\boldsymbol{\Lambda})$ represent the feasible domain set for the hyperparameters $\boldsymbol{\Lambda}$. On the other side, in a poisoning attack, the adversary aims to inject a set of $n_\text{p}$ malicious data points, $\mathcal{D}_\text{p} = \{(\textbf{x}_{\text{p}_k} ,y_{\text{p}_k})\}^{n_\text{p}}_{k=1}$, in the training dataset to maximise some arbitrary objective, $\mathcal{A}$, evaluated on a set of target data points $\mathcal{D}_\text{target}$. As described in \cite{munoz2017towards} different attack scenarios can be considered depending on both the set of target data points and the attacker's objective, including indiscriminate and targeted attacks. In these settings, we propose to formulate the problem for the attacker as the multiobjective bilevel optimisation problem: 
 
\begin{equation}
\begin{aligned}
\min_{\boldsymbol{\Lambda} \in \Phi(\boldsymbol{\Lambda})}  \mathcal{L}(\mathcal{D}_\text{val}, & \  \textbf{w}^\star),  \max_{\mathcal{D}_\text{p} \in \Phi(\mathcal{D}_\text{p})}    \mathcal{A}(\mathcal{D}_\text{target},  \textbf{w}^\star) \\
& \text{s.t.} \quad \textbf{w}^\star\in\argmin_{\textbf{w} \in \mathcal{W}}  \mathcal{L}\left(\mathcal{D}_\text{tr}',  \boldsymbol{\Lambda}, \textbf{w}\right),\\
\end{aligned}
\label{eqAttacker}
\end{equation} 

where $\mathcal{D}_\text{tr}' = \mathcal{D}_\text{tr} \cup \mathcal{D}_\text{p}$ is the poisoned dataset and $\Phi(\mathcal{D}_\text{p})$ is the feasible domain for the attacker.

\begin{figure*}[!t]
	\begin{centering}
		\begin{subfigure}[b]{0.316\textwidth}
			\includegraphics[width=\textwidth]{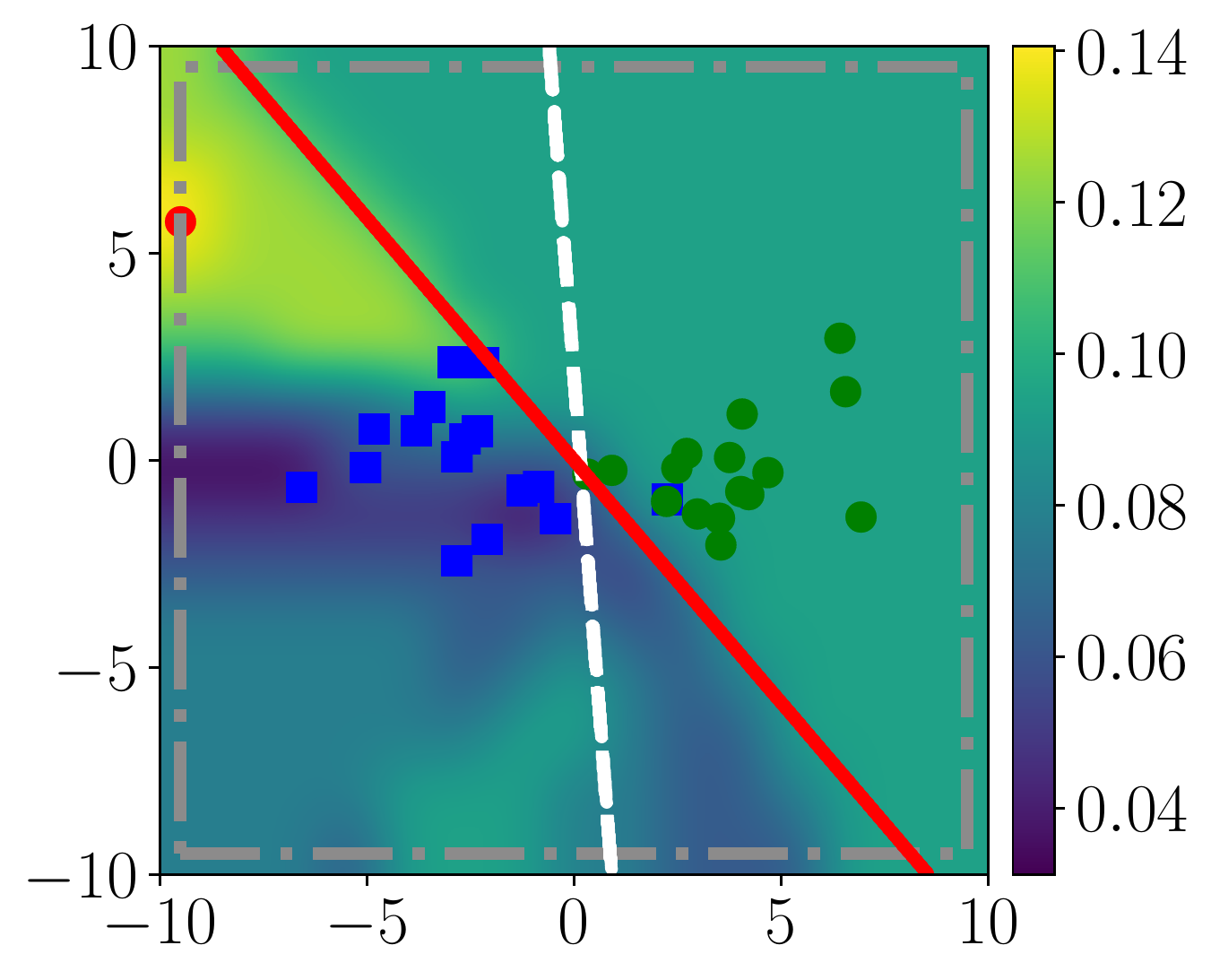}
		\end{subfigure}
		\enskip % Control spacing between left and right figure, can use \enskip, \quad, \qquad, \hfill\centering
		\begin{subfigure}[b]{0.316\textwidth}
			\includegraphics[width=\textwidth]{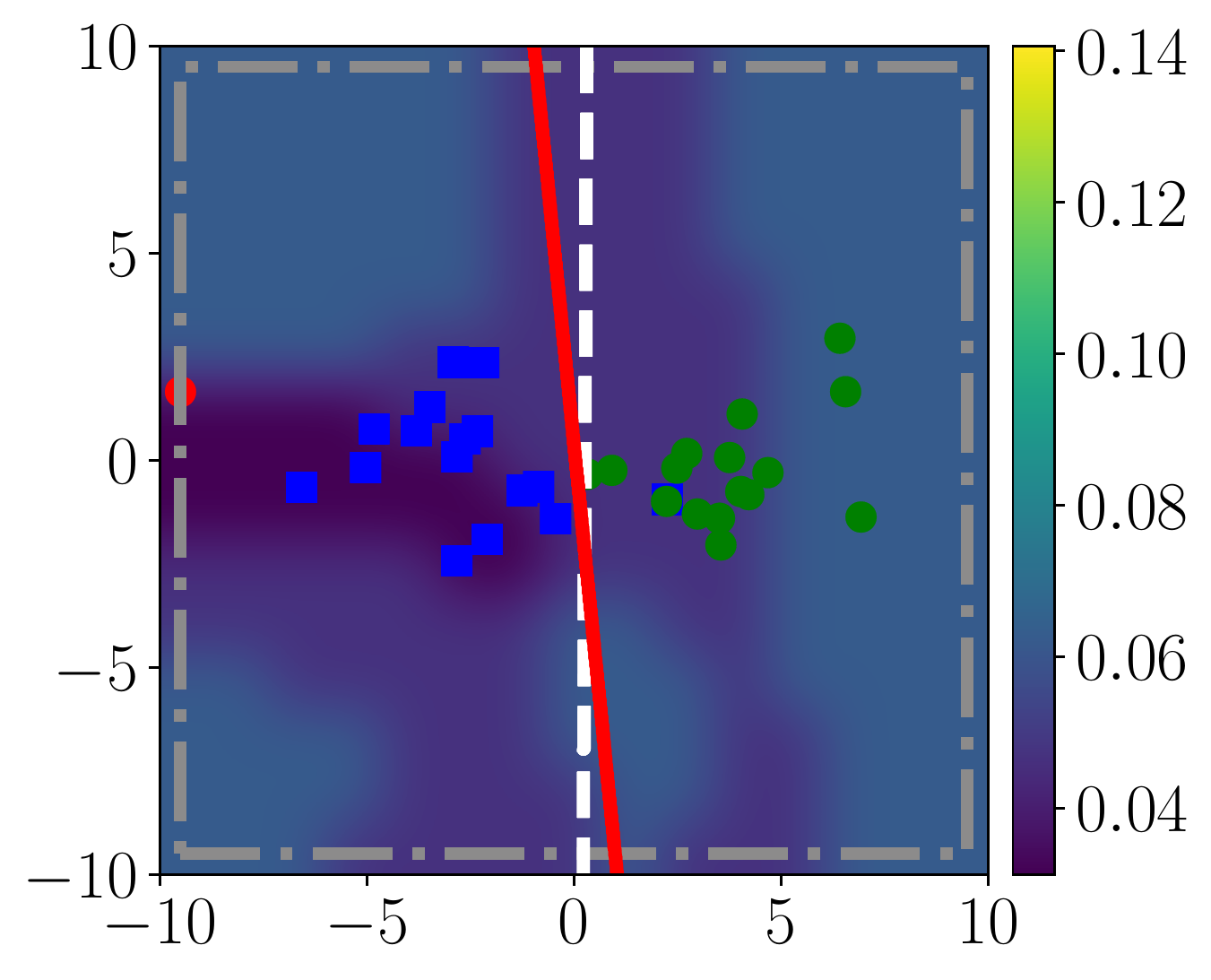}
		\end{subfigure}
		\enskip % Control spacing between left and right figure, can use \enskip, \quad, \qquad, \hfill\centering
		\begin{subfigure}[b]{0.323\textwidth}
			\includegraphics[width=\textwidth]{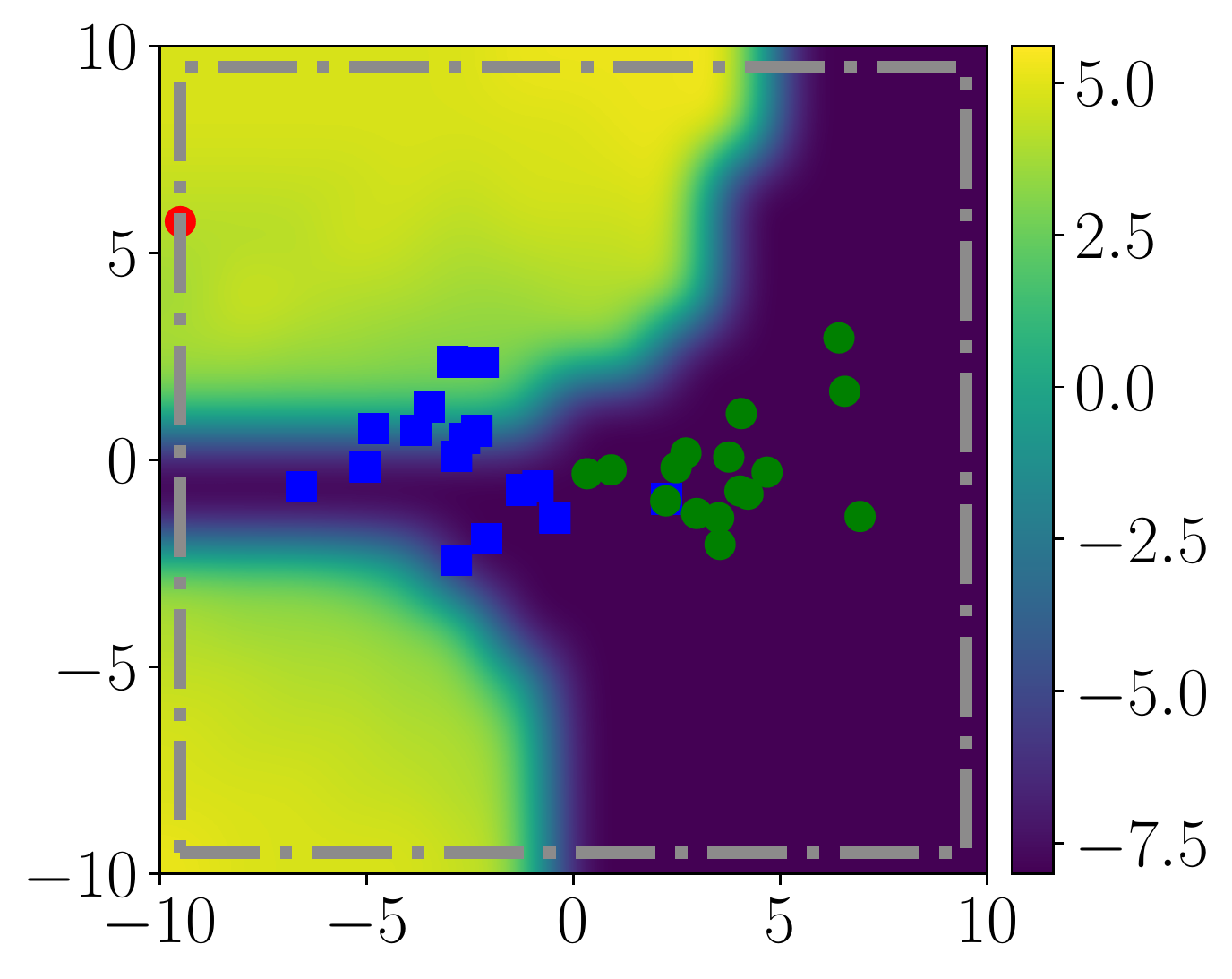}
		\end{subfigure} 
		\enskip % Control spacing between left and right figure, can use \enskip, \quad, \qquad, \hfill\centering
	\end{centering}
	
	\caption{Effect of regularisation on a synthetic example. The blue and green points represent the training data points for each class, and the red point is the poisoning point (labelled as green). The dashed-dotted grey box represent the attacker's constraints. Dashed-white lines and solid-red lines depict the decision boundaries for LR classifiers trained on clean and poisoned datasets respectively. (Left) Standard LR with no regularisation. (Centre) LR with $L_2$ regularisation. The colour-maps in the two plots represent the validation error as a function of the poisoning point. (Right) Value of $\lambda$ learned by solving (\ref{eqHyperparams}) as a function of the injected poisoning point.}
	\label{fig:synthetic}
\end{figure*}

Previous work in the research literature have neglected the effect of the hyperparameters in the problem for the attacker, e.g. the regularisation hyperparameter for the cost function for SVMs \cite{biggio2012poisoning} or for embedded feature selection methods \cite{xiao2015feature}. From the general formulation we propose in (\ref{eqAttacker}) it is clear that the poisoning points in $\mathcal{D}_\text{tr}'$ have an effect not only on the parameters of the classifier, but also on its hyperparameters. Then, testing the robustness of the learning algorithms with attacks that ignore this effect can produce misleading results, as we show, for example, in the synthetic experiment in Fig.~\ref{fig:synthetic} for the case of $L_2$ regularisation. By ignoring the effect on the hyperparameters, these attacks overestimate the adversary's capabilities to influence the learning algorithm. 

Our novel attack formulation in (\ref{eqAttacker}) allows to model a wide variety of attack scenarios, depending on the attacker's objective and the combinations between the target, validation and training data points. In the remainder of the paper we will focus on analysing \textit{worst-case scenarios for indiscriminate poisoning attacks}, i.e. those where the attacker---which has perfect knowledge---aims to increase the overall classification error in the target system. To achieve such a goal, the attacker aims to maximise the loss evaluated on the defender's validation set, i.e. $\mathcal{A}(\mathcal{D}_\text{target},  \textbf{w}^{\star}) = \mathcal{L}(\mathcal{D}_\text{val},  \  \textbf{w}^{\star})$. Then, the problem for the attacker can also be formulated as a bilevel optimisation problem where the outer objective is a minimax problem: 

\begin{equation}
\begin{aligned}
\min_{\boldsymbol{\Lambda} \in \Phi(\boldsymbol{\Lambda})} & \ \max_{\mathcal{D}_\text{p} \in \Phi(\mathcal{D}_\text{p})} \mathcal{L}(\mathcal{D}_\text{val}, \textbf{w}^\star) \\
& \text{s.t.} \quad \textbf{w}^\star\in\argmin_{\textbf{w} \in \mathcal{W}}  \mathcal{L}\left(\mathcal{D}_\text{tr}', \boldsymbol{\Lambda}, \textbf{w}\right).\\
\end{aligned}
\label{eqAttacker2}
\end{equation} 

None of previous approaches described in the research literature have modelled attacks as multiobjective bilevel optimisation problems as in (\ref{eqAttacker}) and (\ref{eqAttacker2}). Beyond the scope of our paper, our approach can also be useful to test the robustness of defensive algorithms to data poisoning in worst-case scenarios when those algorithms contain hyperparameters. 

\subsection{Solving General Optimal Poisoning Attacks}
\label{subsec:genpois}
Solving the multiobjective bilevel optimisation problems in (\ref{eqAttacker}) and (\ref{eqAttacker2}) is strongly NP-Hard \cite{bard2013practical} and, even if the inner problem is convex, the bilevel problem is, in general, non-convex. However, it is possible to use gradient-based approaches to get (possibly) suboptimal solutions, i.e. finding local optima for the problem in (\ref{eqAttacker}) and saddle points for the minimax problem in (\ref{eqAttacker2}). For the sake of clarity, in the remainder we will focus on the solution for the problem (\ref{eqAttacker2}), as we shall use it in our experiments to show the robustness of $L_2$ regularisation to indiscriminate poisoning attacks. The solution of (\ref{eqAttacker}) follows a similar procedure.

To compute the hypergradients for the outer objective, we assume that the loss function, $\mathcal{L}$, is convex and its  first and second derivatives are Lipschitz-continuous functions. Similar to \cite{biggio2012poisoning,mei2015using,xiao2015feature,munoz2017towards} we assume that the label of the poisoning points is set a priori, so the attacker just needs to learn the features for the poisoning points, ${\bf X}_\text{p}$. For the sake of clarity, in the following description we shall use $\mathcal{A}$ to denote the loss function evaluated on $\mathcal{D}_\text{val}$ in the outer objective, i.e. $\mathcal{L}(\mathcal{D}_\text{val}, {\bf w}^{\star})$, and $\mathcal{L}$ to refer to the loss function evaluated on $\mathcal{D}_\text{tr}'$ in the inner objective, i.e. $\mathcal{L}(\mathcal{D}_\text{tr}', \boldsymbol{\Lambda}, {\bf w}^{\star})$; both are evaluated on the optimal solution $\textbf{w}^{\star}$ of the inner problem. We can compute the hypergradients in the outer problem by leveraging the conditions for stationarity in the inner problem, i.e. $\nabla_{\textbf{w}}\mathcal{L} = {\bf 0}$, and applying the implicit function theorem \cite{mei2015using}. Then, the hypergradients can be computed as:

\begin{equation}
\nabla_{\textbf{X}_\text{p}} \mathcal{A} = -\left( \nabla_{\textbf{X}_\text{p}}\nabla_{\bf{w}} \mathcal{L} \right)\tran  \left( \nabla^2_{\bf{w}} \mathcal{L} \right)^{-1} \nabla_{\bf{w}} \mathcal{A}, \hspace{1cm}
\nabla_{{\boldsymbol \Lambda}} \mathcal{A} = -\left( \nabla_{{\boldsymbol \Lambda}}\nabla_{\bf{w}} \mathcal{L} \right)\tran  \left( \nabla^2_{\bf{w}} \mathcal{L} \right)^{-1} \nabla_{\bf{w}} \mathcal{A},
\label{eqHyperGrads}
\end{equation} 

where we assume that the Hessian $\nabla^2_{\bf{w}} \mathcal{L}$ is not singular. Brute-force computation of (\ref{eqHyperGrads}) requires inverting the Hessian, which scales in time as $\mathcal{O}(d^3)$ and in space as $\mathcal{O}(d^2)$---where $d$ is the dimensionality of the parameters. However, as in \cite{foo2008efficient}, we can rearrange the terms in the second part of (\ref{eqHyperGrads}) and solve first the linear system: $\left( \nabla^2_{{\bf w}} \mathcal{L} \right) {\bf v} = \nabla_{{\bf w}} \mathcal{A}$, and compute~$\nabla_{{\bf X}_\text{p}} \mathcal{A} = - \left( \nabla_{{\bf X}_\text{p}} \nabla_{{\bf w}} \mathcal{L} \right)\tran {\bf v}$.\footnote{ The resulting expression for the case of ${\boldsymbol \Lambda}$ is analogous.} The linear system can be efficiently solved by using conjugate gradient descent. Moreover, Hessian-vector products can be computed exactly and efficiently with the technique proposed in \cite{pearlmutter1994fast}, avoiding the computation and storage of the Hessian (for further details, see Appendix \ref{sec:hvp}).

However, the previous approach requires training the whole learning algorithm to compute the hypergradient. This can be intractable for some learning algorithms such as deep networks, where the number of parameters is huge. To sidestep this problem, different techniques have been proposed to estimate the value of the hypergradients \cite{domke2012generic,maclaurin2015gradient,pedregosa2016hyperparameter,franceschi2017forward,munoz2017towards,franceschi2018bilevel}. These techniques do not require to re-train the learning algorithm every time the hypergradient is computed. Instead, they estimate the hypergradient by truncating the learning in the inner problem to a reduced number of epochs. 

Depending on the order in which the operations are computed, we differentiate two approaches to estimate the hypergradients: Reverse-Mode (RMD) and Forward-Mode Differentiation (FMD) \cite{griewank2008evaluating,franceschi2017forward}. In the first case, RMD requires to first train the learning algorithm for $T$ epochs. Then, the hypergradients estimate is computed by reversing the steps followed by the learning algorithm. In some cases, RMD requires to store all the information collected in the trajectory of the parameters. This can be prohibitive for deep networks where the number of parameters is huge. However, other RMD methods proposed do not require to store this information \cite{maclaurin2015gradient,munoz2017towards}. In contrast, FMD computes the hypergradients estimate as the algorithm is trained. However, the scalability of FMD depends heavily on the number of hyperparameters compared to RMD. So, for problems where the number of hyperparameters is large, as is the case for the poisoning attacks we introduced in (\ref{eqAttacker}) and (\ref{eqAttacker2}), RMD is computationally more efficient. Hence, we used RMD in our experiments. Moreover, compared to grid search, these gradient-based techniques allow reducing the computational cost, as the algorithm does not need to be trained completely and evaluated for each combination of hyperparameters.

Finally, once we have computed the hypergradients, at each hyperiteration we can use projected hypergradient descent/ascent to update the value of the poisoning points and the model's hyperparameters: 

\begin{equation}
{\bf X}_\text{p}  \leftarrow \Pi_{\Phi(\mathcal{D}_\text{p})} \left( {\bf X}_\text{p} + \alpha \ \nabla_{{\bf X}_\text{p}} \mathcal{A} \right), \hspace{1.5cm}
{\boldsymbol \Lambda}  \leftarrow \Pi_{\Phi({\boldsymbol \Lambda})} \left( {\boldsymbol \Lambda} - \beta \ \nabla_{{\boldsymbol \Lambda}} \mathcal{A} \right),
\label{eqUpdates}
\end{equation} 

where $\alpha$ and $\beta$ are respectively the learning rates and $\Pi_{\Phi(\mathcal{D}_\text{p})}$ and $\Pi_{\Phi({\boldsymbol \Lambda})}$ are the projection operators for the features of the poisoning points, ${\bf X}_\text{p}$, and the hyperparameters, ${\boldsymbol \Lambda}$, so that their updated values are within the corresponding feasible domains, $\Phi(\mathcal{D}_\text{p})$ and $\Phi({\boldsymbol \Lambda})$. In our case we used standard gradient ascent/descent to solve (\ref{eqAttacker2}). The analysis of other alternatives to solve minimax games, such as \emph{optimistic gradient descent/ascent} \cite{daskalakis2018limit} is left for future work.

\section{\texorpdfstring{$L_2$}~~Regularisation to Mitigate Poisoning Attacks} \label{sec:L2}
Poisoning attacks are intrinsically related to the stability of ML algorithms. Attackers aim to produce large changes in the target algorithm by influencing a reduced set of training points. Xu et al. \cite{xu2011sparse} introduced the following definition of stability: \emph{``an ML algorithm is stable if its output is nearly identical on two datasets, differing on only one sample."} This concept of stability has also been studied in the field of robust statistics, in which robustness formally denotes this notion of stability \cite{rubinstein2009antidote}. It is not our intention here to provide a formal analysis on the stability of the ML algorithms, but to show that stability is an important property for the design of ML algorithms robust to data poisoning. For instance, in \cite{munoz2019poisoning} it was shown empirically that for the same fraction of attack points, the effect of a poisoning attack reduces when the number of training points increases, as the stability of the algorithm increases. Thus, data augmentation can help to limit the attacker's ability to degrade the overall system's performance, leading to significant improvements in generalisation \cite{simard2003best}. However data augmentation may not be as effective to defend against more targeted attacks or backdoors.

$L_2$ (or Tikhonov) regularisation is a well-known mechanism to increase the stability of ML algorithms \cite{xu2011sparse}. In $L_2$ regularisation we add a penalty term to the original loss function which shrinks the norm of the model's parameters, so that $\mathcal{L}( \mathcal{D}_\text{tr}, \boldsymbol{\Lambda}, \textbf{w})=\mathcal{L}(\mathcal{D}_\text{tr}, \textbf{w})+\frac{e^{\lambda}}{2}\left|\left|\textbf{w}\right|\right|_2^2,$ where $\lambda$ is the hyperparameter that controls the strength of the regularisation term. The exponential form is used to ensure a positive contribution of the regularisation term for the loss function and to help learning $\lambda$, for example by using (\ref{eqHyperparams}), as this hyperparameter is usually searched over a log-spaced grid \cite{pedregosa2016hyperparameter}. Different $L_2$ regularisation schemes can be considered: e.g., in neural networks, we could have one regularisation term for the parameters at each layer or even a different term for each parameter \cite{foo2008efficient}.

Xiao et al. \cite{xiao2015feature} analysed the robustness of embedded feature selection, including $L_2$ regularisation, for linear classifiers against optimal poisoning attacks. Although their experimental results showed that $L_2$ was slightly more robust compared to $L_1$ regularisation and \emph{elastic-net}, all the classifiers tested where significantly vulnerable to indiscriminate optimal poisoning attacks. However, these results relied on the assumption that the regularisation hyperparameter was constant regardless of the fraction of poisoning data. This approach can produce misleading results, as the poisoning points can have a significant effect on the value of the hyperparameter learned, for example by using (\ref{eqHyperparams}). 

In Fig.~\ref{fig:synthetic} we show a synthetic example with a binary classifier to illustrate the limitations of the approach considered in \cite{xiao2015feature}. Here, $32$ points for the two classes ($16$ per class) were drawn from two different bivariate Gaussian distributions and we trained a Logistic Regression (LR) classifier. Fig.~\ref{fig:synthetic}(left) shows the effect of injecting a single poisoning point (red point, labelled as green) that aims to maximise the error (measured on a separate validation set with $64$ data points) against an LR classifier with no regularisation.\footnote{ The details of the experiment can be found in Appendix \ref{sec:setsynthetic}.} The dashed-white line represents the decision boundary learned when training on the clean dataset, whereas the red line depicts the decision boundary of the LR trained on the poisoned dataset. We can observe that a single poisoning point can significantly alter the decision boundary.  Fig.~\ref{fig:synthetic}(centre), shows a similar scenario, but training an LR classifier with $L_2$ regularisation, setting $\lambda=\log(20)\approx3$. Here, we can observe that the effect of the poisoning point on the classifier is much reduced and the decision boundary shifts only slightly. In the background of these two figures we represent the error evaluated on the validation set for the LR trained on a poisoned dataset as a function of the location of the poisoning point. We can observe that, when there is no regularisation (left) the error can significantly increase when we inject the poisoning point in certain regions. On the contrary, when regularisation is applied (centre), the colour-map is more uniform, i.e. the algorithm is quite stable regardless of the position of the poisoning point and the increase in the validation error after the attack is very small. 

Fig.~\ref{fig:synthetic}(right) shows how the value of the regularisation hyperparameter, $\lambda$, changes as a function of the poisoning point. The colour-map in the background represents the value of $\lambda$ learned by solving (\ref{eqHyperparams}), i.e. the $\lambda$ that minimises the error on the validation set. We can observe that $\lambda$ can change significantly depending on the position of the poisoning point. Thus, $\lambda$ is much bigger for the regions where the poisoning point can influence more the classifier (Fig.~\ref{fig:synthetic}(left)). Then, when the poisoning attack can have a very negative impact on the classifier's performance, the importance of the regularisation term, controlled by $\lambda$, increases. It is clear that selecting the value of $\lambda$ appropriately can have a significant impact on the classifier's robustness. Furthermore, when testing the robustness of $L_2$-regularised classifiers we must consider the interplay between the attack strength and the value of  $\lambda$.

\section{Experiments}
\label{sec:experiment}
In this section, we evaluate the effectiveness of the attack strategy in (\ref{eqAttacker2}) against LR and feed-forward Deep Neural Networks (DNNs). For LR, we use three different binary classification problems: MNIST (`0' vs `8') \cite{lecun1998gradient}, Fashion-MNIST (FMNIST) (\emph{T-shirt} vs \emph{pullover}) \cite{xiao2017fashion}, and ImageNet (\emph{dog} vs \emph{fish}) \cite{russakovsky2015imagenet} preprocessed as in \cite{koh2017understanding}, where the first two datasets have $784$ features, and the latter has $2,048$ features. For the DNN, we focus on ImageNet. All datasets are balanced, and for all of them we use $512$ training and $171$ validation points, drawn at random from the original pool of training points. For testing, we use $1,954$ points for MNIST, $2,000$ for FMNIST, and $600$ for ImageNet.\footnote{The details of the experimental setup  and hardware used for the experiments can be found in Appendix \ref{sec:expset}.} All the results in our experiments are the average of $10$ repetitions with different random data splits for training and validation, whereas the test set is fixed. For all the attacks we measure the average test error for different attack strengths, where the number of poisoning points ranges from $0$ to $85$.

\begin{figure*}[!t]
	\begin{centering}
		\begin{subfigure}[b]{0.325\textwidth}
			\includegraphics[width=\textwidth]{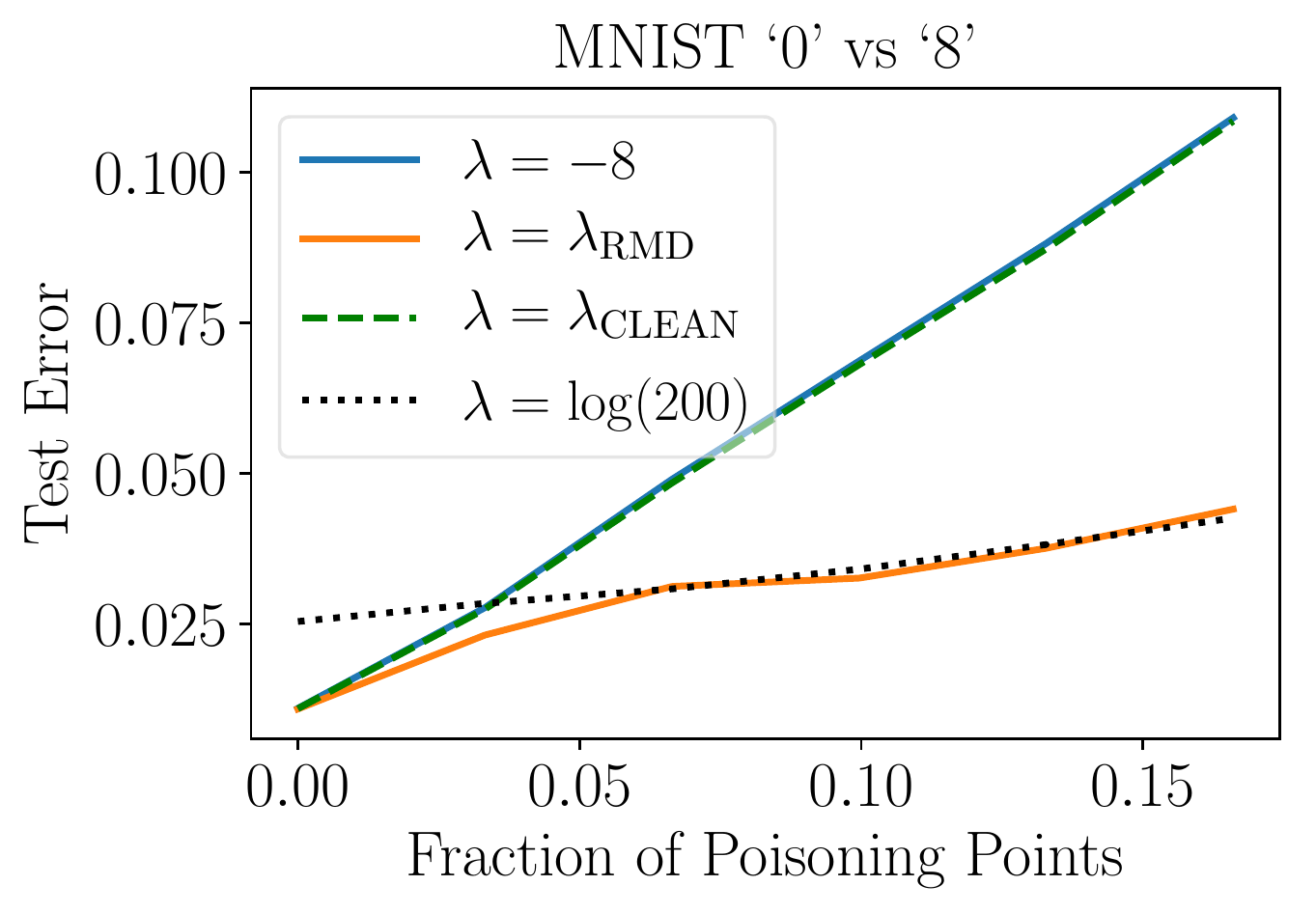}
			\caption{}
		\end{subfigure}
		\enskip % Control spacing between left and right figure, can use \enskip, \quad, \qquad, \hfill
		\begin{subfigure}[b]{0.307\textwidth}
			\includegraphics[width=\textwidth]{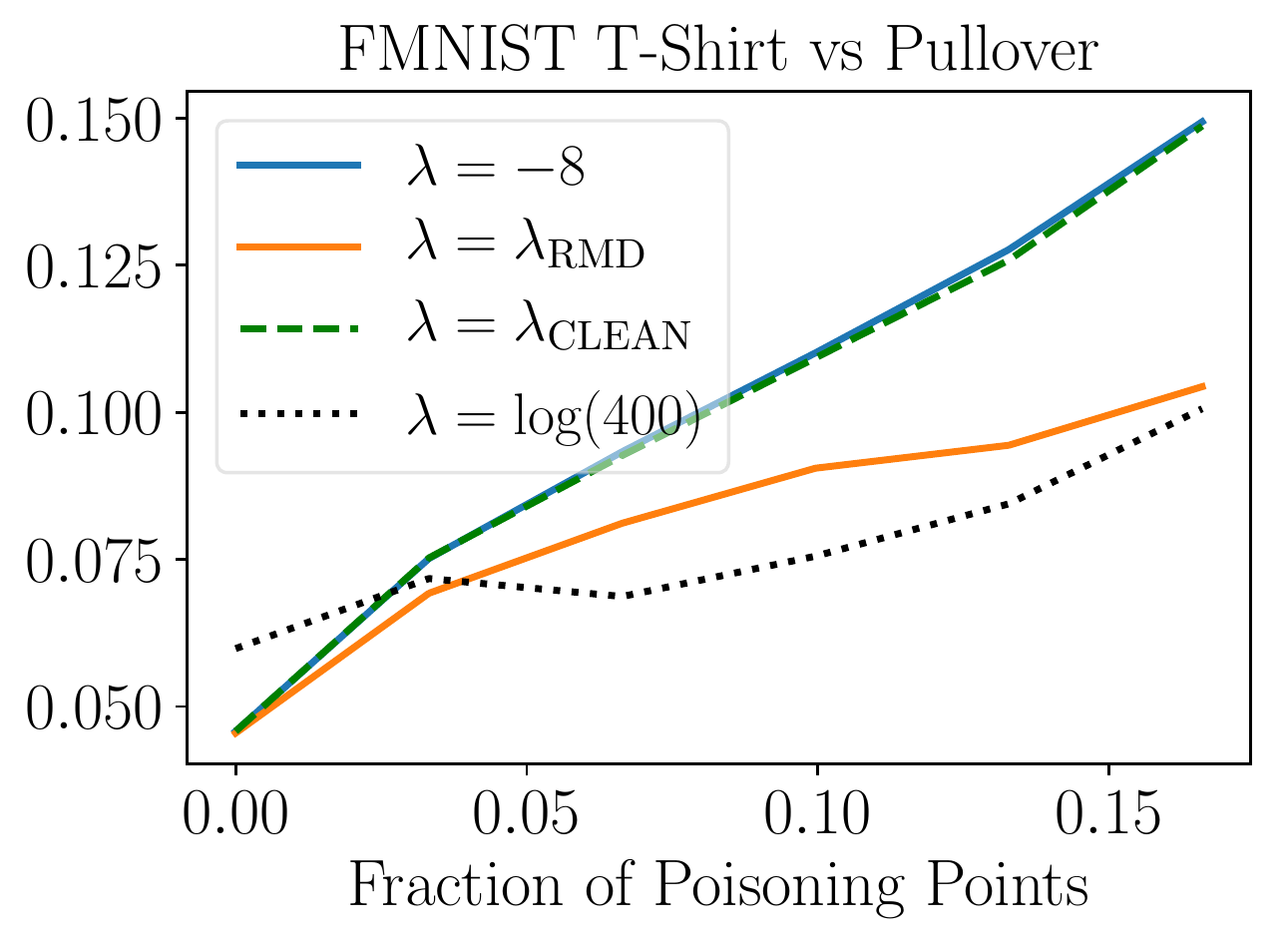}
			\caption{}
		\end{subfigure}
		\enskip % Control spacing between left and right figure, can use \enskip, \quad, \qquad, \hfill
		\begin{subfigure}[b]{0.326\textwidth}
			\includegraphics[width=\textwidth]{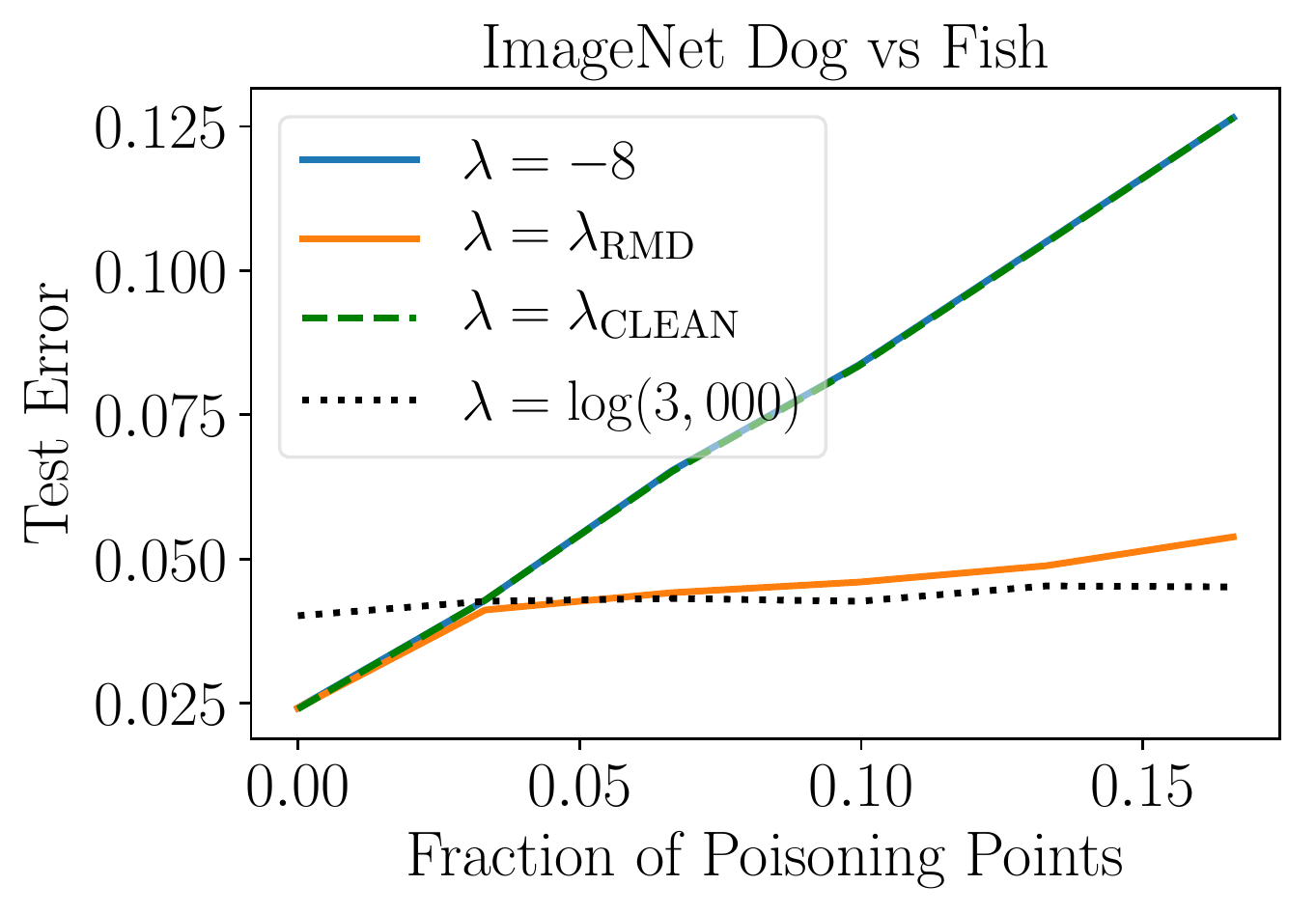}
			\caption{}
		\end{subfigure}
		\enskip % Control spacing between left and right figure, can use \enskip, \quad, \qquad, \hfill
		\begin{subfigure}[b]{0.315\textwidth}
			\includegraphics[width=\textwidth]{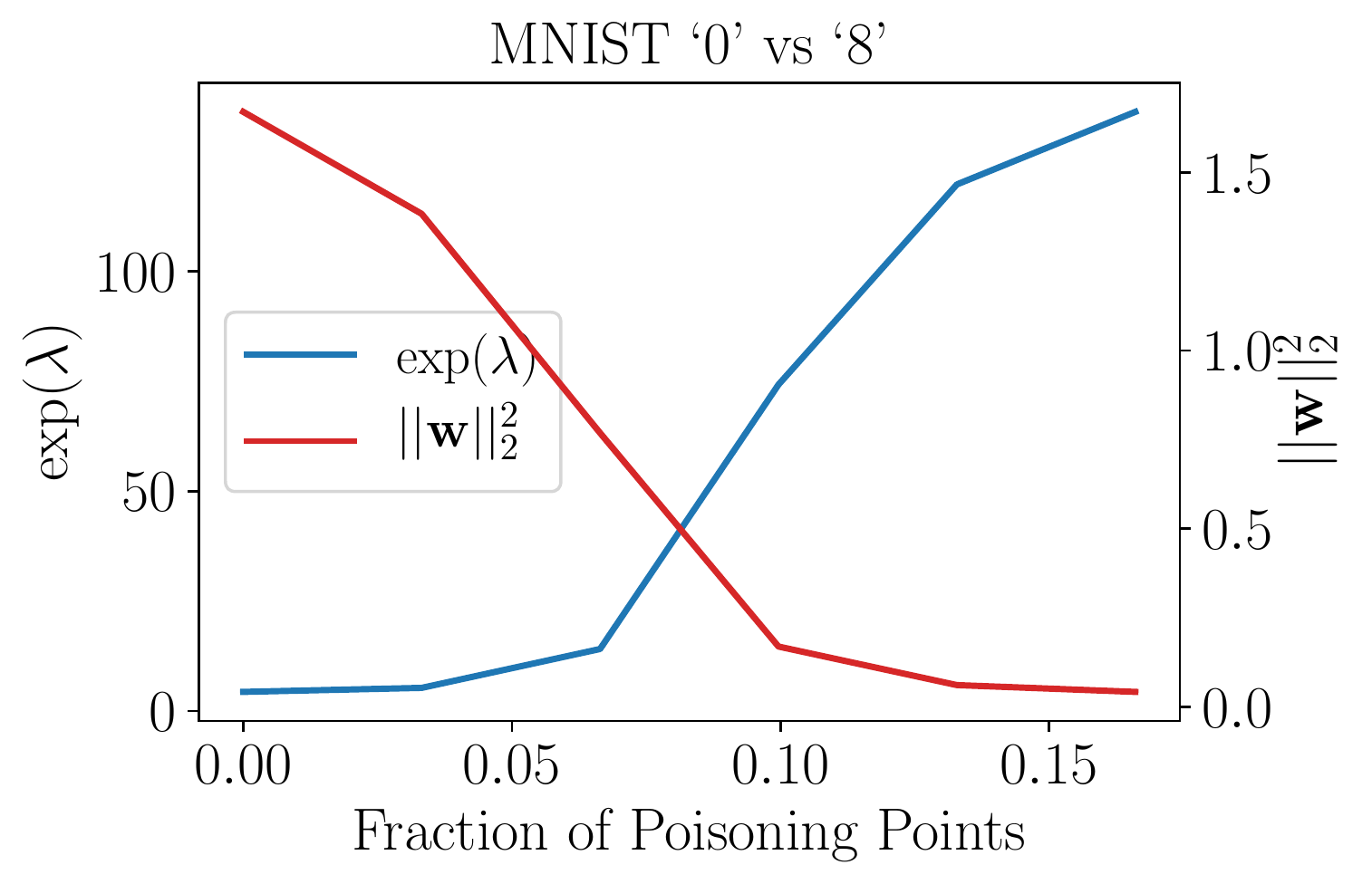}
			\caption{}
		\end{subfigure}
		\enskip % Control spacing between left and right figure, can use \enskip, \quad, \qquad, \hfill
		\begin{subfigure}[b]{0.316\textwidth}
			\includegraphics[width=\textwidth]{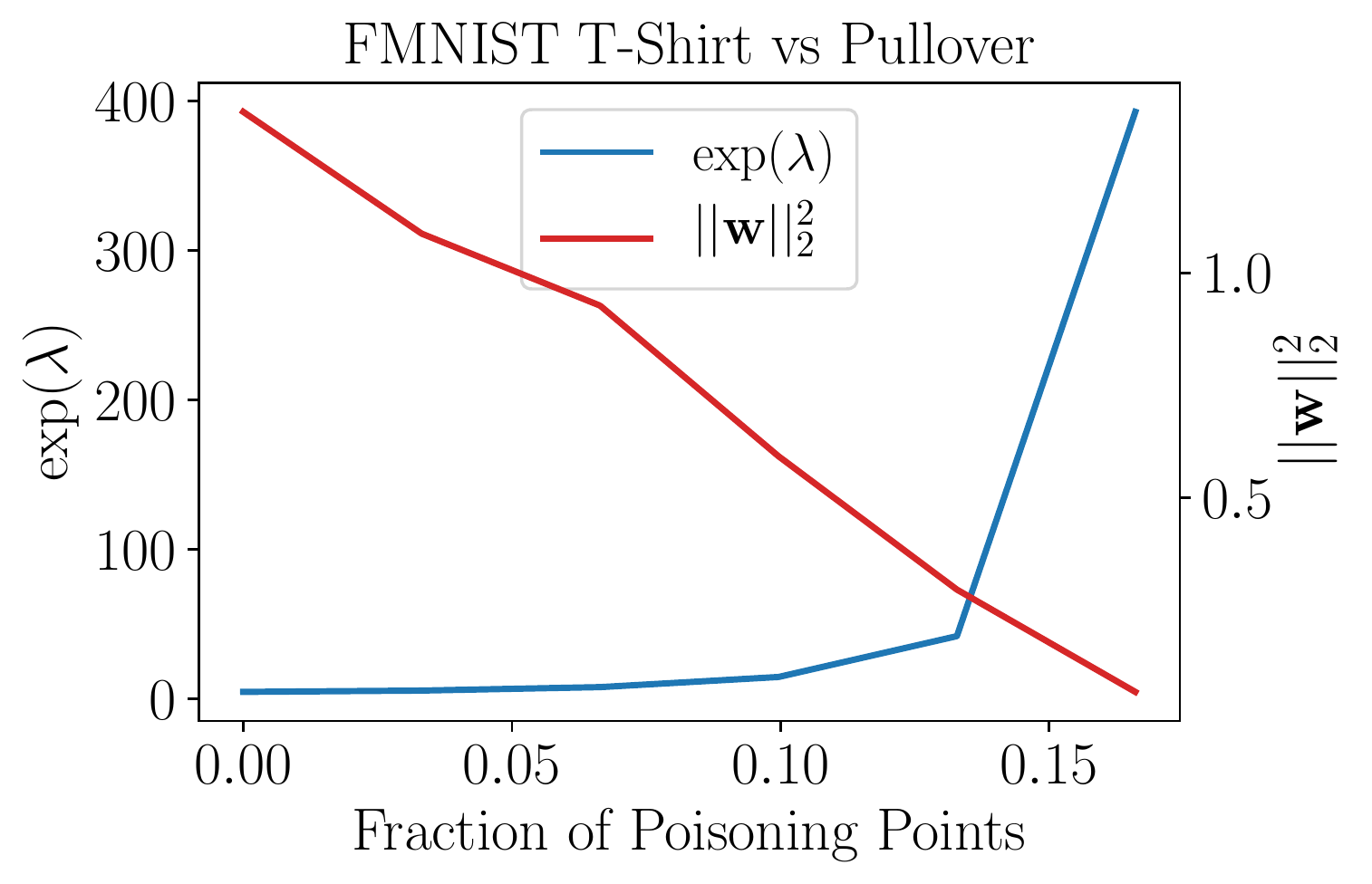}
			\caption{}
		\end{subfigure}
		\enskip % Control spacing between left and right figure, can use \enskip, \quad, \qquad, \hfill
		\begin{subfigure}[b]{0.322\textwidth}
			\includegraphics[width=\textwidth]{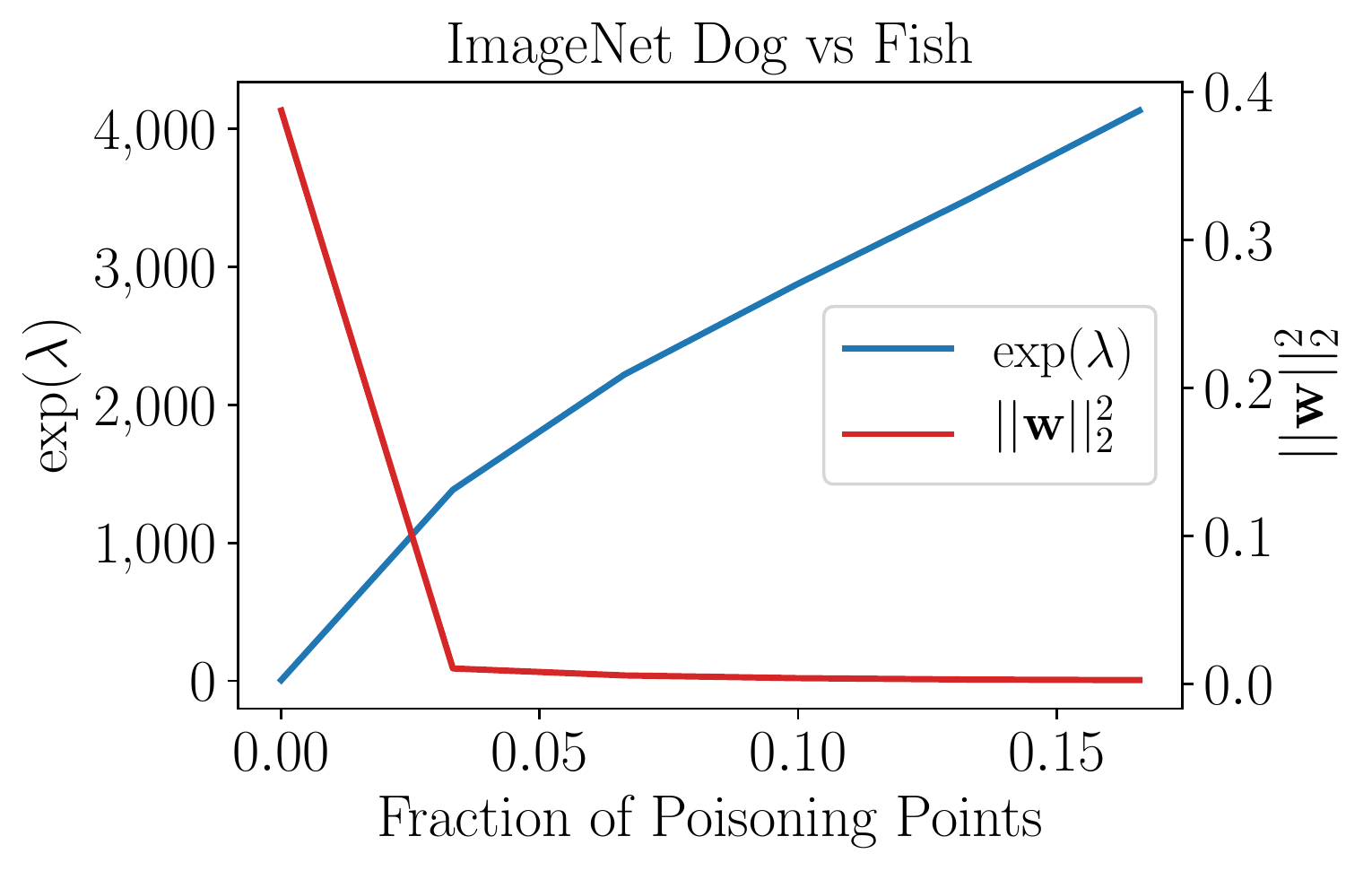}
			\caption{}
		\end{subfigure}

	\end{centering}

	\caption{Results for the optimal attack against LR: The first row represents the average test error on (a) MNIST, (b) FMNIST, and (c) ImageNet. The second row contains the plots for the average $\lambda$ and $||\textbf{w}||_2^2$ for (d) MNIST, (e) FMNIST, and (f) ImageNet.}
	\label{fig:lropt}
\end{figure*}

\subsection{Logistic Regression}
For LR we test the general poisoning attack strategy in (\ref{eqAttacker2}) using the following settings for the computation of the hypergradients with RMD. For MNIST we set $T$, the number of epochs for the inner problem, to 150. For FMNIST and ImageNet we use $T=200$.
For comparison purposes, we craft optimal poisoning attacks, setting the value of $\lambda$ to different constant values: a very small regularisation term ($\lambda = -8$), a very large one ($\lambda = \log (200)$ for MNIST, $\lambda = \log (400)$ for FMNIST, and $\lambda = \log (3,000)$ for ImageNet), and the value of $\lambda$ optimised with $5$-fold cross-validation, training the model on the clean dataset ($\lambda_\text{CLEAN}$). With the small and large constant values for $\lambda$ we aim to show the trade-off between accuracy and robustness to different attack strengths. The case of $\lambda_\text{CLEAN}$ is similar to the settings used in \cite{xiao2015feature}.

The results are shown in Fig.~\ref{fig:lropt}(a)-(c). We can observe that for the small $\lambda$ and $\lambda_\text{CLEAN}$ the attacks are very effective and the test error  increases significantly when compared to the algorithm's performance on the clean dataset. In contrast, for the biggest $\lambda$ the test error increases only slightly with the increasing fraction of poisoning points, showing a stable performance regardless of the attack strength. However, in the absence of an attack, the algorithm clearly \emph{underfits} and the error is significantly higher compared to the other models. When the value of $\lambda$ is learned ($\lambda_\text{RMD}$), i.e. solving the problem in (\ref{eqAttacker2}), the increase on the test error is quite moderate and, when the ratio of poisoning points is large, presents a similar performance as in the case where the value of $\lambda$ is large. In this sense, the error does not increase further by injecting more poisoning points. In the absence of attack, we can observe that the performance for $\lambda_\text{RMD}$ is as good as in the case of $\lambda_\text{CLEAN}$. 

The results in Fig.~\ref{fig:lropt}(a)-(c) show that the attack and the methodology presented in \cite{xiao2015feature} provide an overly pessimistic view on the robustness of $L_2$ regularisation to data poisoning attacks and that, by appropriately selecting the value of $\lambda$, we can effectively reduce the impact of such attacks. Furthermore, we can also observe that there exists a trade-off between accuracy and robustness: over-regularising (i.e. setting a very large value for $\lambda$) makes the algorithm more robust to the attack, but the performance is degraded when there is no attack. 

In Fig.~\ref{fig:lropt}(d)-(f) we show the value of $\lambda$ learned and the norm of the model's parameters, $||{\bf w}||^2_2$, as a function of the fraction of poisoning points injected for the solution of problem (\ref{eqAttacker2}) with RMD. We can observe that in all cases, the regularisation hyperparameter increases as we increase the fraction of poisoning points, which means that the regularisation term tries to compensate the effect of the malicious samples on the model's parameters. Thus, as expected, $L_2$ provides a natural mechanism to stabilise the model in the presence of attacks. For ImageNet, the order of magnitude of $\lambda$ is higher, as there are more features in the model. In Fig.~\ref{fig:lropt}(d)-(f) we can also observe that, as expected from the properties of $L_2$ regularisation, when $\lambda$ increases, the norm of the parameters decreases. 

\subsection{Deep Neural Networks}
\label{subsec:expdnn}
For DNNs, we consider a vector of regularisation hyperparameters $\lambda$, with one hyperparameter for each layer. Intuitively, the amount of scaling needed by each layer's parameters to compensate for a change in the output is not the same, given the nonlinear activation functions. This can also give us an intuition about which layer is more vulnerable to the poisoning attack. We also propose an additional modification to the RMD algorithm: we apply different initial random parameters $\textbf{w}^{(0)}$ for every update of the poisoning points. This can be interpreted as assembling different randomly initialised DNNs to improve the generalisation of the poisoning points across different parameter initialisations. In this case we set $T=600$.
This scenario is much more challenging for the multiobjective bilevel problem we aim to solve, as the model has two hidden layers ($2,048\times128\times32\times1$) with Leaky ReLU activation functions, which sums up to a total of $266,433$ parameters. 

\begin{figure*}[!tp]
	\begin{centering}
		\begin{subfigure}[b]{0.325\textwidth}
			\includegraphics[width=\textwidth]{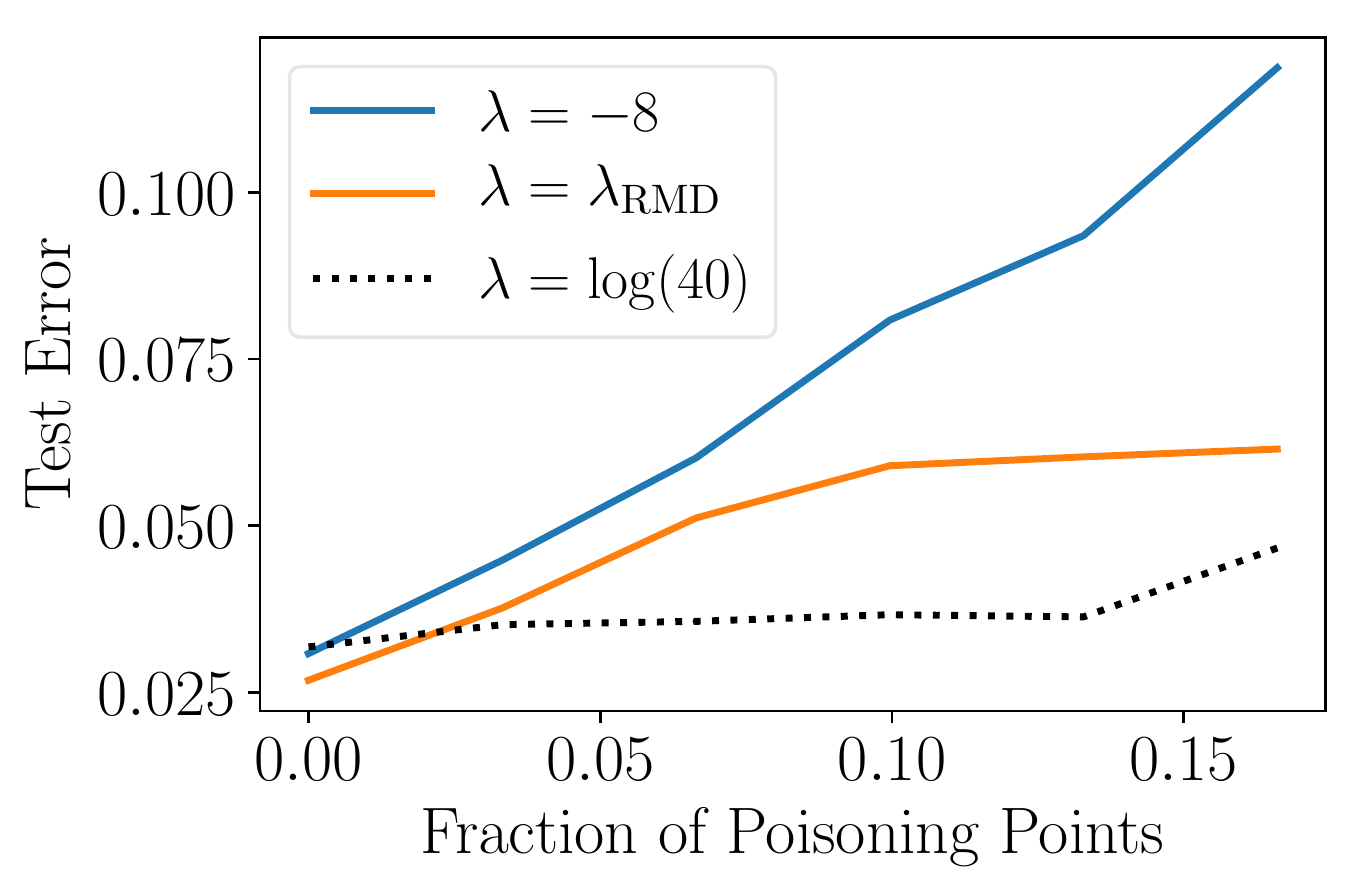}
			\caption{}
		\end{subfigure}
		\enskip % Control spacing between left and right figure, can use \enskip, \quad, \qquad, \hfill
		\begin{subfigure}[b]{0.309\textwidth}
			\includegraphics[width=\textwidth]{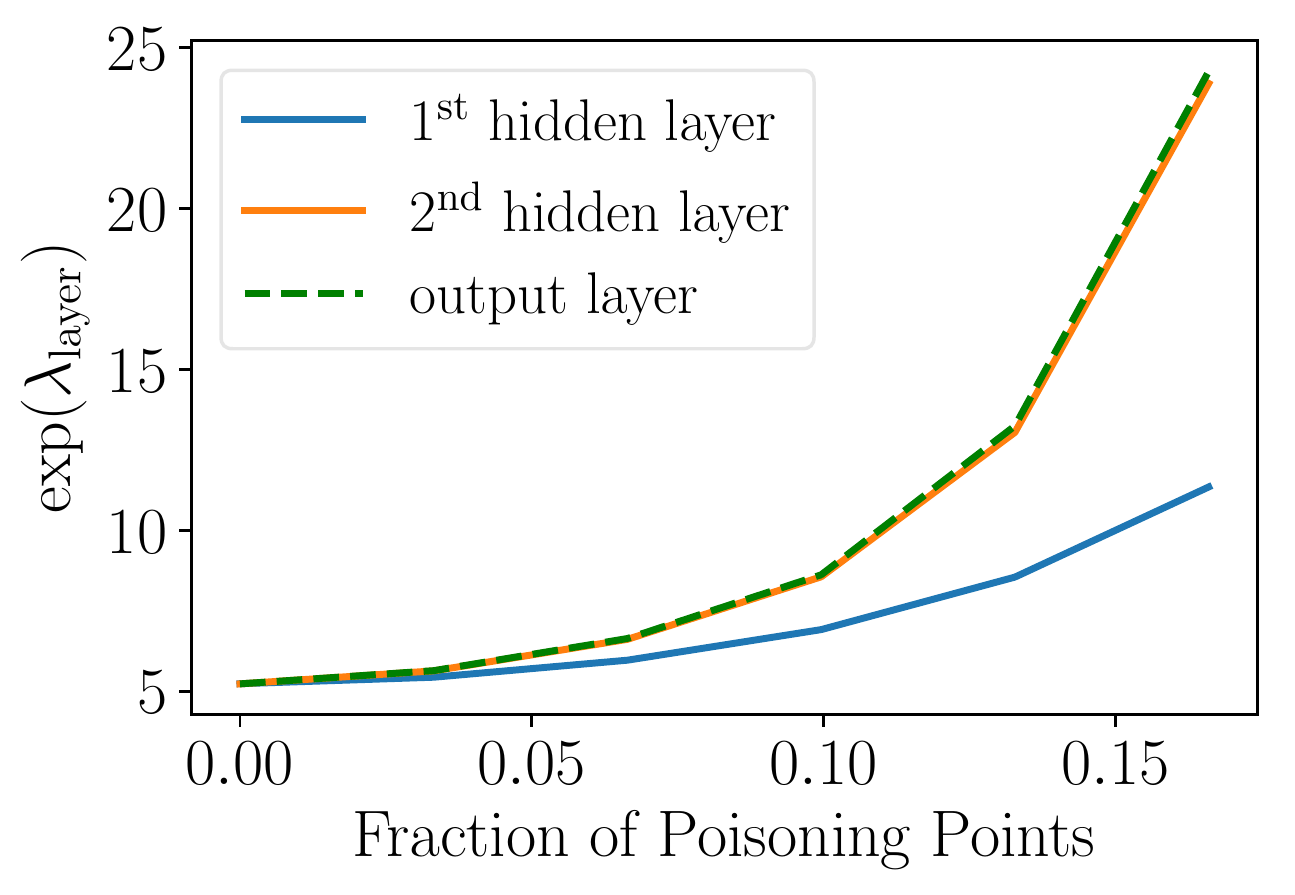}
			\caption{}
		\end{subfigure}
		\enskip % Control spacing between left and right figure, can use \enskip, \quad, \qquad, \hfill
		\begin{subfigure}[b]{0.324\textwidth}
			\includegraphics[width=\textwidth]{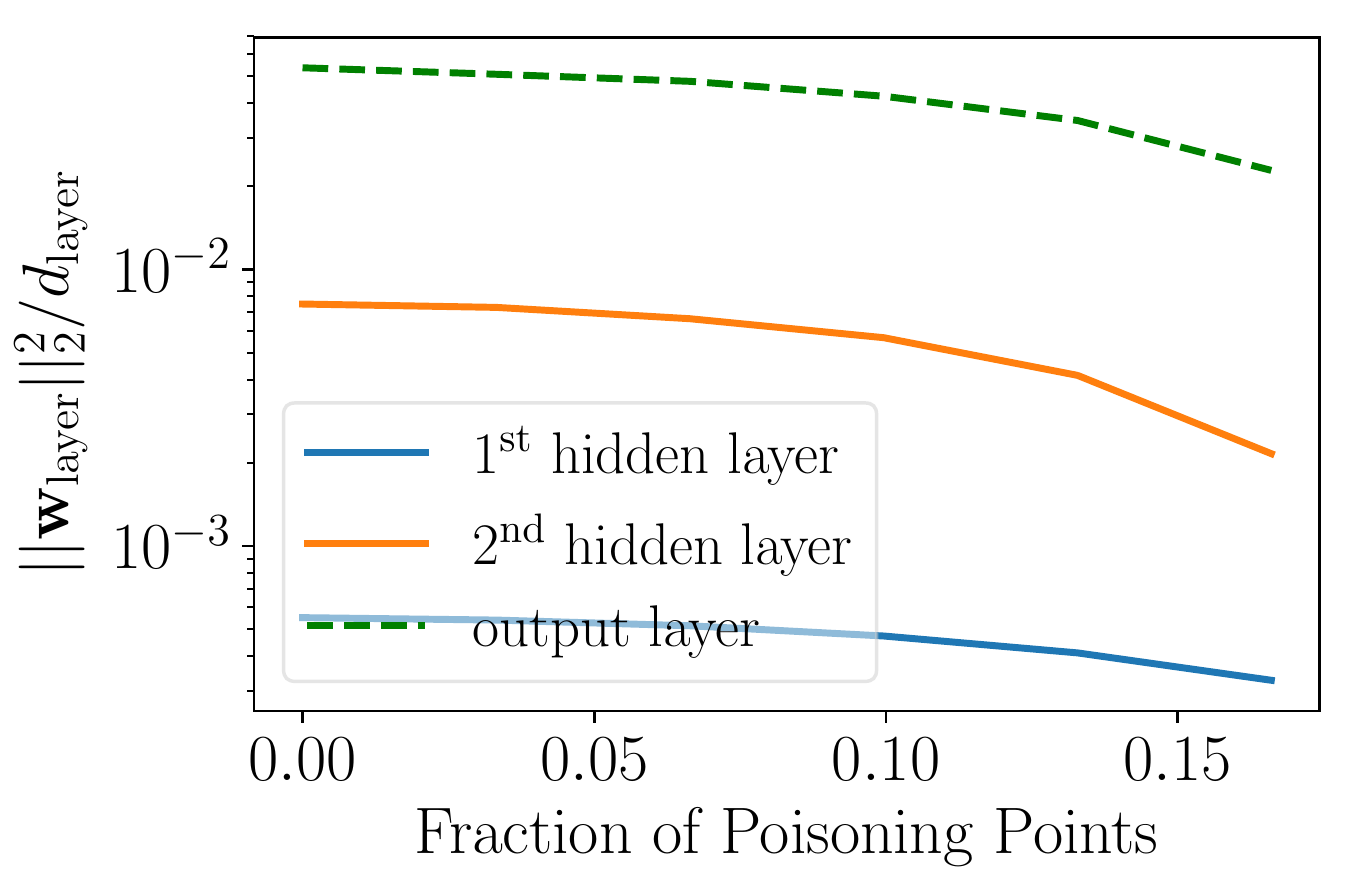}
			\caption{}
		\end{subfigure} 
	\end{centering}

	\caption{Results for the optimal attack against the DNN on ImageNet. (a) Average test error. $\lambda_{\text{RMD}}$ represents the one learned with RMD.  (b) Average $\lambda$ learned with RMD. (c) Average $||\textbf{w}_\text{layer}||_2^2/d_\text{layer}$, where $d_\text{layer}$ represents the number of parameters of the corresponding layer. This normalisation allows comparing $||\textbf{w}_\text{layer}||_2^2$ for each layer regardless of the number of the parameters.}
	\label{fig:dnnopt}
\end{figure*}

As in the previous experiment, we performed attacks with different strengths for the DNN trained with small ($\lambda=-8$) and large ($\lambda= \log(40)$) values for the regularisation hyperparameter, constant for all the layers. In this case, we omitted the case where $\lambda$ is set with $5$-fold cross-validation on the clean dataset as the search space is large, which makes it computationally very expensive. As before, we denote with $\lambda_{\text{RMD}}$ the case where the regularisation hyperparameter is learned (\ref{eqAttacker2}). The results in Fig.~\ref{fig:dnnopt}(a) are consistent with those obtained for the case of LR. For $\lambda=-8$ the algorithm is very vulnerable to the poisoning attack and its test error significantly increases up to $12\%$ when the training dataset is poisoned with $16.6\%$ malicious points. On the contrary, for $\lambda= \log(40)$ the algorithm's performance remains quite stable for increasing fractions of poisoning points. For $\lambda_{\text{RMD}}$ the test error increases moderately and stabilises when the fraction of poisoning points is $10\%$. Finally, from Fig.~\ref{fig:dnnopt}(a) we can see that when there is no attack, the test error for $\lambda_{\text{RMD}}$ is smaller than the other two cases. As discussed before, although over-regularising may be appealing to make the algorithm more robust to poisoning attacks, the performance in the absence of attack may be significantly worse. 

In Fig.~\ref{fig:dnnopt}(b) we can observe that the $\lambda$ learned for the second and output layers increase faster than the one for the first layer. This result suggests that the latter layers are more vulnerable to the poisoning attack. In other words, the poisoning attack tries to produce more changes in the parameters of the network in those layers and, at the same time, the network tries to resist to those changes by increasing the value of the corresponding regularisation hyperparameters. Finally, in Fig.~\ref{fig:dnnopt}(c) we show the $L_2$ norm of the parameters for each layer divided by the number of parameters in each case. The results are consistent with those in Fig.~\ref{fig:dnnopt}(b): The norm of the parameters in the last two layers significantly decrease when we increase the fraction of poisoning points. On the other side, regardless of the attack strength, the norm of the parameters is significantly bigger for the output layer.

\section{Conclusions}
\label{sec:conclusion}
In this paper we introduce a novel poisoning attack strategy for evaluating the robustness of ML classifiers that contain hyperparameters in worst-case scenarios. The attack can be formulated as a multiobjective bilevel optimisation problem that can be solved with gradient-based techniques. We show the limitations of previous approaches, such as \cite{biggio2012poisoning,xiao2015feature}, where the model's hyperparameters are considered constant regardless of the type and strength of the attack. As shown in our experiments, this approach can provide a misleading (and, often, pessimistic) view of the algorithms' robustness. We show that these poisoning attacks can have a strong influence on the hyperparameters learned, and thus, their influence must be considered when assessing the robustness of the algorithms. 

We also show that, contrary to the results reported in \cite{xiao2015feature}, $L_2$ regularisation can help to to mitigate the effect of poisoning attacks, increasing the stability of the learning algorithms. When the attacker increases the strength of the attack, the regularisation hyperparameter also increases to compensate the instability that the attacker tries to produce in the algorithm. Our experimental evaluation shows the benefits of using $L_2$ regularisation as a good practice to reduce the impact of poisoning attacks. We also illustrate the trade-off between robustness and accuracy as a function of the regularisation hyperparameter. Our results highlight the importance of stability as a desired property for the design of more robust algorithms to data poisoning. Further research work will include the analysis of attacks in multiclass settings, other forms of regularisation, such as $L_1$ and elastic-net, and the analysis of the robustness to data poisoning for other learning algorithms that contains hyperparameters.

\bibliography{paper}

\newpage

\appendix

\section{Hessian-Vector Products}

\label{sec:hvp}

Let $\textbf{x}\in \mathcal{X} \subseteq \mathbb{R}^m$, $\textbf{y} \in \mathcal{Y} \subseteq \mathbb{R}^n$, $\textbf{v}\in\mathbb{R}^m$, and $f(\textbf{x},\textbf{y})\in\mathbb{R}$. If the second partial derivatives of $f(\textbf{x},\textbf{y})$ are continuous \emph{almost everywhere}\footnote{A property
	that holds almost everywhere holds throughout all space except on a set of \emph{measure zero}. Intuitively, a set
	of measure zero occupies a negligible volume in the measured space \cite{goodfellow2016deep}.} in $\mathcal{X}$ and $\mathcal{Y}$ (mild condition), the Hessian-vector products $\left(\nabla_\textbf{x}^2f(\textbf{x},\textbf{y})\right)\textbf{v}$ and $\left(\nabla_\textbf{y}\nabla_\textbf{x}f(\textbf{x},\textbf{y})\right)\tran\textbf{v}$ can be computed exactly and efficiently by using the following identities  \cite{pearlmutter1994fast}:

\begin{gather}
	\left(\nabla_\textbf{x}^2f(\textbf{x},\textbf{y})\right)\textbf{v}=\nabla_\textbf{x}\left(\textbf{v}\tran\nabla_\textbf{x}f(\textbf{x},\textbf{y})\right),\hspace{1.5cm}
	\left(\nabla_\textbf{y}\nabla_\textbf{x}f(\textbf{x},\textbf{y})\right)\tran\textbf{v}=\nabla_\textbf{y}\left(\textbf{v}\tran\nabla_\textbf{x}f(\textbf{x},\textbf{y})\right).
\end{gather} 

The left and right expressions scale as $\mathcal{O}(m)$ and $\mathcal{O}(\max(m, n))$, respectively, both in time and space. Analogous expressions can be obtained when $\textbf{v}\in\mathbb{R}^n$. An elegant aspect of this technique is that, for machine learning models optimised with gradient-based methods, the equations for exactly evaluating
the Hessian-vector products emulate closely those for standard forward and backward propagation. Hence, the application of existing automatic differentiation frameworks to compute this product is typically straightforward \cite{pearlmutter1994fast, bishop2006pattern}. 

\section{Projected Hypergradient Descent/Ascent Algorithm}

\label{sec:hypdesasc}

Alg. \ref{alg:adreg} describes the procedure to solve the multiobjective bilevel problem proposed in Sect. \ref{sec:generalAttacks}. Essentially, this algorithm implements projected hypergradient descent/ascent to optimise, in a coordinate manner, the poisoning points---replaced into the training set---and the set of hyperparameters. 

Alg. \ref{alg:adreg} receives as inputs $\mathcal{M}$ (the target model), $\mathcal{A}$ and $\mathcal{L}$ (outer and inner objective functions),  $\mathcal{D}_\text{val}$ and $\mathcal{D}_\text{tr}$ (validation and training datasets), $n_\text{p}$ (number of poisoning points to optimise), $\mathcal{P}=\left\{p_k\right\}_{k=1}^{n_\text{p}}$ (indices of the $n_\text{p}$ points---uniformly sampled without duplicates---of $\mathcal{D}_\text{tr}$ to be replaced by $\mathcal{D}_\text{p}$),  $T_\text{mul}$ and $T$ (number of hyperiterations for the outer problem and number of epochs for the inner problem), and $\alpha$, $\beta$, and $\eta$ (learning rates for the features of the poisoning points, ${\bf X}_\text{p}$, for the hyperparameters, ${\boldsymbol \Lambda}$, and for the parameters of the inner problem, respectively). In addition, let $\textbf{w}$ and $\mathcal{D}_\text{p}$ denote the the model's parameters and the labelled poisoning points, correspondingly. At the end of the algorithm, it outputs the optimised poisoning points and hyperparameters, i.e. $\mathcal{D}_\text{p}^{(T_\text{mul})}$ and $\boldsymbol{\Lambda}^{(T_\text{mul})}$. For the sake of clarity, in the algorithms included in this appendix and the following ones we shall make use of the following notation:

\begin{gather*}
	\mathcal{A}\left(\textbf{w}^{(T)}\right)=\mathcal{A}\left(\mathcal{D}_\text{val}, \textbf{w}^{(T)}\right)=\mathcal{A}\left(\mathcal{M}\left(\textbf{X}_\text{val}, \textbf{w}^{(T)}\right), \textbf{y}_\text{val}\right),\\
	\mathcal{L}\left(\textbf{w}^{(t)}\right)=\mathcal{L}\left(\mathcal{D}_\text{tr}'^{(\tau)}, \boldsymbol{\Lambda}^{(\tau)}, \textbf{w}^{(t)}\right)=\mathcal{L}\left(\mathcal{M}\left(\textbf{X}_\text{tr}'^{(\tau)}, \boldsymbol{\Lambda}^{(\tau)}, \textbf{w}^{(t)}\right), \textbf{y}_\text{tr}'\right),
\end{gather*}

where $\textbf{X}_\text{val}$ and $\textbf{X}_\text{tr}'$ are the features of the validation and poisoned training sets, and $\textbf{y}_\text{val}$ and $\textbf{y}_\text{tr}'$ are their respective labels.

To reduce the computational burden, we consider the simultaneous optimisation of a batch of $n_\text{p}$ poisoning points, $\mathcal{D}_\text{p} = \{(\textbf{x}_{\text{p}_k} ,y_{\text{p}_k})\}^{n_\text{p}}_{k=1}$. We generate the initial values of $\mathcal{D}_\text{p}$ by cloning $n_\text{p}$ samples---uniformly sampled without duplicates---of $\mathcal{D}_\text{val}$. Their labels are initially flipped and
kept fixed during the optimisation. This process is carried out in the function \texttt{initDp} (Line \ref{lin:initdp}). Then, these $n_\text{p}$ poisoning samples replace the $n_\text{p}$ clean samples of $\mathcal{D}_\text{tr}$ whose indices are in the set $\mathcal{P}$ (Line \ref{lin:upddtr1}).  On the other hand, the hyperparameters are initialised in \texttt{initL} (Line \ref{lin:initl}). In the case study of this paper, the hyperparameters correspond to the regularisation ones.

In a bilevel optimisation problem, the outer problem depends on the solution of the inner problem. Hence, in order to solve the bilevel problem, every time (hyperiteration) the variables in the outer problem (i.e. $\textbf{X}_\text{p}$ and $\boldsymbol{\Lambda}$ in this case) are updated, the variables in the inner problem (i.e. the model's parameters) need to be previously initialised and optimised. Thus, let \texttt{initW} (Line \ref{lin:initw1}) be a particular initialisation for the model's parameters. $\texttt{hypGradXp}$ (Line \ref{lin:hypgradxp}) and $\texttt{hypGradL}$ (Line \ref{lin:hypgradl}) refer to the particular optimisation algorithm used to train the model's parameters and compute the corresponding hypergradients. In this work, this algorithm is Reverse-Mode Differentiation (RMD) (Alg. \ref{alg:bg}). Then, $\Pi_{\Phi(\mathcal{D}_\text{p})}$ (Line \ref{lin:phga}) and $\Pi_{\Phi({\boldsymbol \Lambda})}$ (Line \ref{lin:phgd}) are the projection operators for ${\bf X}_\text{p}$ and ${\boldsymbol \Lambda}$, so that their updated values are within the corresponding feasible domains, $\Phi(\mathcal{D}_\text{p})$ and $\Phi({\boldsymbol \Lambda})$. Line \ref{lin:upddtr2} updates the features of the poisoning samples in the poisoned training set.

In our experiments, we simulate different ratios of poisoning points in a cumulative manner: Once the optimisation of the current batch of poisoning points and hyperparameters is finished,\footnote{The criterion to finish the loop that optimises the variables of the outer level problem is given by the number of hyperiterations.} this batch of poisoning points is fixed and the next batch of poisoning points is replaced into the remaining clean training set, to carry out their corresponding optimisation.  To accelerate their optimisation, the hypergradients for the poisoning points are normalised with respect to their $L_2$ norm.\footnote{The analysis of other techniques to accelerate the optimisation, such as adaptive learning rates, is left for future work.} Regarding the hyperparameters, their values are updated continuously for the whole set of batches of poisoning points.

\begin{algorithm}[!t]
	\caption{Projected Hypergradient Descent/Ascent}
	\label{alg:adreg}
	
	\begin{flushleft}
		{\bfseries Input:} $\mathcal{M}$, $\mathcal{A}$, $\mathcal{L}$, $\mathcal{D}_\text{val}$, $\mathcal{D}_\text{tr}$,  $n_\text{p}$, $\mathcal{P}$, $T_\text{mul}$, $T$, $\alpha$, $\beta$, $\eta$ \\
		{\bfseries Output:} $\mathcal{D}_\text{p}^{(T_\text{mul})}$, $\boldsymbol{\Lambda}^{(T_\text{mul})}$
	\end{flushleft}
	
	\begin{algorithmic}[1]
		\STATE $\mathcal{D}_\text{p}^{(0)} \leftarrow \texttt{initDp}\left(\mathcal{D}_\text{val}, n_\text{p}\right)$ \COMMENT{generate $n_\text{p}$ samples for $\mathcal{D}_\text{p}^{(0)}$ }
		
		\label{lin:initdp}

		\STATE $\mathcal{D}_\text{tr}'^{(0)}  \leftarrow \left( \mathcal{D}_\text{tr} \setminus  \{(\textbf{x}_{\text{tr}_k} ,y_{\text{tr}_k})\}_{k\in \mathcal{P}} \right) \cup \mathcal{D}_\text{p}^{(0)}$ \COMMENT{replace $n_\text{p}$ samples of $\mathcal{D}_\text{\text{tr}}$ by  $\mathcal{D}_\text{p}^{(0)}$ }
		
		\label{lin:upddtr1}

		\STATE $\boldsymbol{\Lambda}^{(0)}\leftarrow \texttt{initL}(\mathcal{M})$ \COMMENT{initialise $\boldsymbol{\Lambda}$}
		
		\label{lin:initl}

		\FOR{$\tau=0$ \textbf{to} $T_\text{mul}-1$}
		
		\STATE 	$\textbf{w}^{(0)} \leftarrow \texttt{initW}(\mathcal{M})$ \COMMENT{initialise \textbf{w}}
		\label{lin:initw1}
		\STATE $ \nabla_{\textbf{X}_\text{p}}\mathcal{A}\left(\textbf{w}^{(T)}\right)
		\leftarrow \texttt{hypGradXp}\left(\mathcal{M},\mathcal{A}, {\mathcal{L}, \mathcal{D}_\text{val}, \mathcal{D}_\text{tr}'^{(\tau)},  \boldsymbol{\Lambda}^{(\tau)}}, \textbf{w}^{(0)}, T, \eta\right)$
		\label{lin:hypgradxp} \COMMENT{Alg. \ref{alg:bg}}	
		\STATE $\textbf{X}_\text{p}^{(\tau+1)}\leftarrow \Pi_{\Phi \left(\mathcal{D}_\text{p}\right)} \left(\textbf{X}_\text{p}^{(\tau)} + \alpha\nabla_{\textbf{X}_\text{p}}\mathcal{A}\left(\textbf{w}^{(T)}\right)\right)$ \COMMENT{projected hypergradient ascent}
		\label{lin:phga}

		\STATE $\textbf{X}_\text{tr}'^{(\tau+1)}  \leftarrow \left( \textbf{X}_\text{tr}'^{(\tau)} \setminus \textbf{X}_\text{p}^{(\tau)} \right) \cup \textbf{X}_\text{p}^{(\tau+1)}$ \COMMENT{update $\textbf{X}_\text{\text{tr}}'^{(\tau)}$ with $\textbf{X}_\text{p}^{(\tau+1)}$ }
		
		\label{lin:upddtr2}
		
		\STATE $ \nabla_{\boldsymbol{\Lambda}}\mathcal{A}\left(\textbf{w}^{(T)}\right)
		\leftarrow \texttt{hypGradL}\left({\mathcal{M},\mathcal{A}, \mathcal{L},  \mathcal{D}_\text{val}, \mathcal{D}_\text{tr}'^{(\tau+1)},  \boldsymbol{\Lambda}^{(\tau)}},\textbf{w}^{(0)}, T, \eta\right)$ \COMMENT{Alg. \ref{alg:bg}}
		\label{lin:hypgradl}

		\STATE $\boldsymbol{\Lambda}^{(\tau+1)}\leftarrow \Pi_{\Phi\left(\boldsymbol{\Lambda}\right)}\left(\boldsymbol{\Lambda}^{(\tau)} - \beta\nabla_{\boldsymbol{\Lambda}}\mathcal{A}\left(\textbf{w}^{(T)}\right)\right)$ \COMMENT{projected hypergradient descent}
		\label{lin:phgd}
		
		\ENDFOR
		
	\end{algorithmic}
	
\end{algorithm}

\section{Reverse-Mode Differentiation Algorithm}

As described in \cite{franceschi2017forward}, we can think of the training algorithm as a discrete-time dynamical system, described by a set of states ${\bf s}^{(t)} \in\mathbb{R}^{d_\text{s}}$, with $t = 0, \ldots, T$, where each state depends on model's parameters, the accumulated gradients and/or the velocities. Then, from a reduced number of training iterations, $T$, we can estimate the hypergradients from the values of the parameters collected in the set of states. Depending on the order to compute the operations we can differentiate two approaches to estimate the hypergradients: Reverse-Mode (RMD) and Forward-Mode Differentiation (FMD) \cite{griewank2008evaluating,franceschi2017forward}. In the first case, RMD requires first to train the learning algorithm for $T$ epochs, and then, to compute ${\bf s}^{(0)}$ to ${\bf s}^{(T)}$. Then, the hypergradients estimate is computed by reversing the steps followed by the learning algorithm from ${\bf s}^{(T)}$ down to ${\bf s}^{(0)}$. On the other hand, FMD computes the estimate of the hypergradients as the algorithm is trained, i.e. from ${\bf s}^{(0)}$ to ${\bf s}^{(T)}$ (i.e. the estimates can be computed in parallel with the training procedure).

In some cases, as in \cite{domke2012generic}, RMD requires to store all the information collected in the states in the forward pass. This can be prohibitive in some cases as, for example, for deep networks where the number of parameters is huge. However, other RMD methods proposed in the literature do not require to store this information \cite{maclaurin2015gradient,munoz2017towards}. To estimate the hypergradients, RMD requires to compute a forward and a backward pass through the set of states. In contrast, FMD just needs to do the forward computation. However, the scalability of FMD depends heavily on the number of hyperparameters compared to RMD. Then, for problems where the number of hyperparameters is large, as it is the case for the poisoning attacks we introduced in Sect. \ref{sec:generalAttacks}, RMD is computationally more efficient to estimate the hypergradients. For this reason, we used RMD in our experiments.

Here we include the Reverse-Mode Differentiation (RMD) algorithm (Alg. \ref{alg:bg}) \cite{domke2012generic, maclaurin2015gradient, munoz2017towards}, which we use to compute the hypergradient estimate at the outer level problem (both for the features of the poisoning points, $\textbf{X}_\text{p}$, and the hyperparameters, $\boldsymbol{\Lambda}$). We make use of a similar notation to the algorithms presented in \cite{maclaurin2015gradient, munoz2017towards}. The resulting algorithm for the case of $\boldsymbol{\Lambda}$ is analogous.

\begin{algorithm}[!t]
	\caption{Reverse-Mode Differentiation}
	\label{alg:bg}
	
	\begin{flushleft}
		
		{\bfseries Input:}  $\mathcal{M}$, $\mathcal{A}$, $\mathcal{L}$,  $\mathcal{D}_\text{val}$, $\mathcal{D}_\text{tr}'^{(\tau)}$, $\boldsymbol{\Lambda}^{(\tau)}$, $\textbf{w}^{(0)}$,  $T$, $\eta$ \\
		{\bfseries Output:} $\nabla_{\textbf{X}_\text{p}}\mathcal{A}\left(\textbf{w}^{(T)}\right)$
		
	\end{flushleft}
	
	\begin{algorithmic}[1]
		
		\FOR{$t=0$ \textbf{to} $T-1$}
		
		\STATE $\textbf{g} \leftarrow  \nabla_{\textbf{w}_{}}\mathcal{L}\left(\textbf{w}^{(t)}\right)$ 
		
		\STATE $\textbf{w}^{(t+1)}\leftarrow \textbf{w}^{(t)} - \eta \textbf{g}$ \COMMENT{stochastic gradient descent}
		\ENDFOR
		
		\STATE $d\textbf{X}_\text{p}^{(T)} \leftarrow \textbf{0}$
		
		\STATE $d\textbf{w}^{(T)} \leftarrow \nabla_{\textbf{w}_{}}\mathcal{A}\left(\textbf{w}^{(T)}\right)$
		\FOR{$t=T-1$ \textbf{down to} $0$}
		
		\STATE $d\textbf{g}_\text{w} \gets \left(\nabla^2_{\textbf{w}_{}}\mathcal{L}\left(\textbf{w}^{(t)}\right)\right)d\textbf{w}^{(t+1)} $ \COMMENT{Hessian-vector product (Appendix \ref{sec:hvp})}
		\STATE $d\textbf{g}_{\text{X}_\text{p}} \gets \left(\nabla_{\textbf{X}_\text{p}}\nabla_{\textbf{w}_{}}\mathcal{L}\left(\textbf{w}^{(t)}\right)\right)\tran d\textbf{w}^{(t+1)} $ \COMMENT{Hessian-vector product (Appendix \ref{sec:hvp})}
		
		\STATE	$
		d\textbf{w}^{(t)}  \leftarrow d\textbf{w}^{(t+1)}  - \eta d\textbf{g}_\text{w}
		$
		
		\STATE	$
		d\textbf{X}_\text{p}^{(t)}  \leftarrow d\textbf{X}_\text{p}^{(t+1)}  - \eta d\textbf{g}_{\text{X}_\text{p}}
		$

		\ENDFOR
		\STATE $ \nabla_{\textbf{X}_\text{p}}\mathcal{A}\left(\textbf{w}^{(T)}\right)
		\leftarrow d\textbf{X}_\text{p}^{(0)}$

	\end{algorithmic}
\end{algorithm}

\section{Experimental Settings}

\label{sec:expset}

In the following specifications, $T_{\mathcal{D}_\text{p}}$ denotes the number of hyperiterations at the outer problem when the poisoning points are learned and $\lambda$ is fixed, $T_\lambda$ when the training set is clean and $\lambda$ is learned, and $T_\text{mul}$ when $\textbf{X}_\text{p}$ and $\lambda$ are learned in a coordinate manner. Finally, $\eta_\text{tr}$ is the learning rate when testing the attack.

\subsection{Architecture of the Models and Initialisation of Their Parameters}

The Logistic Regression (LR) classifier's parameters are always initialised with zeros, for all the datasets. 

In contrast, the Deep Neural Network (DNN) model---trained on ImageNet---has two hidden layers ($2,048\times128\times32\times1$) with Leaky ReLU activation functions (with a negative slope of $0.1$), which sums up to a total of $266,433$ parameters. These parameters are initially filled with values according to the Xavier Initialisation method \cite{glorot2010understanding}, using a uniform distribution for all the parameters except the bias terms, which are initialised with a value of $10^{-2}$.

\subsection{Synthetic Example}

\label{sec:setsynthetic}

For the synthetic experiment shown in Sect. \ref{sec:L2}, we sample the attacker's data from two bivariate Gaussian distributions, $\mathcal{N}(\boldsymbol{\mu}_0, \boldsymbol{\Sigma}_0)$ and $\mathcal{N}(\boldsymbol{\mu}_1, \boldsymbol{\Sigma}_1)$, with mean vectors $\boldsymbol{\mu}_i$ and covariance matrices $\boldsymbol{\Sigma}_i$ for each class, $i=0,1$:

\begin{equation*}
	\begin{aligned}
		\boldsymbol{\mu}_0 & = \begin{bmatrix} -3.0 \\ \hspace{\minuslength}0.0 \end{bmatrix}, & \qquad \boldsymbol{\Sigma}_0 & = \begin{bmatrix} 2.5 & 0.0 \\ 0.0 & 1.5 \end{bmatrix}, \\
		\boldsymbol{\mu}_1  & = \begin{bmatrix} 3.0 \\ 0.0 \end{bmatrix}, &         \boldsymbol{\Sigma}_1 & = \begin{bmatrix} 2.5 & 0.0 \\ 0.0 & 1.5 \end{bmatrix}. \\
	\end{aligned}
\end{equation*}

The attacker uses $32$ points ($16$ per class) for training and $64$ ($32$ per class) for validation, and one poisoning point cloned from the validation set (in the example of Sect. \ref{sec:L2}, cloned from the set labelled as blue), whose label is flipped. This poisoning point is concatenated into the training set and the features of this point are optimised with RMD. In order to poison the LR classifier, we use a learning rate and a number of hyperiterations for the poisoning point $\alpha=0.4$, $T_{\mathcal{D}_\text{p}}=50$; the feasible domain $\Phi(\mathcal{D}_\text{p})\in[-9.5,9.5]^2$; the learning rate and number of epochs for the inner problem $\eta=0.2$, $T=500$; and when evaluating the attack, $\eta_\text{tr}=0.2$, batch size $=32$ (full batch), and $\text{number of epochs} = 100$. When we apply regularisation, we fix $\lambda=\log(20)\approx3$.

To plot the colour-map that shows the value of the regularisation hyperparameter learned (Fig.~\ref{fig:synthetic}(right)) as a function of the poisoning point injected in the training set, the values of $\lambda$ explored for each possible poisoning point are in the range $[-8, 6]$. Then, the optimal value of $\lambda$ is chosen such that it minimises the error of the model, trained on each combination of the poisoning point (concatenated into the training set) and $\lambda$ in the grid, and evaluated on the validation set.

\subsection{MNIST, FMNIST and ImageNet}

\begin{table*}[t]
	
	\centering
	\caption{Characteristics of the datasets used in the experiments.}{ 
		\begin{tabular}{|l|c|c|c|c|}
			\hline
			Name & \makecell{\# Training \\ Samples} & \makecell{\# Validation \\ Samples} & \makecell{\# Test \\ Samples} & \# Features \\
			\hline
			MNIST (`0' vs `8') & $512$ & $171$ & $1,954$ & $784$ \\
			FMNIST (\emph{T-shirt} vs \emph{pullover}) & $512$ & $171$ & $2,000$ & $784$ \\
			ImageNet (\emph{dog} vs \emph{fish}) & $512$ & $171$ & $600$ & $2,048$ \\
			\hline
	\end{tabular}}
	\label{tabDatasets}
	
\end{table*}

\begin{samepage}

	\begin{table*}[!t]
		
		\centering
		\caption{Experimental settings for the poisoning attack.}{ 
			\begin{tabular}{|l|c|c|c|c|c|}
				\hline
				Name &  $T_{\mathcal{D}_\text{p}}$ &  $T_\lambda$ &  $T_\text{mul}$ &  $\alpha$ & $\beta$ \\
				\hline
				MNIST (`0' vs `8') (LR) & $100$ & $50$ & $100$ & $0.99$ & $0.80$ \\
				FMNIST (\emph{T-shirt} vs \emph{pullover}) (LR) & $100$ & $50$ & $100$ & $0.90$ & $0.30$ \\
				ImageNet (\emph{dog} vs \emph{fish}) (LR) & $100$ & $50$ & $200$ & $0.90$ & $0.40$ \\
				ImageNet (\emph{dog} vs \emph{fish}) (DNN) & $100$ & $75$ & $150$ & $0.99$ & $0.03$ \\
				\hline
			\end{tabular}
			
			\vspace{0.2cm}
			
			\begin{tabular}{|l|c|c|c|c|}
				\hline
				Name & $\Phi(\mathcal{D}_\text{p})$ & $\Phi(\lambda)$ & $\eta$ & $T$  \\
				\hline
				MNIST (`0' vs `8') (LR) & $[0.0, 1.0]^{784}$ & $[-8,\log(200)]$ & $0.10$ & $150$ \\
				FMNIST (\emph{T-shirt} vs \emph{pullover}) (LR) & $[0.0, 1.0]^{784}$ & $[-8,\log(400)]$ & $0.08$ & $200$\\
				ImageNet (\emph{dog} vs \emph{fish}) (LR) & $[-0.5, 0.5]^{2,048}$ & $[-8,\log(20,000)]$ & $0.05$ & $200$ \\
				ImageNet (\emph{dog} vs \emph{fish}) (DNN) & $[-0.5, 0.5]^{2,048}$ & $[-8,\log(60)]$ & $0.01$ & $600$ \\
				\hline
		\end{tabular}}
		\label{tabAttack}
		
	\end{table*}
	
	\begin{table*}[!t]
		\centering
		\caption{Experimental settings for training the models.}{ 
			\begin{tabular}{|l|c|c|c|}
				\hline
				Name &  $\eta_\text{tr}$ &  Batch Size &  Number of Epochs \\
				\hline
				MNIST (`0' vs `8') (LR) & $10^{-2}$ & $64$ & $200$ \\
				FMNIST (\emph{T-shirt} vs \emph{pullover}) (LR) & $10^{-2}$ & $64$ & $200$ \\
				ImageNet (\emph{dog} vs \emph{fish}) (LR) & $10^{-3}$ & $64$ & $300$ \\
				ImageNet (\emph{dog} vs \emph{fish}) (DNN) & $10^{-4}$ & $64$ & $4,000$  \\
				\hline
		\end{tabular}}
		\label{tabTrain}
	\end{table*}
	
\end{samepage}

In our experiments, all the results for MNIST `0' vs `8' \cite{lecun1998gradient}, Fashion-MNIST (FMNIST) \emph{T-shirt} vs \emph{pullover} \cite{xiao2017fashion}, and ImageNet \emph{dog} vs \emph{fish} \cite{russakovsky2015imagenet} (preprocessed as in \cite{koh2017understanding}) are the average of $10$ repetitions with different random data splits for training and validation, whereas the test set is fixed. For all the attacks we measure the average test error for different attack strengths, where the number of poisoning points is in the range between $0$ and $85$.  To reduce the computational cost, the size of the batch of poisoning points that are simultaneously optimised (as explained in Appendix \ref{sec:hypdesasc}) is $17$ for all the datasets. This way, we simulate six different ratios of poisoning ranging from $0\%$ to $16.6\%$.

The details of each dataset are included in Table \ref{tabDatasets}. All the datasets are balanced. Moreover, both MNIST and FMNIST sets are normalised to be in the range $[0, 1]^{784}$, whereas for the ImageNet sets we use the same Inception-v3 \cite{szegedy2016rethinking} features as in \cite{koh2017understanding}, and normalised them with respect to their training mean and standard deviation.

For all the experiments, we make use of stochastic gradient descent both to update the parameters in the forward pass of RMD\footnote{To refine the solutions obtained, when the poisoning points are learned and $\lambda$ is fixed, and when the training set is clean and $\lambda$ is learned, if the optimisation algorithm gets stuck in a poor local optimum we restart the bilevel optimisation procedure. In the case of the poisoning points, we reinitialise them with different values uniformly sampled without duplicates from the validation set. For the regularisation hyperparameters, let $\lambda^{(\tau)}$ denote their latest value: We reinitialise them with values uniformly sampled in the range $\left[\lambda^{(\tau)}-0.5, \lambda^{(\tau)}+0.5\right]$. It is however noteworthy that these reinitialisations were not crucial to obtain the results shown in this paper. The exploration of warm-restart techniques in the case of multiobjective bilevel problems is left for future work.} (full batch training), and to train the model when testing the attack (mini-batch training). The details of the attack settings are shown in Table \ref{tabAttack}, whereas the ones for training are in Table \ref{tabTrain}. Additionally, $\lambda_\text{CLEAN}$ is optimised with $5$-fold cross-validation---training the model on the clean dataset, as in \cite{xiao2015feature}. For all the datasets the ranges of values of $\lambda$ explored to compute $\lambda_\text{CLEAN}$ is $[-8, 1]$ (no better performance was observed for larger values of $\lambda$).  On the other hand, to accelerate the optimisation of $\lambda_\text{RMD}$, when the training set is clean these hyperparameters are warm-started with a value $\lambda = \log(5)\approx1.61$.

\subsection{Details of the Hardware Used}

All the experiments are run on $2 \times 11$~GB NVIDIA GeForce\textregistered~GTX 1080 Ti GPUs. The RAM memory is $64$~GB ($4\times16$~GB) Corsair VENGEANCE DDR4 $3000~\text{MHz}$. The processor (CPU) is Intel\textregistered~Core\texttrademark~i7 Quad Core Processor i7-7700k ($4.2$~GHz) $8$~MB Cache.

\section{Additional Results}

\subsection{Test False Positive and False Negative Rate}

\subsubsection{Logistic Regression}

\begin{figure*}[tp]
	\begin{subfigure}[b]{0.325\textwidth}
		\includegraphics[width=\textwidth]{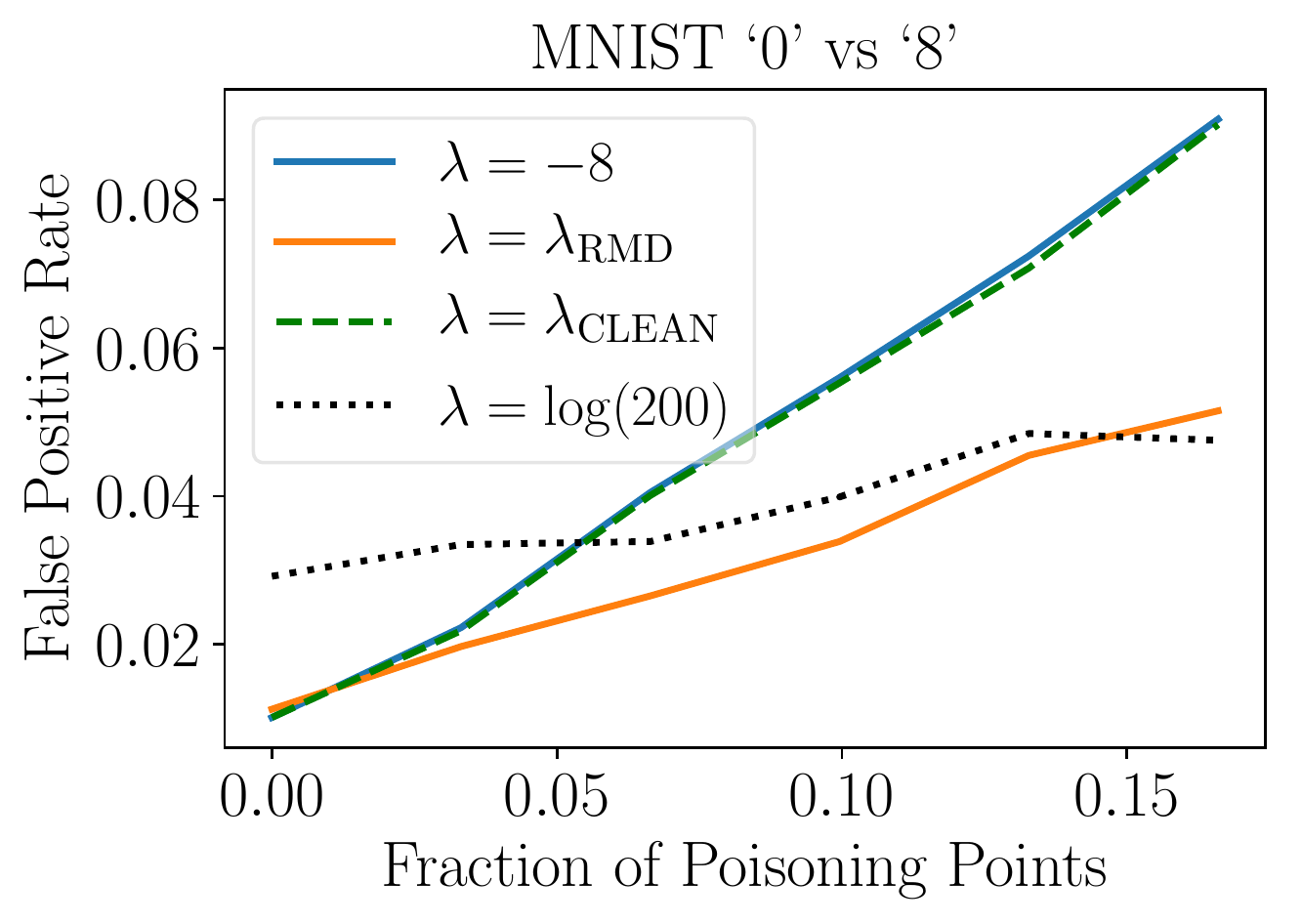}
		\caption{}
	\end{subfigure}
	\enskip % Control spacing between left and right figure, can use \enskip, \quad, \qquad, \hfill
	\begin{subfigure}[b]{0.306\textwidth}
		\includegraphics[width=\textwidth]{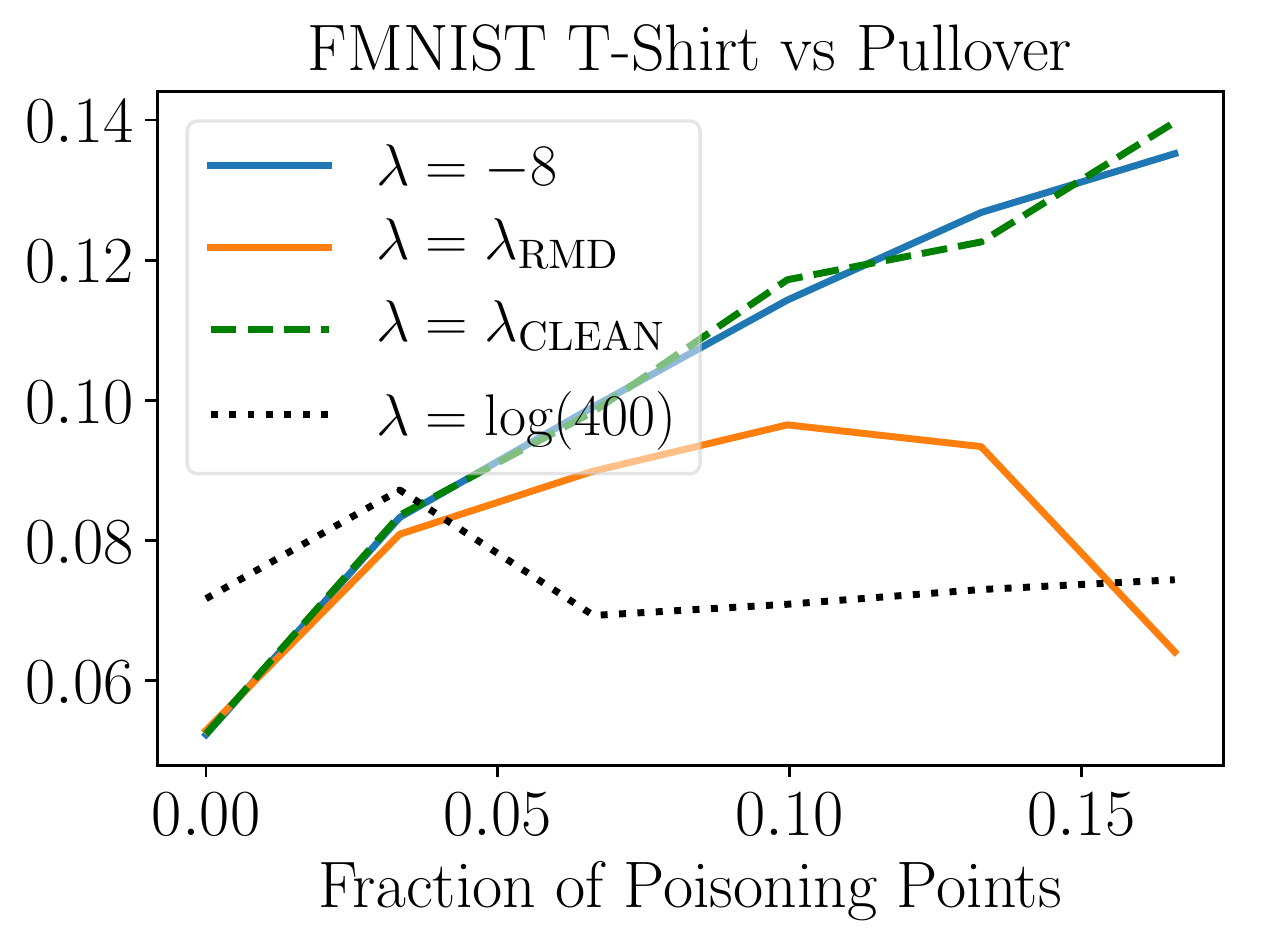}
		\caption{}
	\end{subfigure}
	\enskip % Control spacing between left and right figure, can use \enskip, \quad, \qquad, \hfill
	\begin{subfigure}[b]{0.33\textwidth}
		\includegraphics[width=\textwidth]{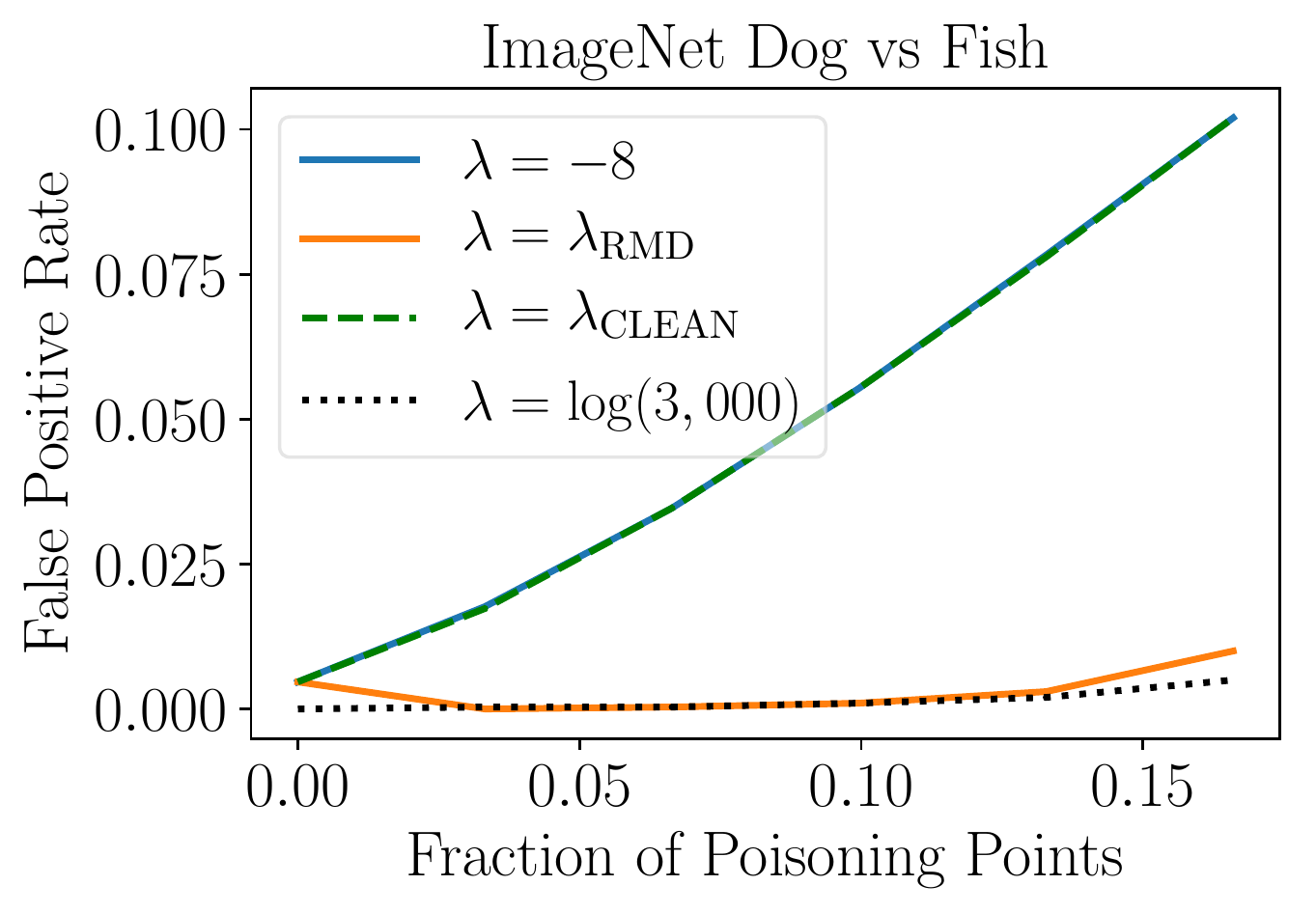}
		\caption{}
	\end{subfigure}
	\enskip % Control spacing between left and right figure, can use \enskip, \quad, \qquad, \hfill
	\begin{subfigure}[b]{0.323\textwidth}
		\includegraphics[width=\textwidth]{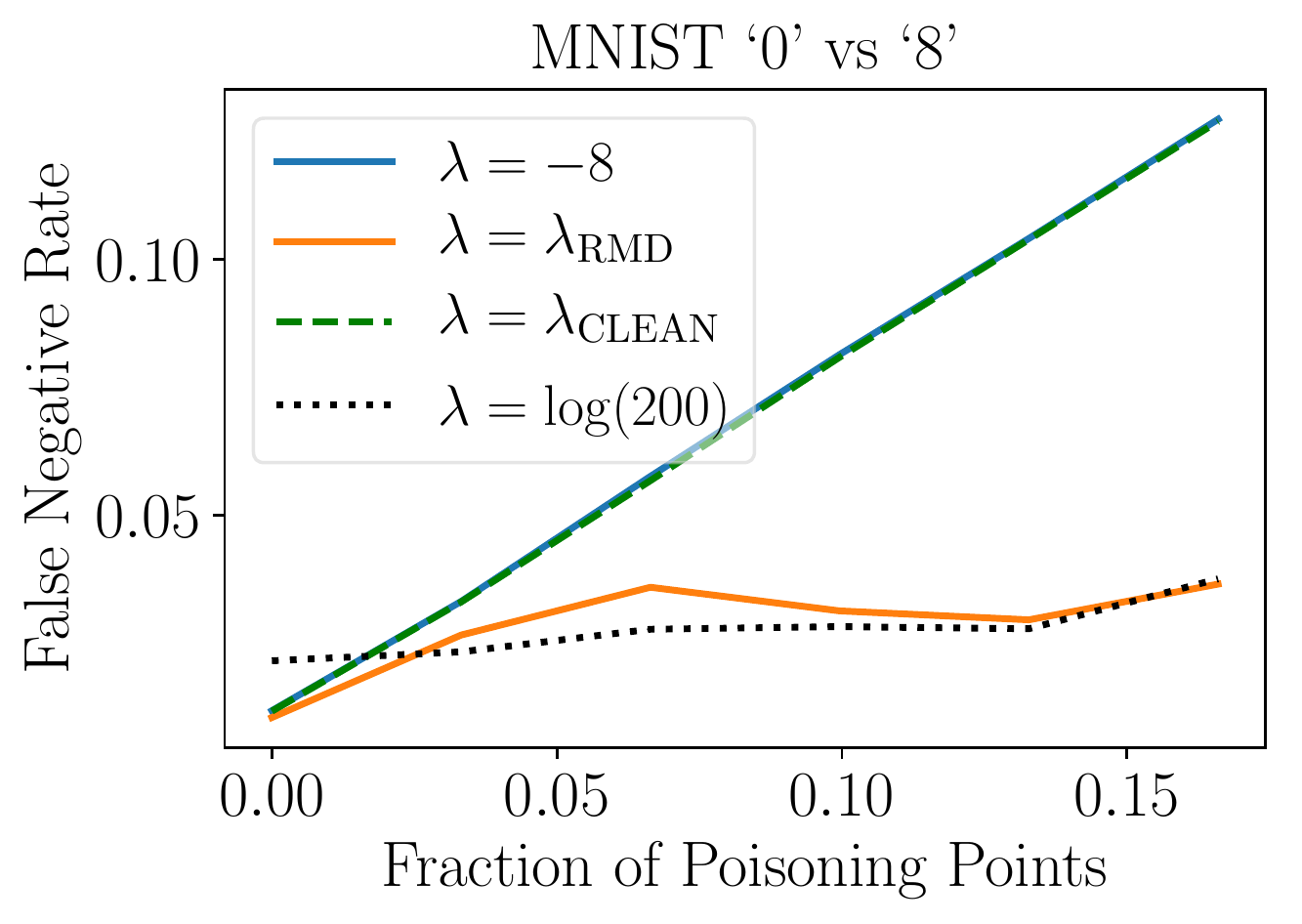}
		\caption{}

	\end{subfigure}
	\enskip % Control spacing between left and right figure, can use \enskip, \quad, \qquad, \hfill
	\begin{subfigure}[b]{0.306\textwidth}
		\includegraphics[width=\textwidth]{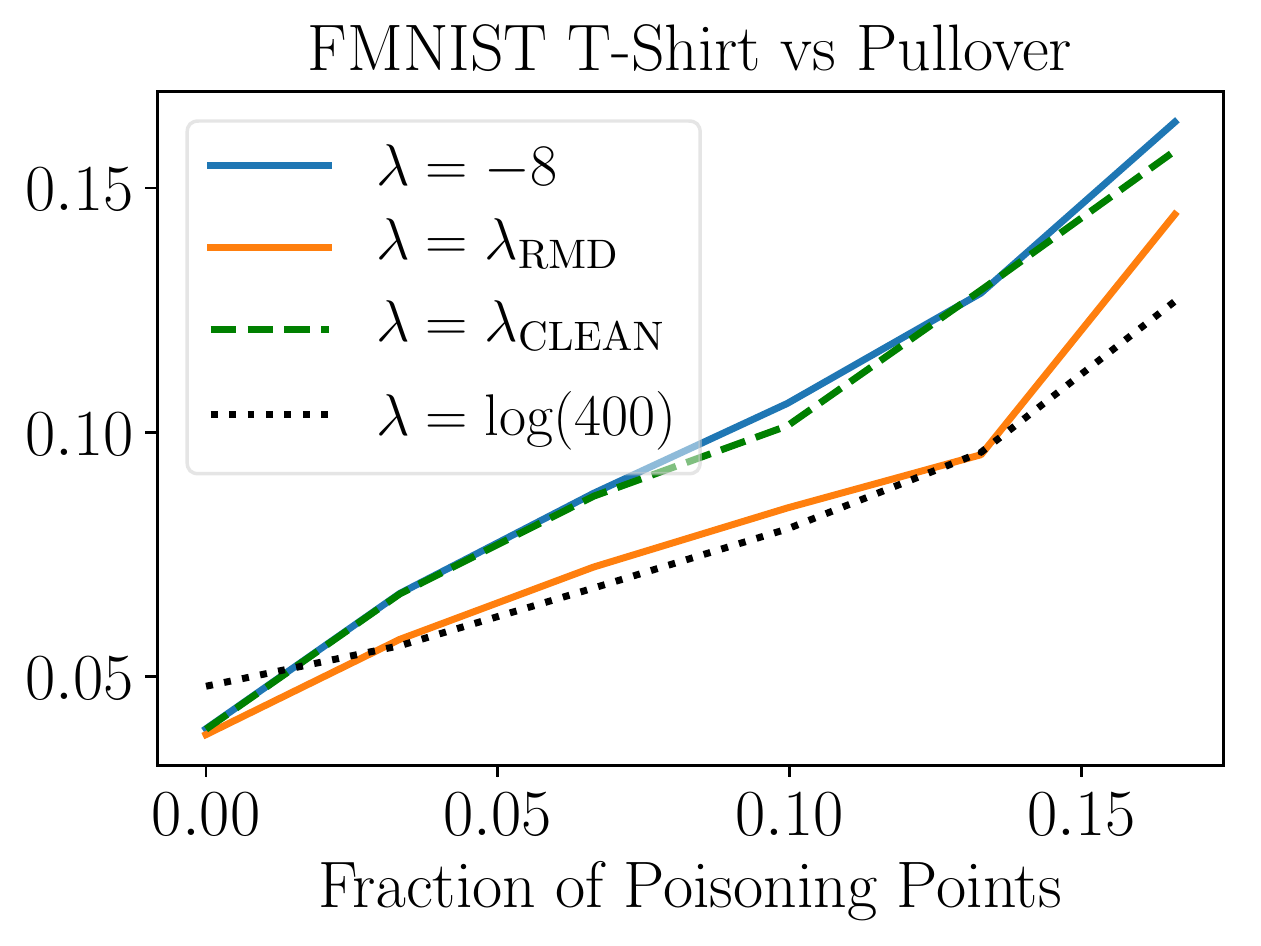}
		\caption{}		
	\end{subfigure}
	\enskip % Control spacing between left and right figure, can use \enskip, \quad, \qquad, \hfill
	\begin{subfigure}[b]{0.33\textwidth}
		\includegraphics[width=\textwidth]{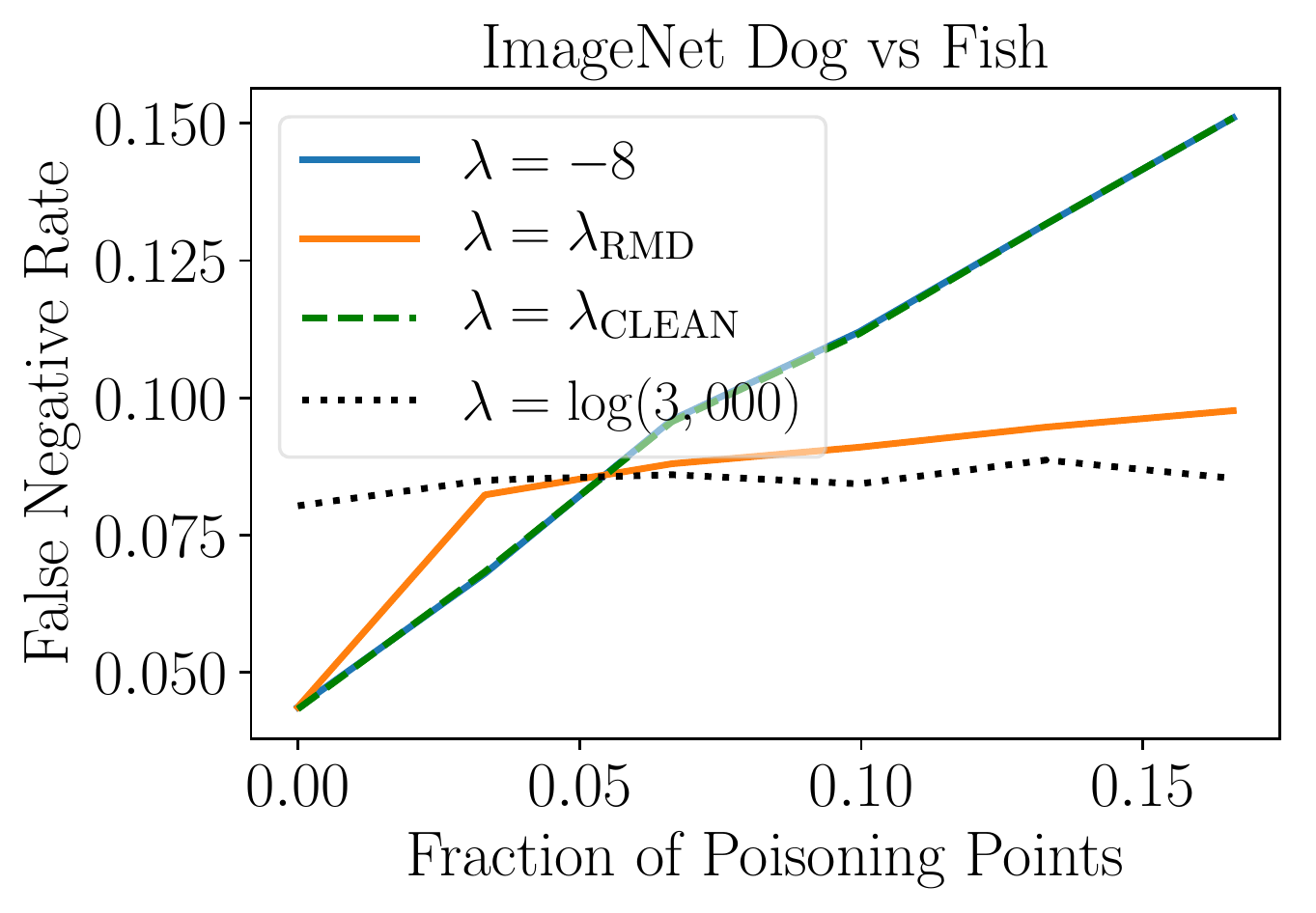}
		\caption{}

	\end{subfigure}

	\caption{Average test false positive and false negative rate for LR. $\lambda_{\text{RMD}}$ represents the one learned with RMD. The first row represents the test false positive rate: (a) MNIST, (b) FMNIST, and (c)~ImageNet. The second row depicts the test false negative rate: (d) MNIST, (e) FMNIST, and (f)~ImageNet.}
	\label{fig:lrfpr}
\end{figure*}

\begin{figure*}[!t]
	\begin{subfigure}[b]{0.323\textwidth}
		\includegraphics[width=\textwidth]{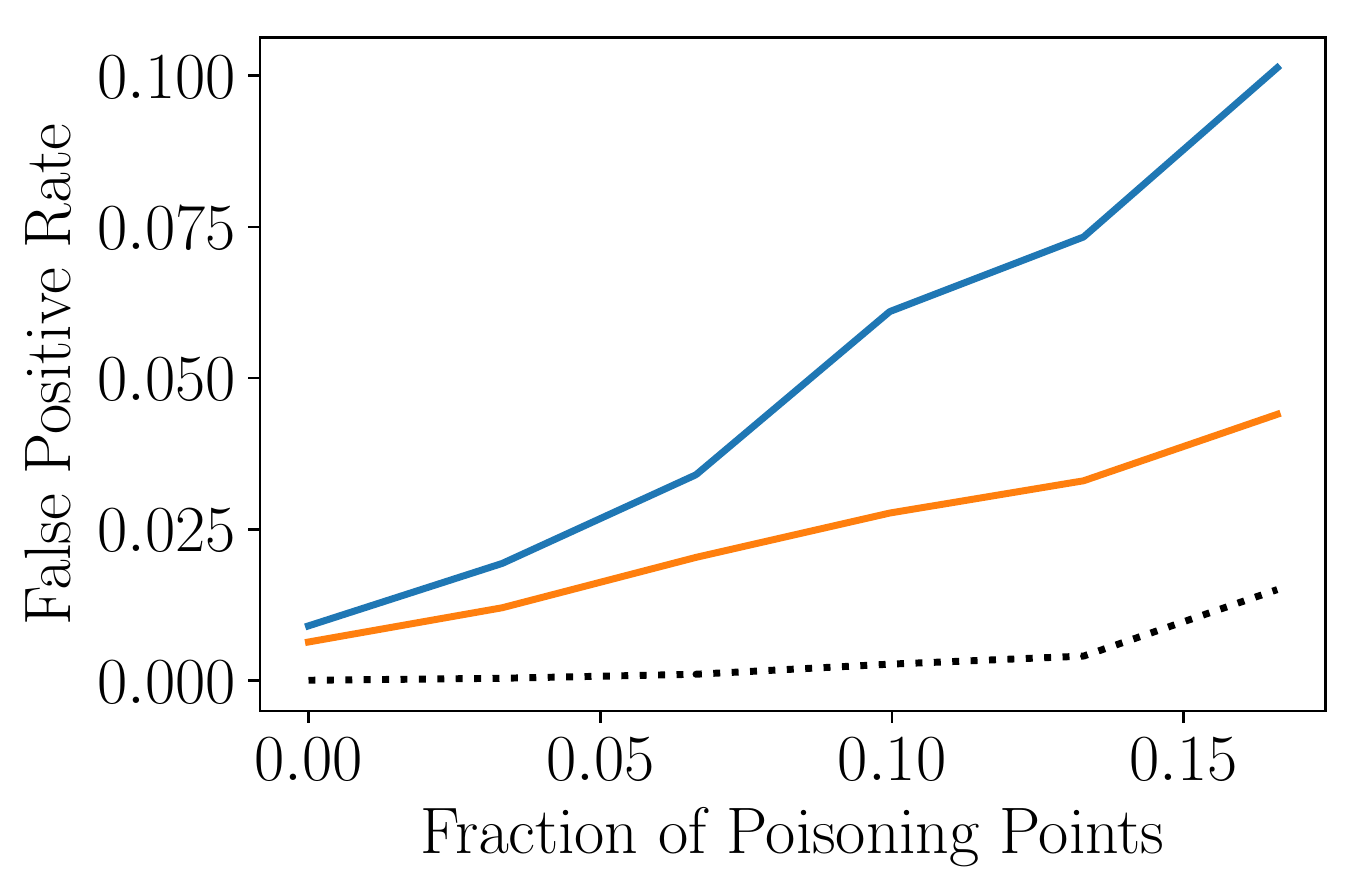}
		\caption{}
		
	\end{subfigure}
	\enskip % Control spacing between left and right figure, can use \enskip, \quad, \qquad, \hfill
	\begin{subfigure}[b]{0.317\textwidth}
		
		\includegraphics[width=\textwidth]{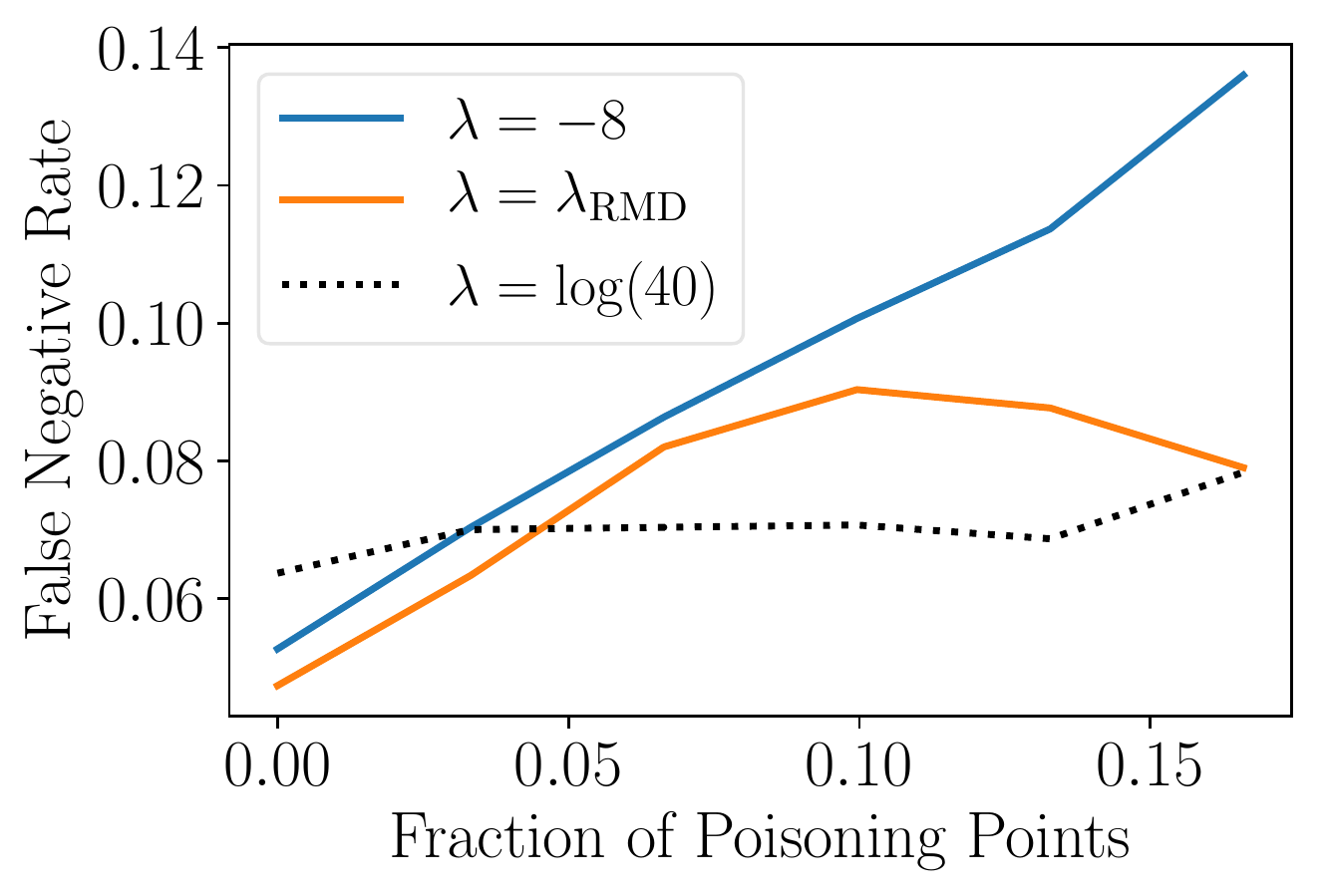}
		\caption{}
	\end{subfigure}

	\caption{Average test false positive and false negative rate for the DNN on ImageNet.  $\lambda_{\text{RMD}}$ represents the one learned with RMD: (a) Test false positive rate and (b) test false negative rate.}
	\label{fig:dnnfpr}
\end{figure*}

In Fig. \ref{fig:lrfpr} we represent the results for the test false positive and false negative rate for LR on MNIST, FMNIST, and
ImageNet. The results are consistent with the test error shown in Fig.~\ref{fig:lropt}(upper row): The rates for $\lambda_\text{RMD}$ lower bound the ones where regularisation is not applied ($\lambda=-8$), whose trend is very similar to the case of $\lambda_\text{CLEAN}$. For MNIST, the false positive rate for $\lambda_\text{RMD}$ (Fig. \ref{fig:lrfpr}(a)) is lower than the one for the large value of $\lambda$ until a $15\%$ of poisoning points; the false negative rate for $\lambda_\text{RMD}$ (Fig. \ref{fig:lrfpr}(d)) is lower than the one for the large regularisation term until a $2\%$ of poisoning points, and follows a similar trend from that point on. On the other side, for FMNIST, the false positive rate when learning $\lambda$ (Fig. \ref{fig:lrfpr}(b)) is lower than the one for $\lambda=\log(400)$ when the fraction of poisoning points is lower than $5\%$, and it reaches a maximum at around $10\%$, where it starts to decrease, being again lower at approximately $15\%$ of poisoning points; the false negative rate for $\lambda_\text{RMD}$ (Fig. \ref{fig:lrfpr}(e)) is lower until a $2.5\%$ of poisoning, and from then on it follows a similar pattern to the one for $\lambda=\log(400)$. Finally, for ImageNet, the false positive rate when $\lambda$ is learned (Fig. \ref{fig:lrfpr}(c)) follows a very similar trend to the one for $\lambda=\log(3,000)$; in the case of the false negative rate (Fig. \ref{fig:lrfpr}(f)), it is lower than the one for the large regularisation term until a $3\%$ of poisoning points, and from that point on it presents a similar pattern.

\subsubsection{Deep Neural Networks}

In Fig. \ref{fig:dnnfpr} we depict the results for the test false positive and false negative rate for the DNN on ImageNet. The results present a similar trend to the test error (Fig.~\ref{fig:dnnopt}(a)), as again the rates corresponding to $\lambda=-8$ are lower bounded by the ones for $\lambda_\text{RMD}$. The false positive rate for $\lambda_\text{RMD}$ (Fig. \ref{fig:dnnfpr}(a)) appears to be greater than the plot corresponding to $\lambda=\log(40)$, which does not seem affected by the attack. Regarding the false negative rate for $\lambda_\text{RMD}$ (Fig. \ref{fig:dnnfpr}(b)), it is lower than the one for $\lambda=\log(40)$ until a $5\%$ of poisoning points, and after a roughly $10\%$ of poisoning points it slightly decreases when the fraction of poisoning points increases, until a $16.6\%$ of poisoning points, where it becomes equal to the one for the strong regularisation term.

\subsection{Histograms of the Models' Parameters}

\subsubsection{Logistic Regression}

\begin{figure*}[!t]
	\begin{subfigure}[b]{0.316\textwidth}
		\includegraphics[width=\textwidth]{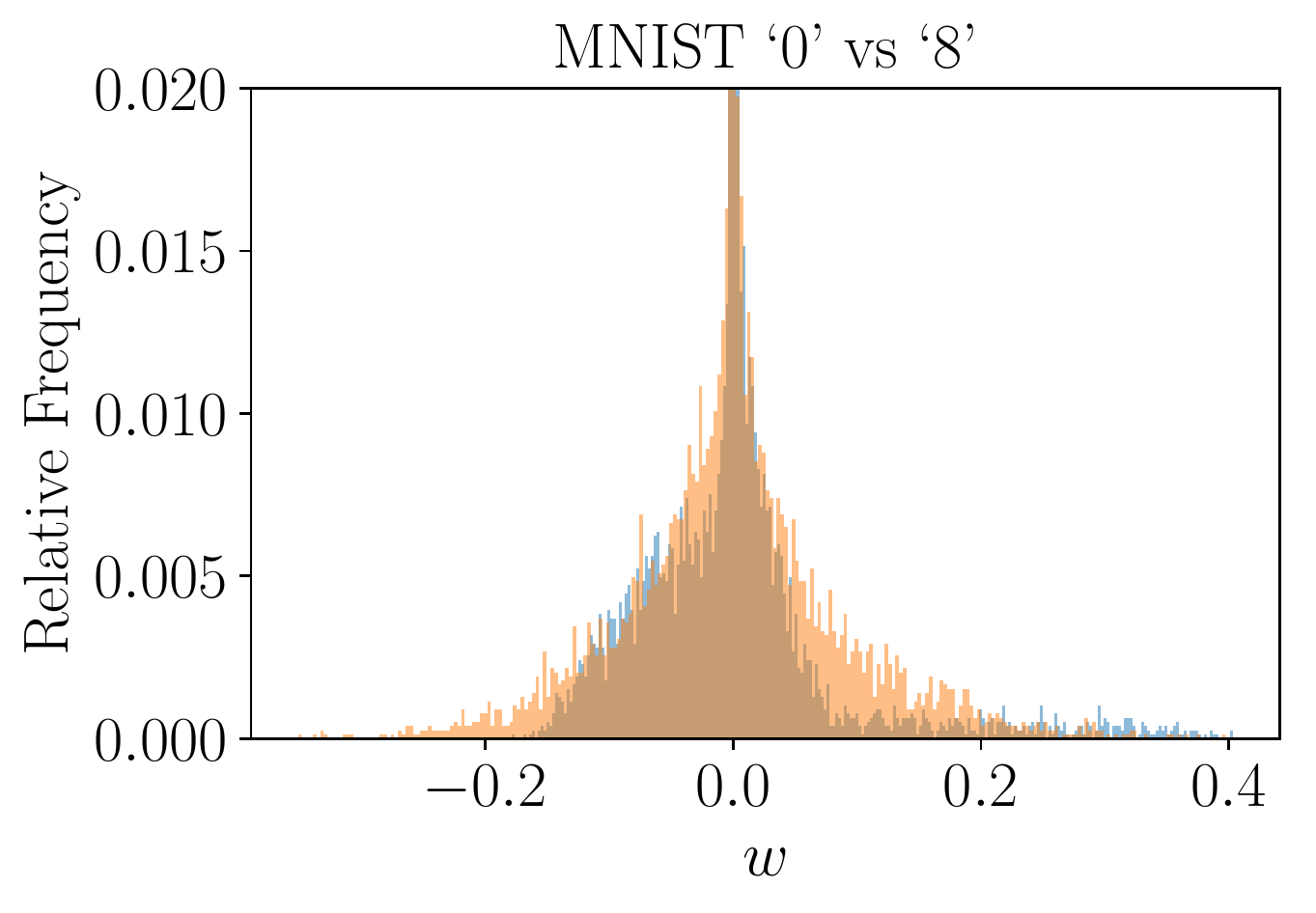}
		\caption{}
	\end{subfigure}
	\enskip % Control spacing between left and right figure, can use \enskip, \quad, \qquad, \hfill
	\begin{subfigure}[b]{0.295\textwidth}
		\includegraphics[width=\textwidth]{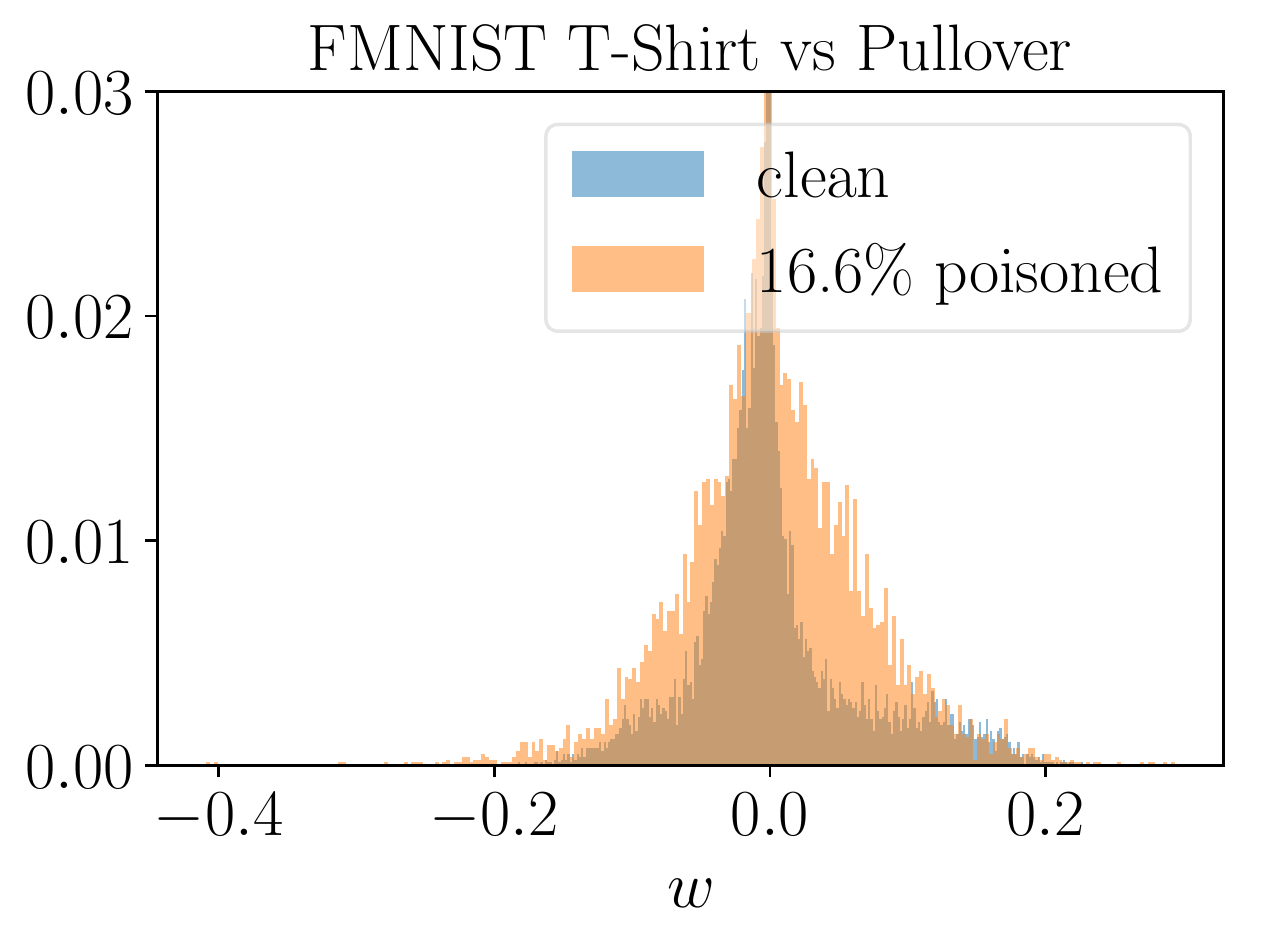}
		\caption{}
	\end{subfigure}
	\enskip % Control spacing between left and right figure, can use \enskip, \quad, \qquad, \hfill
	\begin{subfigure}[b]{0.319\textwidth}
		\includegraphics[width=\textwidth]{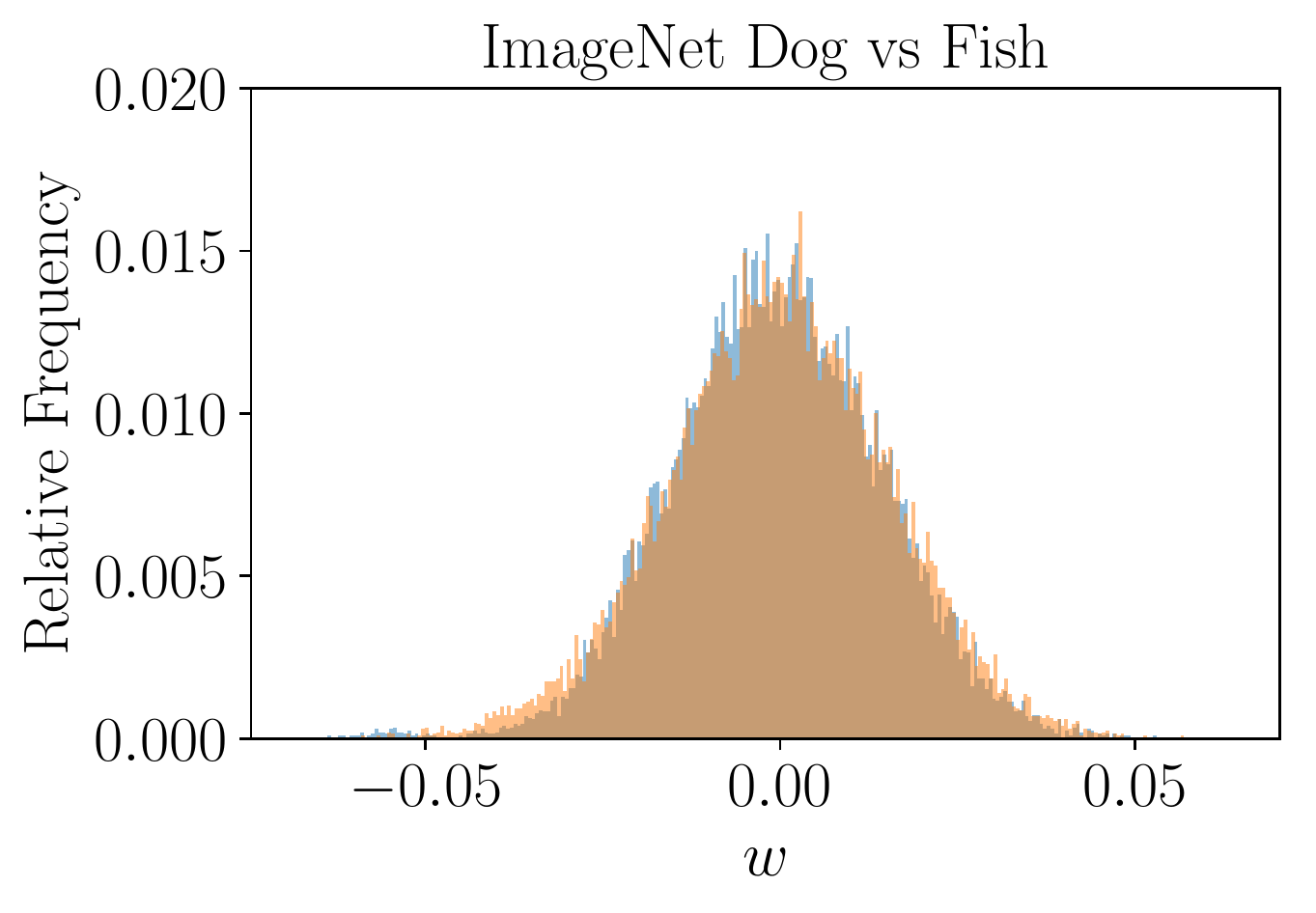}
		\caption{}
	\end{subfigure}
	\enskip % Control spacing between left and right figure, can use \enskip, \quad, \qquad, \hfill
	\begin{subfigure}[b]{0.323\textwidth}
		\includegraphics[width=\textwidth]{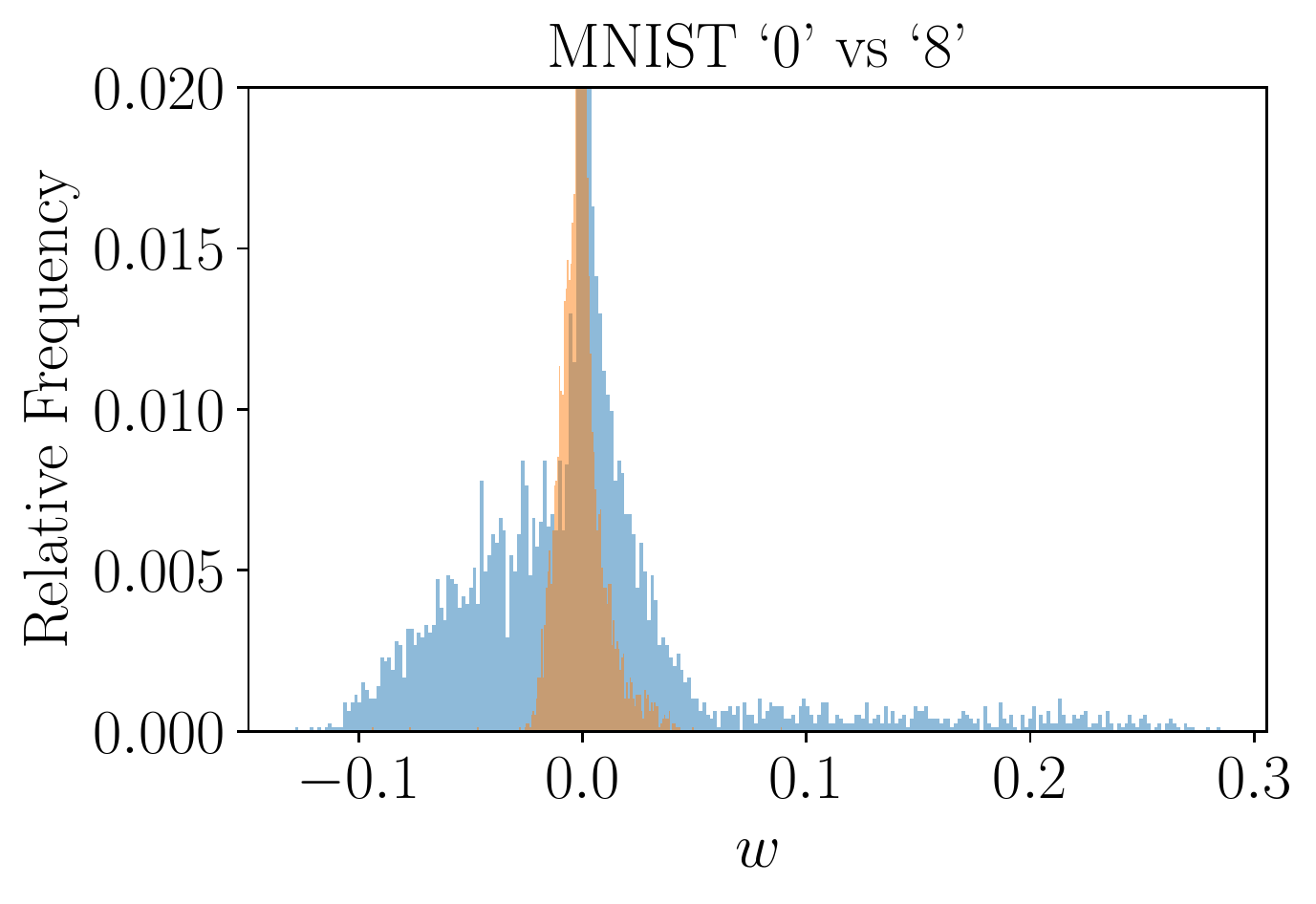}
		\caption{}
	\end{subfigure}
	\enskip % Control spacing between left and right figure, can use \enskip, \quad, \qquad, \hfill
	\begin{subfigure}[b]{0.3\textwidth}
		\includegraphics[width=\textwidth]{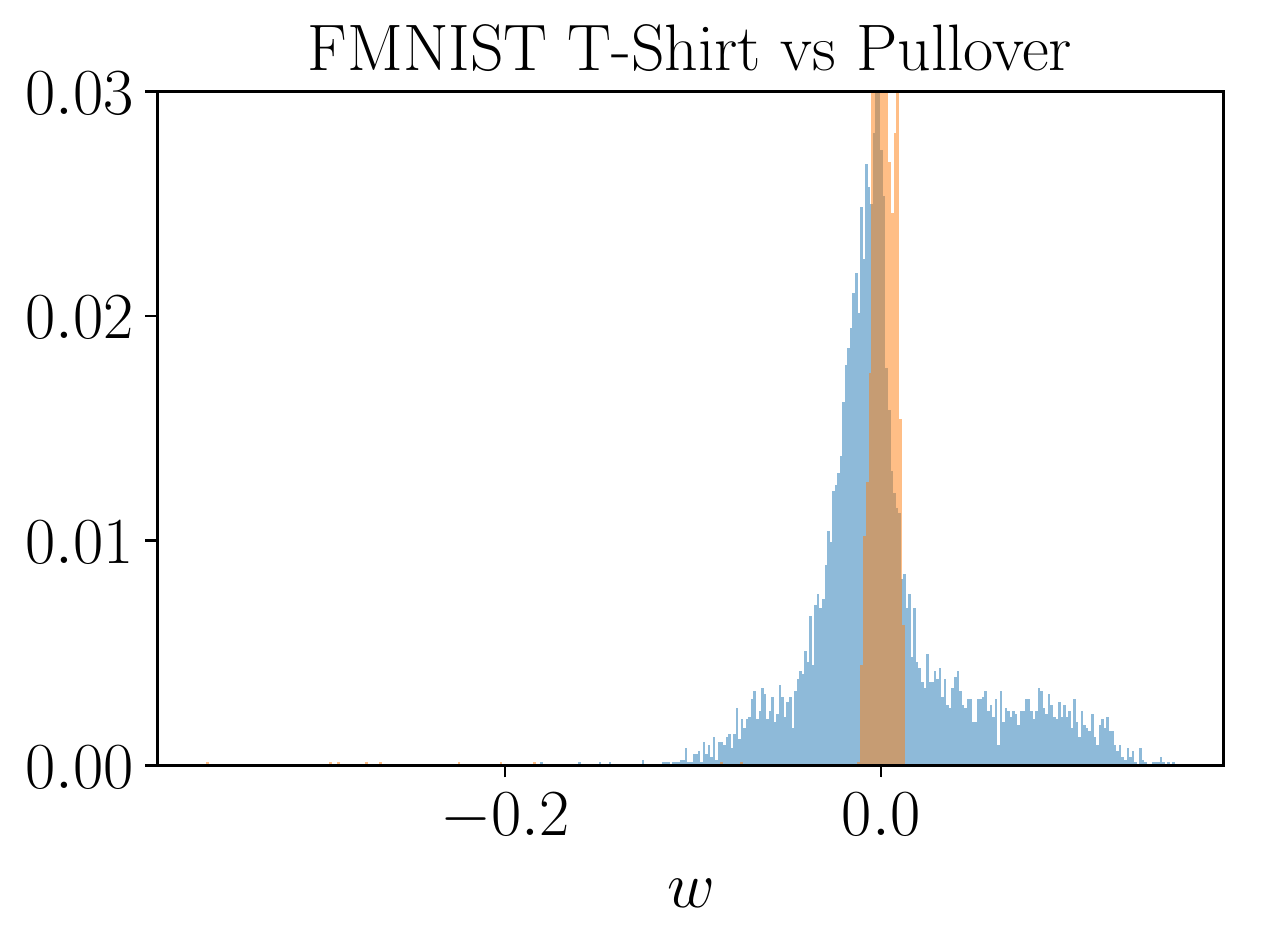}
		\caption{}
	\end{subfigure}
	\enskip % Control spacing between left and right figure, can use \enskip, \quad, \qquad, \hfill
	\begin{subfigure}[b]{0.323\textwidth}
		\includegraphics[width=\textwidth]{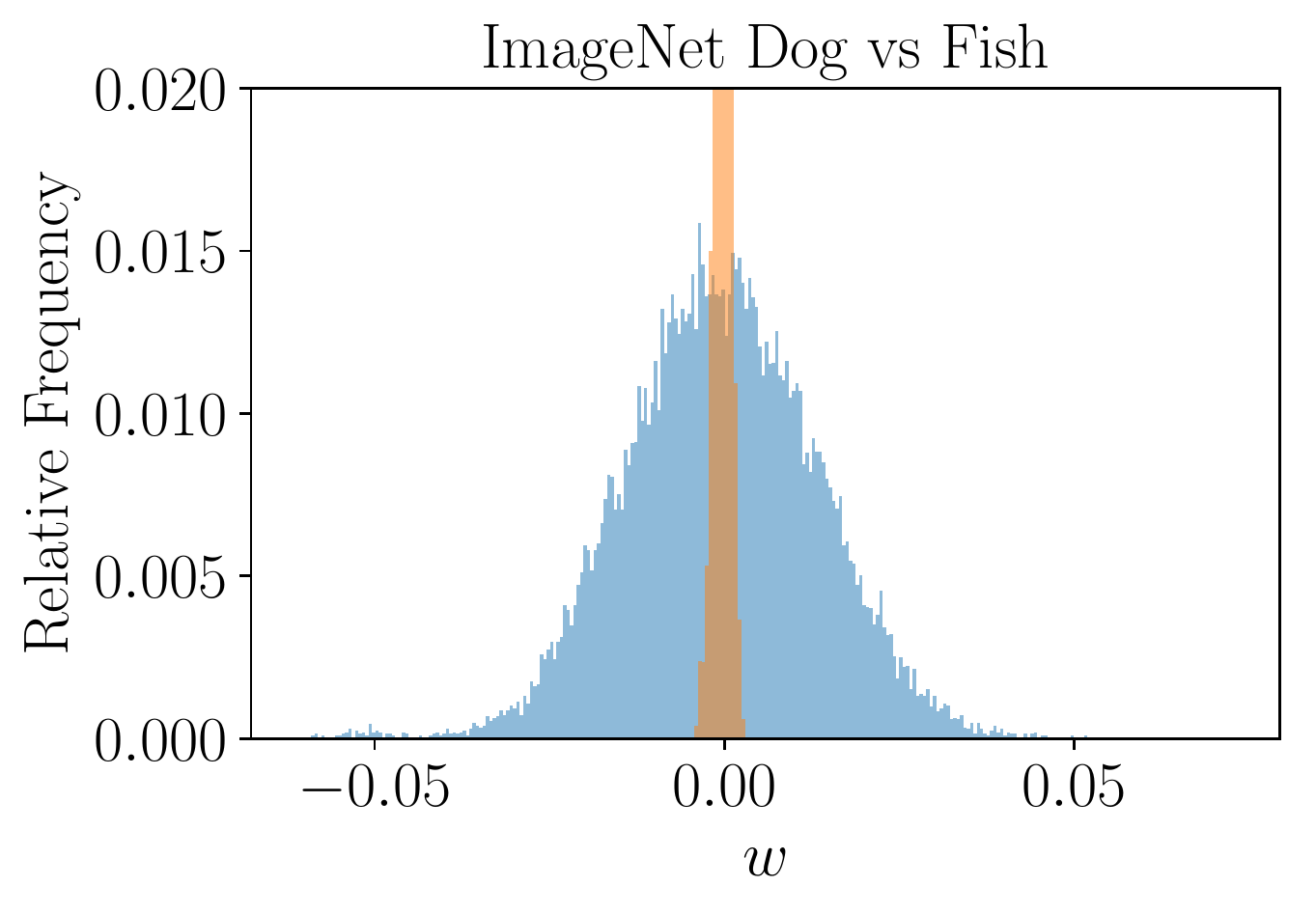}
		\caption{}
	\end{subfigure}

	\caption{Histograms (in terms of relative frequency) of the LR's parameters. For a better visualisation, the upper part of some subfigures has been omitted. The first row represents the case where no regularisation is applied: (a) MNIST, (b) FMNIST, and (c) ImageNet. The second row shows the case where $\lambda$ is learned with RMD: (d) MNIST, (e) FMNIST, and (f) ImageNet.}
	\label{fig:lrhist}
\end{figure*}

\begin{figure*}[!t]
	\begin{subfigure}[b]{0.323\textwidth}
		\includegraphics[width=\textwidth]{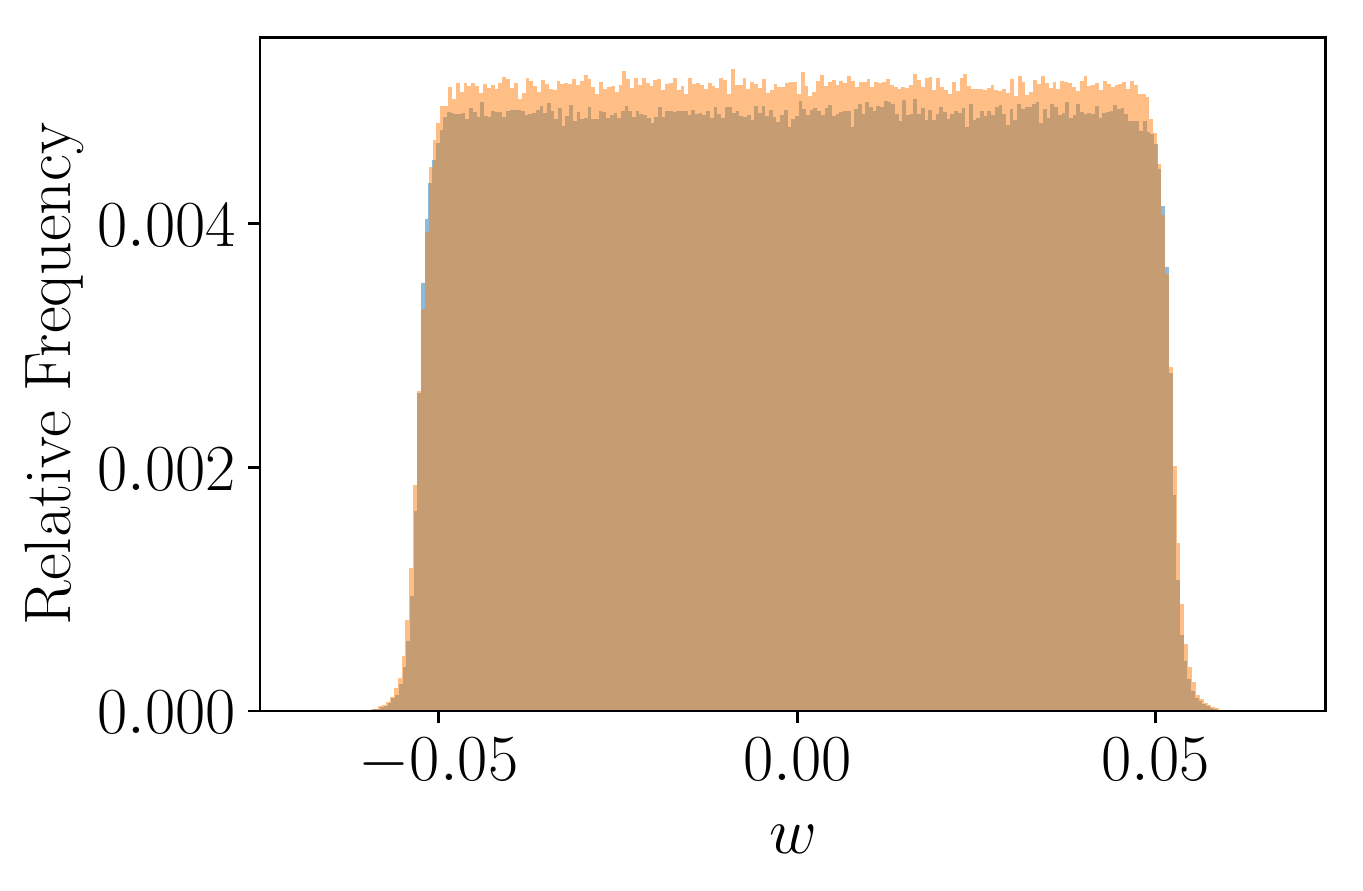}
		\caption{}
	\end{subfigure}
	\enskip % Control spacing between left and right figure, can use \enskip, \quad, \qquad, \hfill
	\begin{subfigure}[b]{0.308\textwidth}
		\includegraphics[width=\textwidth]{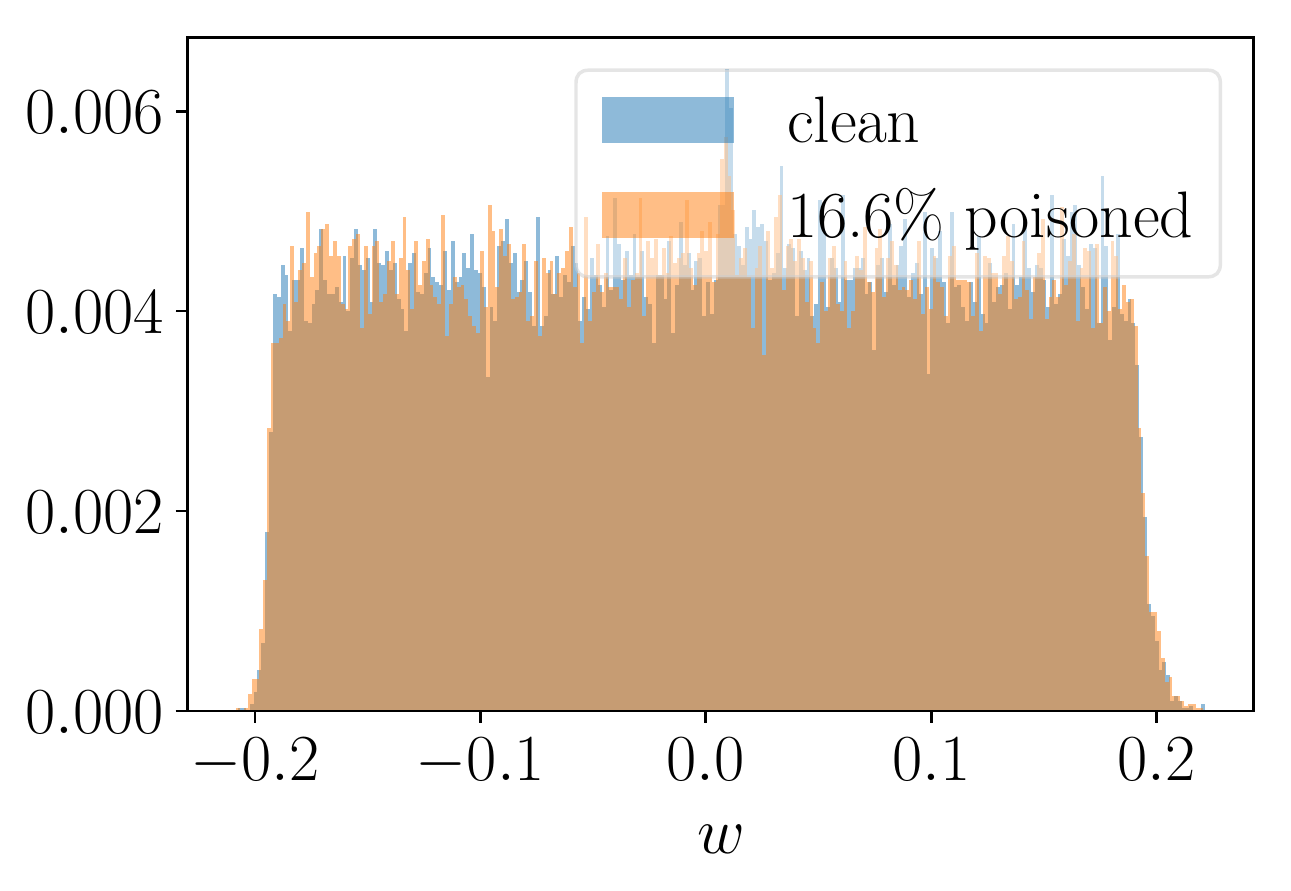}
		\caption{}
	\end{subfigure}
	\enskip % Control spacing between left and right figure, can use \enskip, \quad, \qquad, \hfill
	\begin{subfigure}[b]{0.31\textwidth}
		\includegraphics[width=\textwidth]{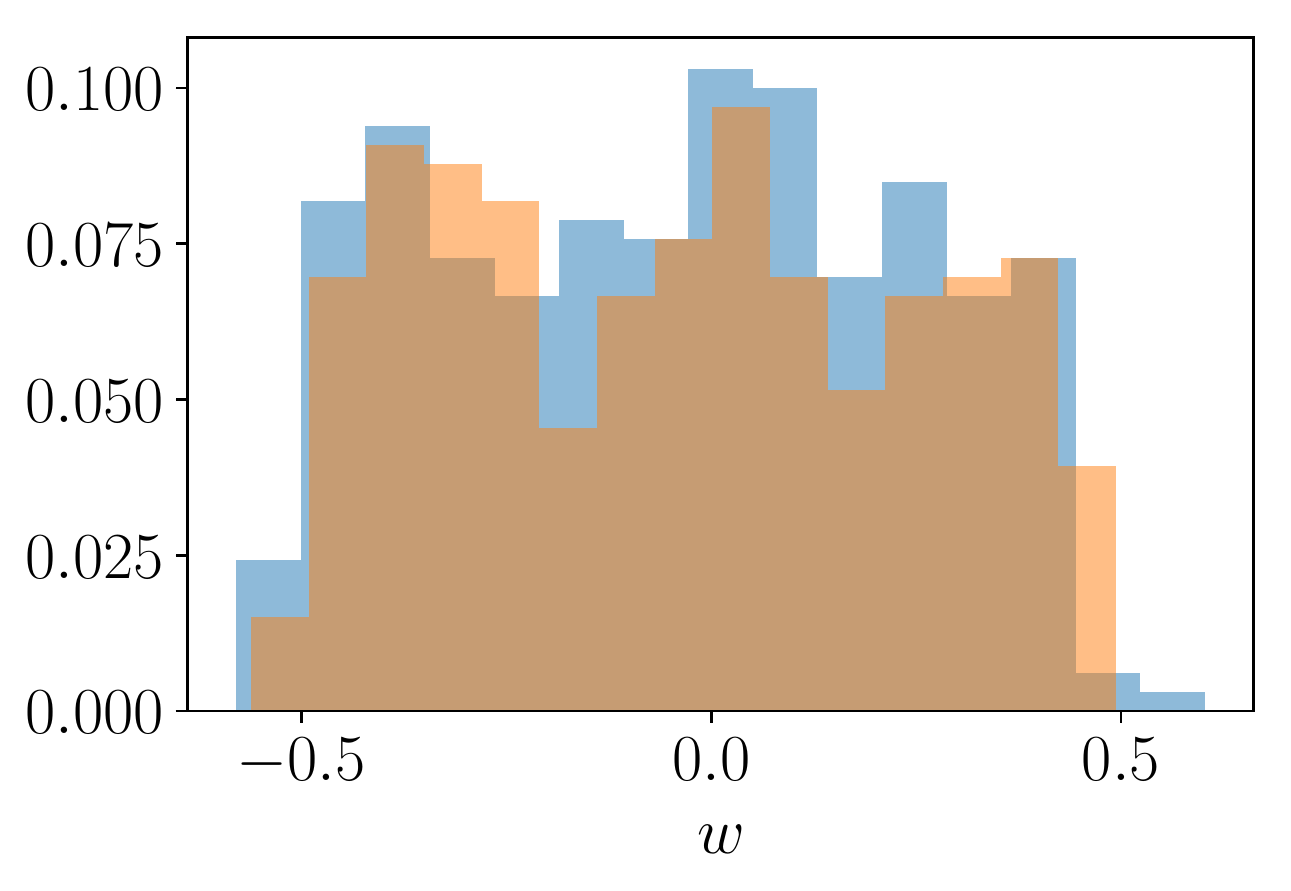}
		\caption{}
	\end{subfigure}
	\enskip % Control spacing between left and right figure, can use \enskip, \quad, \qquad, \hfill
	\begin{subfigure}[b]{0.329\textwidth}
		\includegraphics[width=\textwidth]{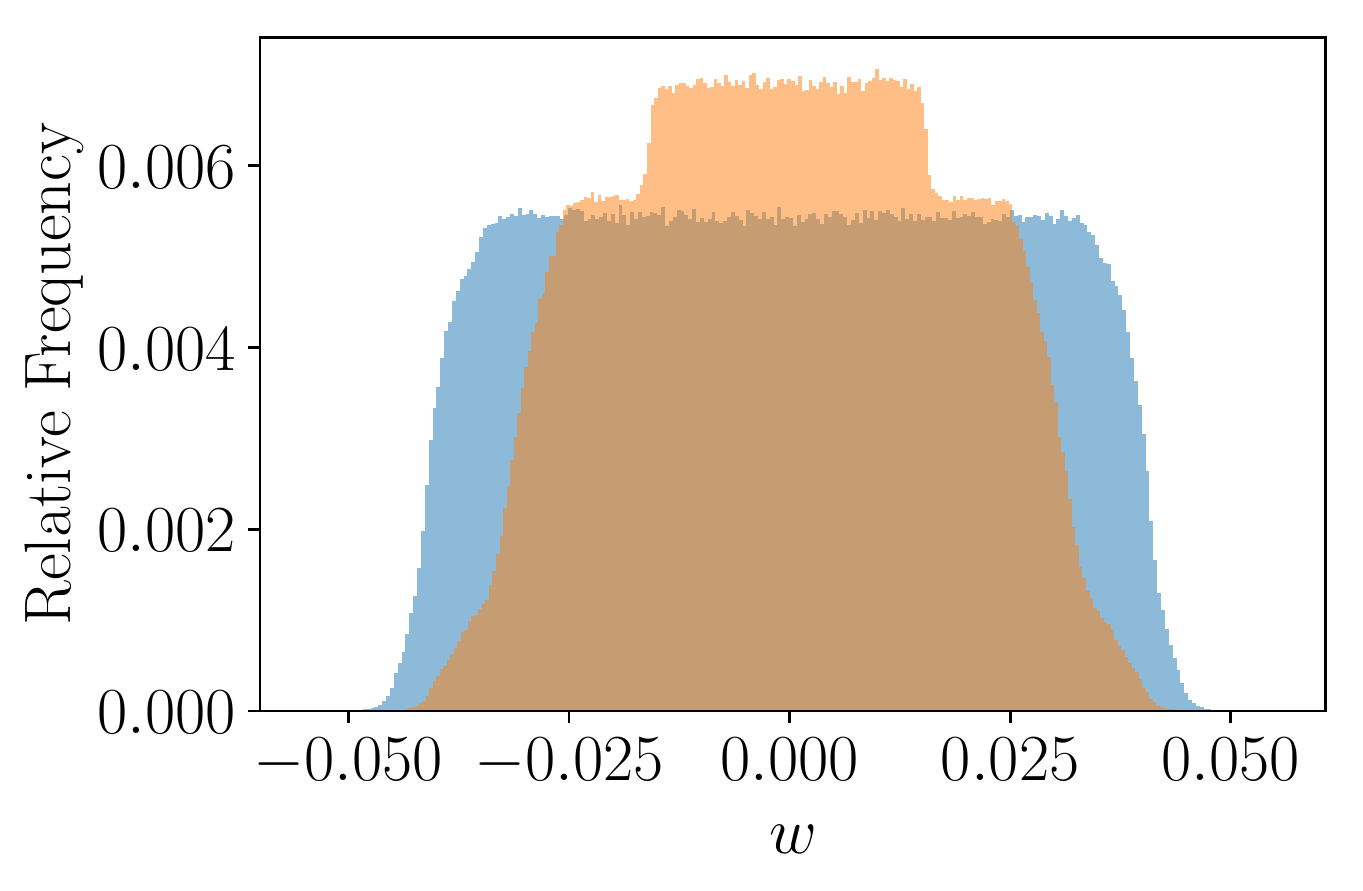}
		\caption{}
	\end{subfigure}
	\enskip % Control spacing between left and right figure, can use \enskip, \quad, \qquad, \hfill
	\begin{subfigure}[b]{0.317\textwidth}
		\includegraphics[width=\textwidth]{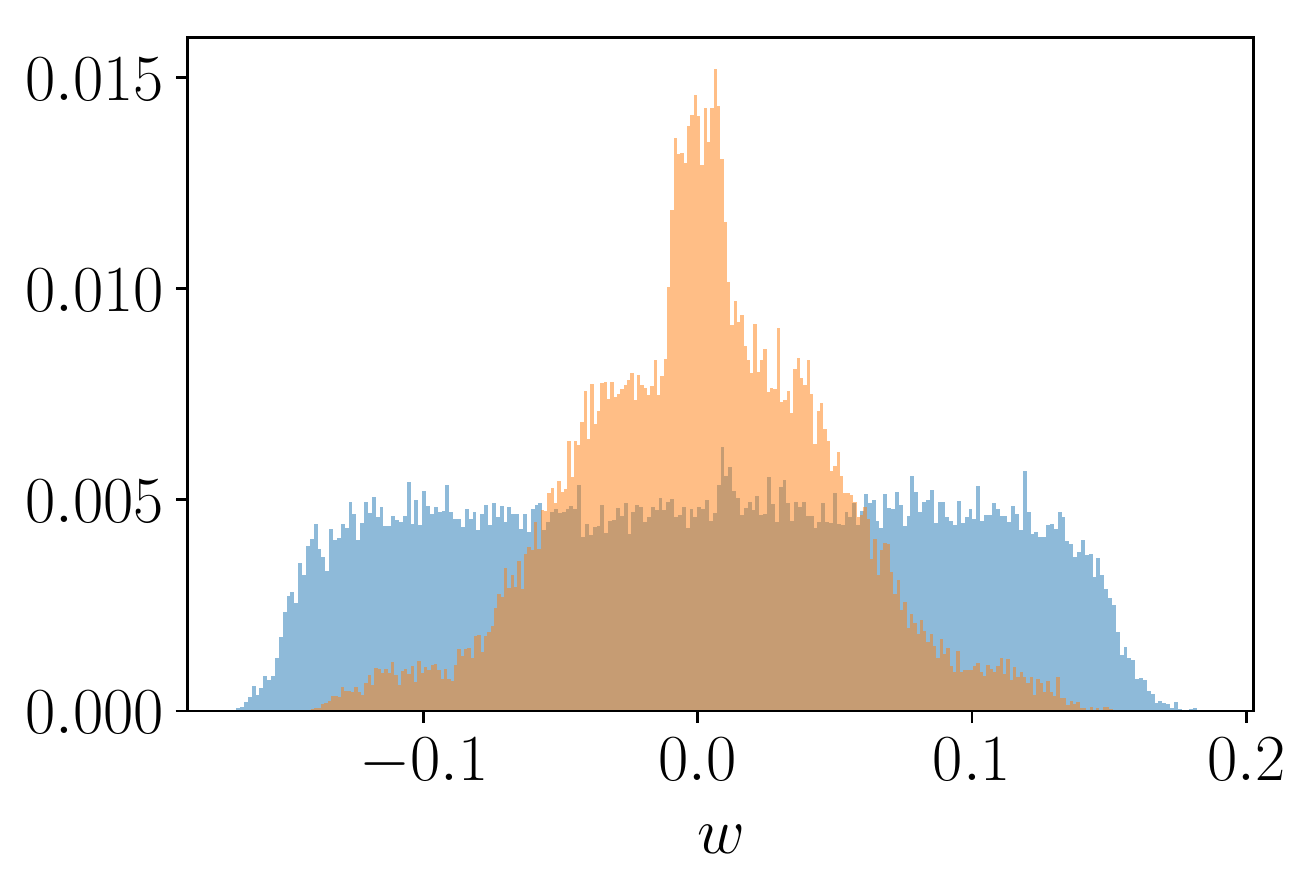}
		\caption{}
	\end{subfigure}
	\enskip % Control spacing between left and right figure, can use \enskip, \quad, \qquad, \hfill
	\begin{subfigure}[b]{0.305\textwidth}
		\includegraphics[width=\textwidth]{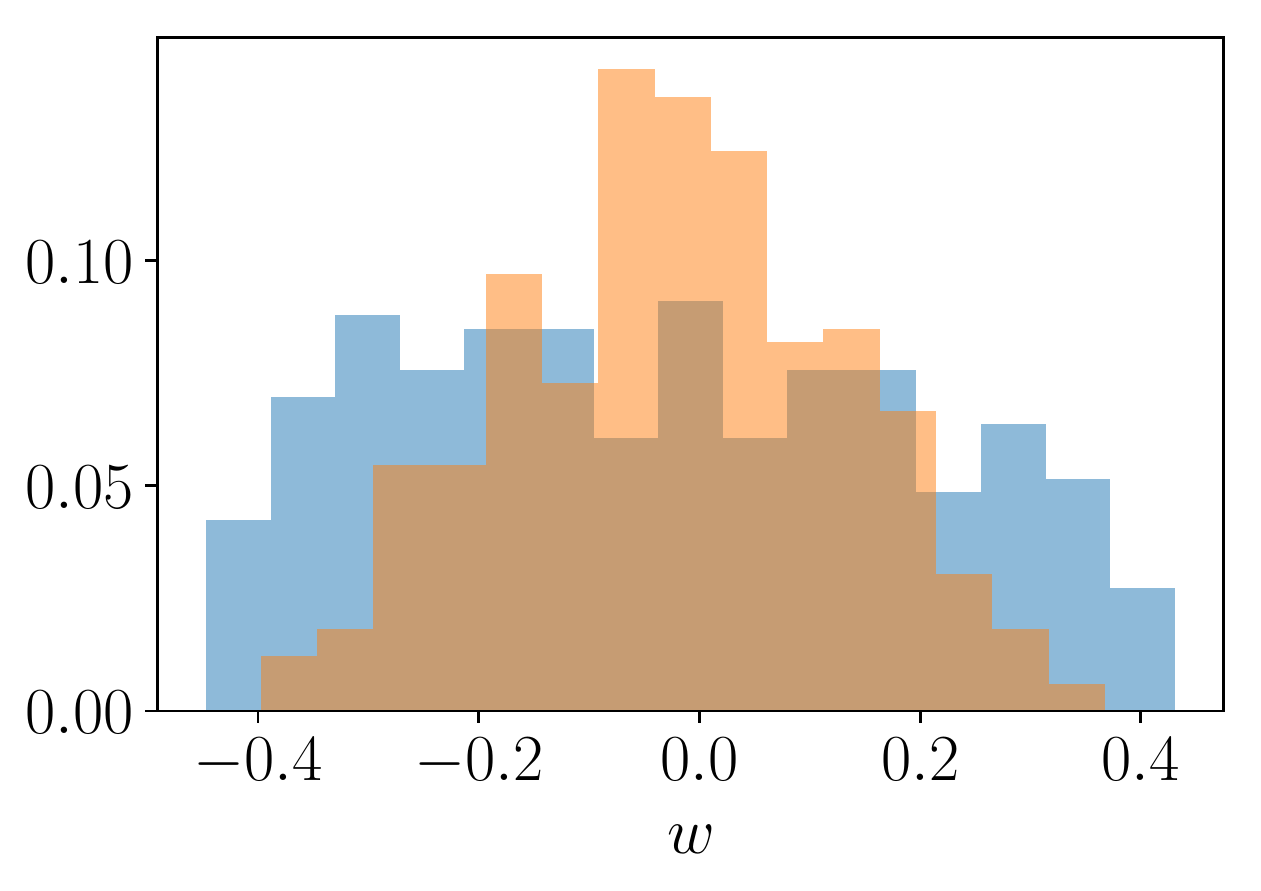}
		\caption{}
	\end{subfigure}

	\caption{Histograms (in terms of relative frequency) of the DNN's parameters on ImageNet.  The first row represents the case where regularisation is not applied: (a) first hidden layer, (b) second hidden layer, and (c) output layer. The second row shows the corresponding results when $\lambda$ is learned with RMD: (a) first hidden layer, (b) second hidden layer, and (c) output layer.}
	\label{fig:dnnhist}
\end{figure*}

In Fig. \ref{fig:lrhist} we represent the histogram of the values of the parameters, $w$, of the LR classifier on MNIST, FMNIST and ImageNet, when the training sets are clean (blue bars) and when a $16.6\%$ of each training set is replaced by poisoning points  (orange bars). To appreciate better the distribution of the parameters, we omit the upper part of some plots. The first row depicts the cases when no regularisation is applied, and the second row shows the case where $\lambda$ is learned using RMD. We can clearly appreciate the effect of the regularisation:
For all the datasets, the range of values of the parameters (blue bars of the second row) is narrowed down, and when the attacker injects poisoning points, this forces the model to compress more these values (orange bars of the second row) close to $0$, as the value of $\lambda$ increases. This leads to a more stable model under possible malicious manipulations of the training data.

\subsubsection{Deep Neural Networks}

In Fig. \ref{fig:dnnhist} we represent the histogram of the values of the parameters of the DNN classifier on ImageNet,  for the clean dataset (blue bars) and when a $16.6\%$ of the training set is replaced by poisoning points (orange bars). The first, second, and third column show the results for the first hidden, second hidden, and output layer, correspondingly. The first row depicts the cases where no regularisation is applied, and the second row when $\lambda$ is learned using RMD. Once again, we can appreciate the effect of the regularisation: For all the layers, the range of values of the parameters (blue bars of the second row) is narrowed down, and when the attacker poisons the training data, this forces the model to bound more these values (orange bars of the second row), as the corresponding values of $\lambda$ for each layer increase.

\subsection{Examples of Poisoning Points Computed}

\begin{figure*}[!t]
	\begin{subfigure}[b]{0.323\textwidth}
		\includegraphics[width=\textwidth]{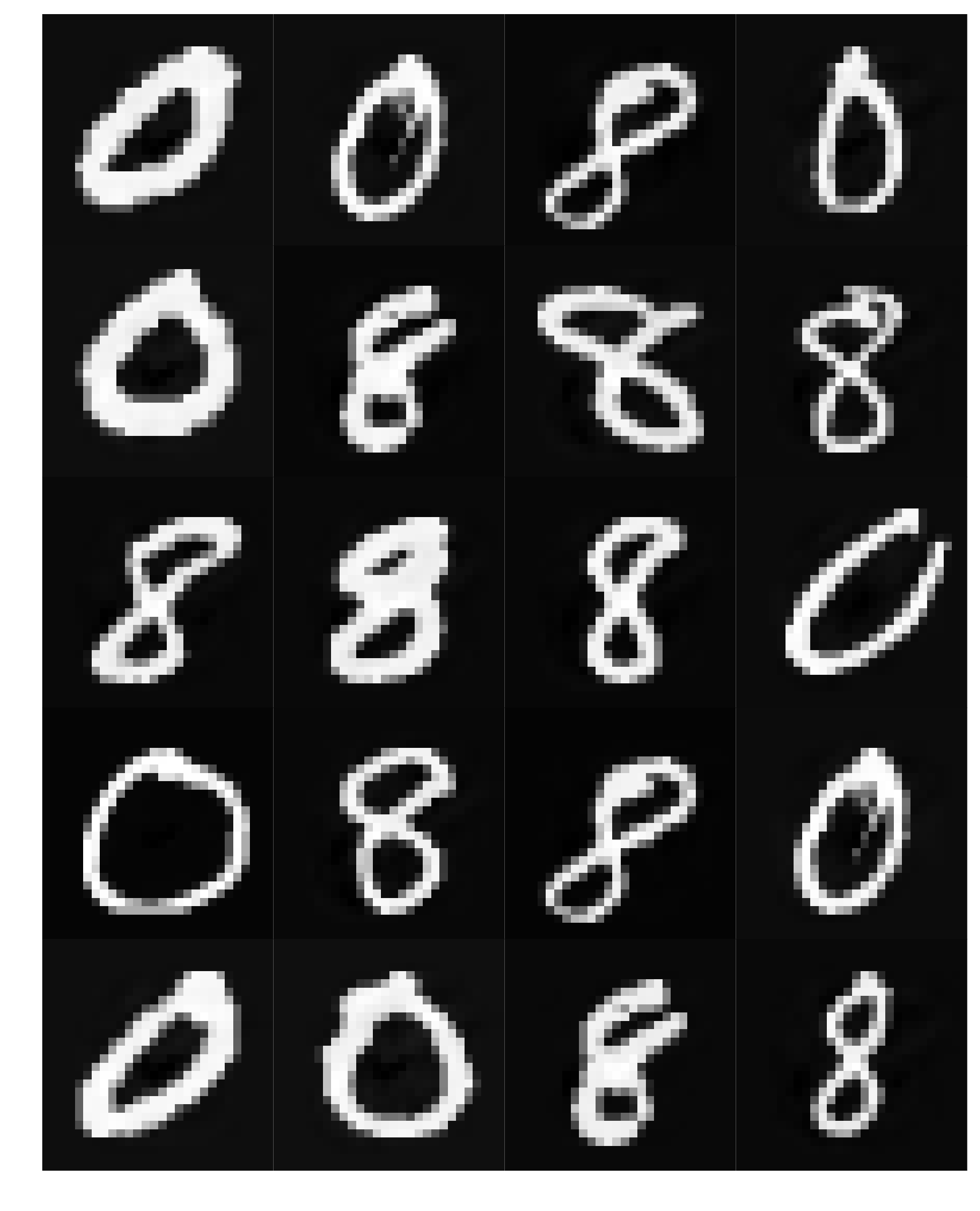}
		\caption{}
		
	\end{subfigure}
	\enskip % Control spacing between left and right figure, can use \enskip, \quad, \qquad, \hfill
	\begin{subfigure}[b]{0.323\textwidth}
		
		\includegraphics[width=\textwidth]{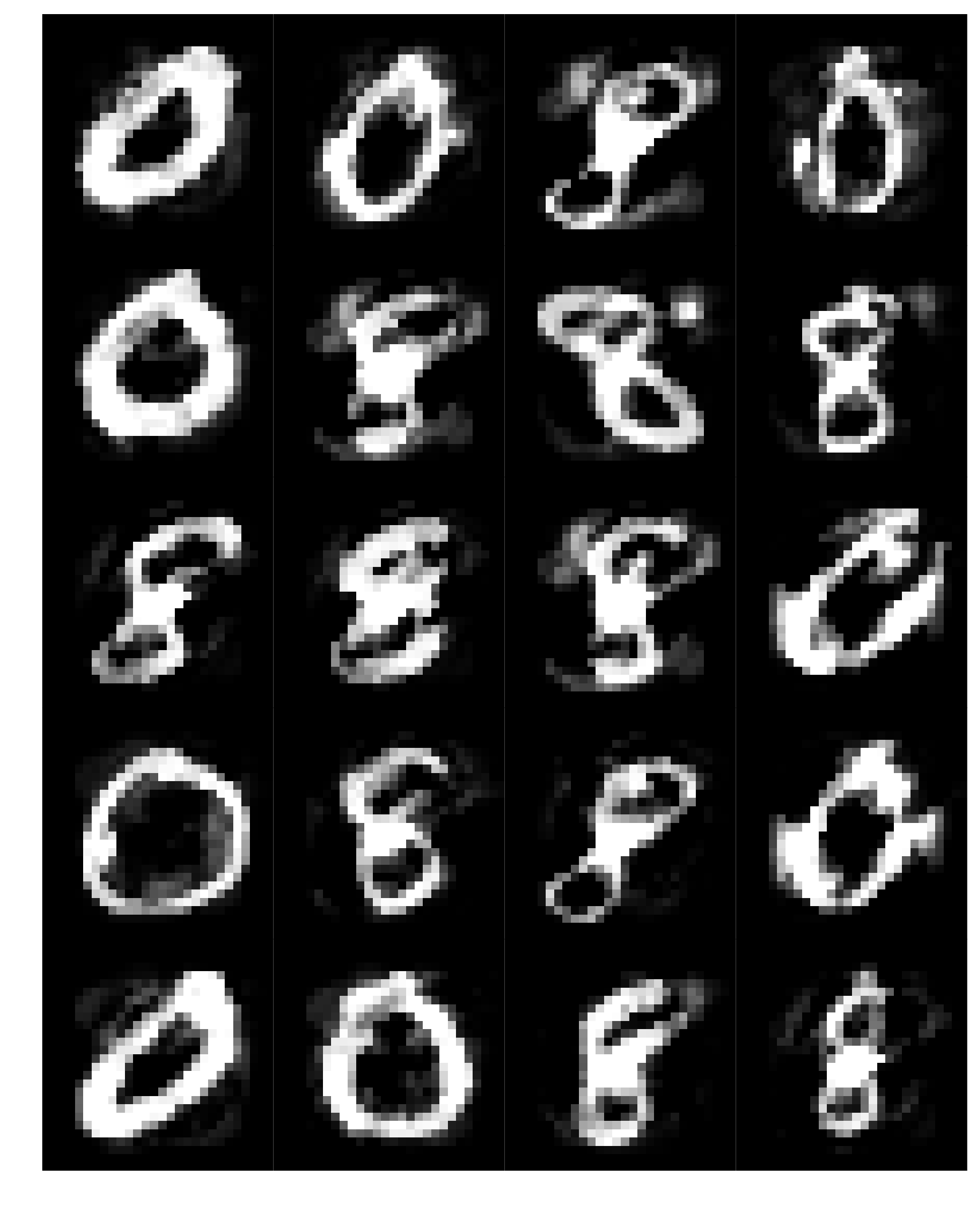}
		\caption{}
	\end{subfigure}
	\begin{subfigure}[b]{0.323\textwidth}
		
		\includegraphics[width=\textwidth]{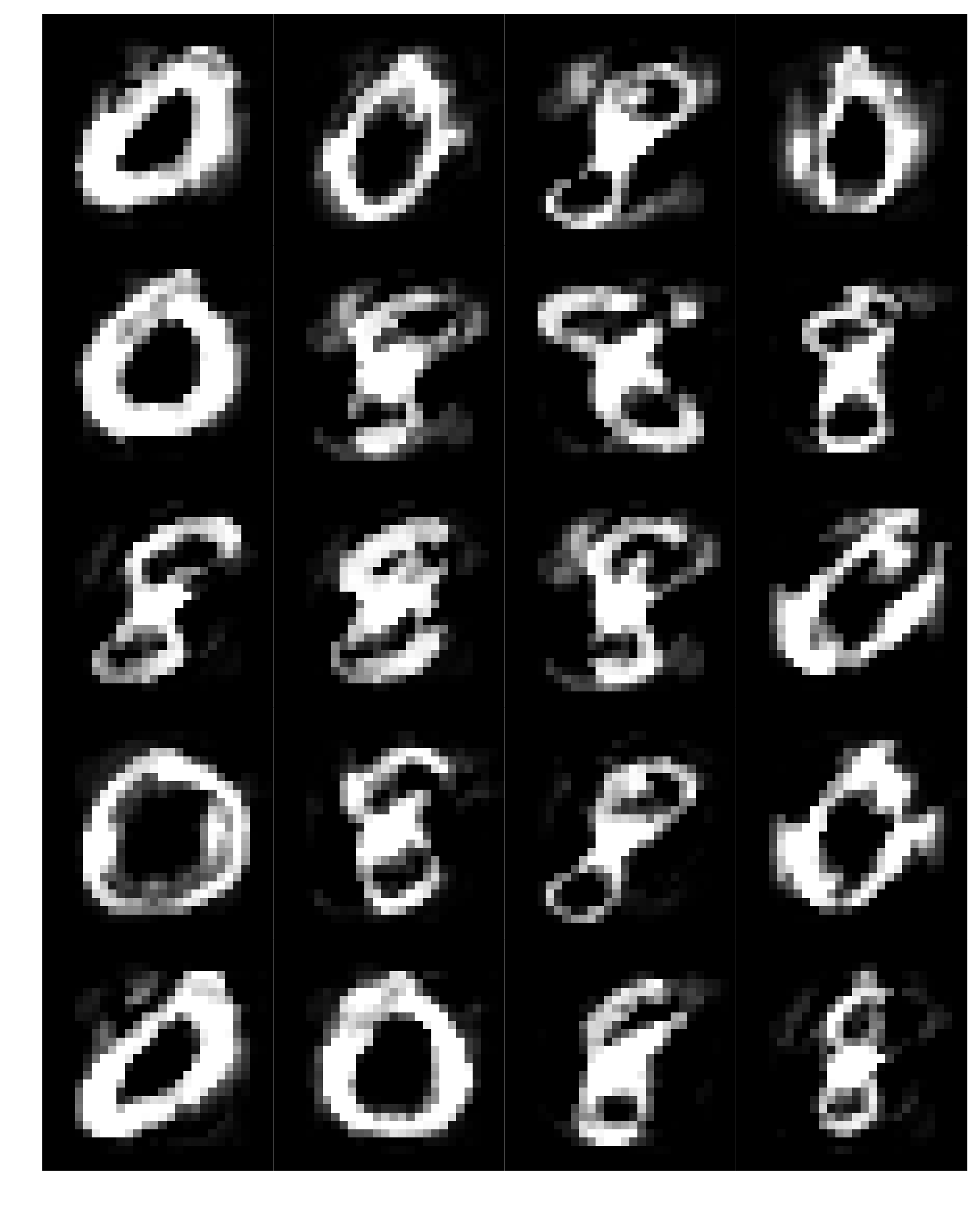}
		\caption{}
	\end{subfigure}

	\caption{Examples of poisoning points computed with RMD for the LR classifier on MNIST: (a)~initial features, (b) features optimised for $\lambda=-8$, and (c) features optimised for $\lambda_\text{RMD}$.}
	\label{fig:plotmnist}
\end{figure*}

\begin{figure*}[!t]
	\begin{subfigure}[b]{0.323\textwidth}
		\includegraphics[width=\textwidth]{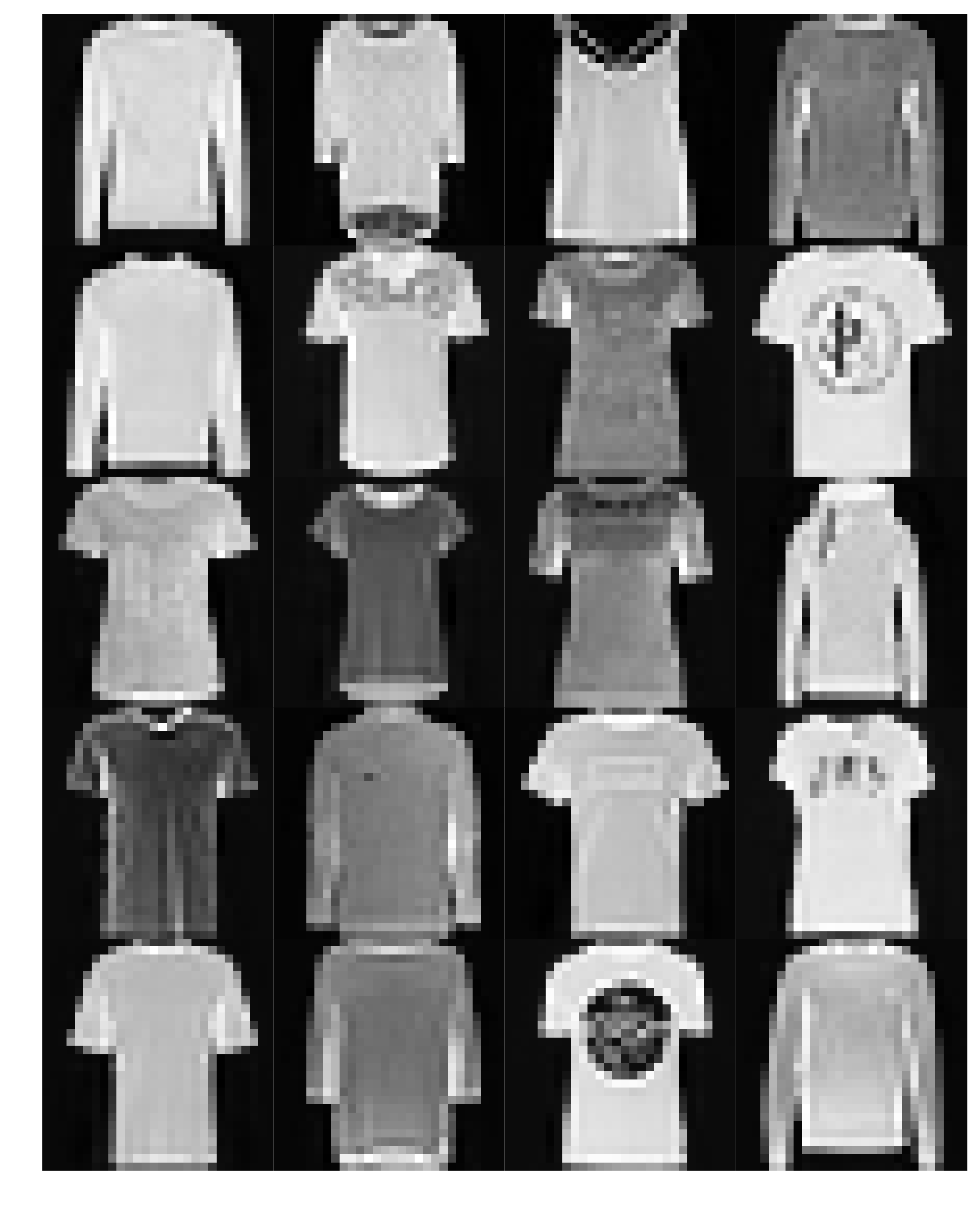}
		\caption{}
		
	\end{subfigure}
	\enskip % Control spacing between left and right figure, can use \enskip, \quad, \qquad, \hfill
	\begin{subfigure}[b]{0.323\textwidth}
		
		\includegraphics[width=\textwidth]{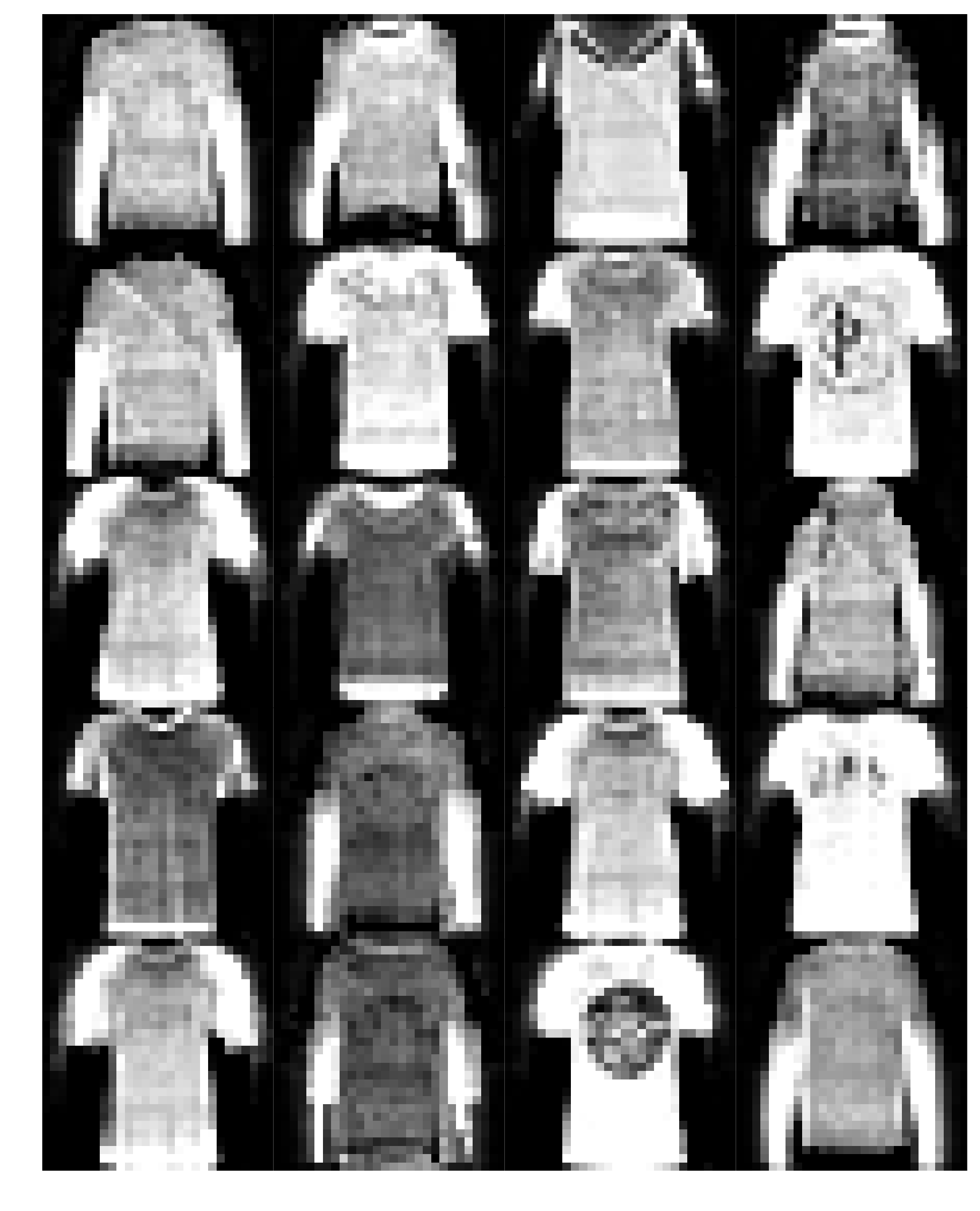}
		\caption{}
	\end{subfigure}
	\begin{subfigure}[b]{0.323\textwidth}
		
		\includegraphics[width=\textwidth]{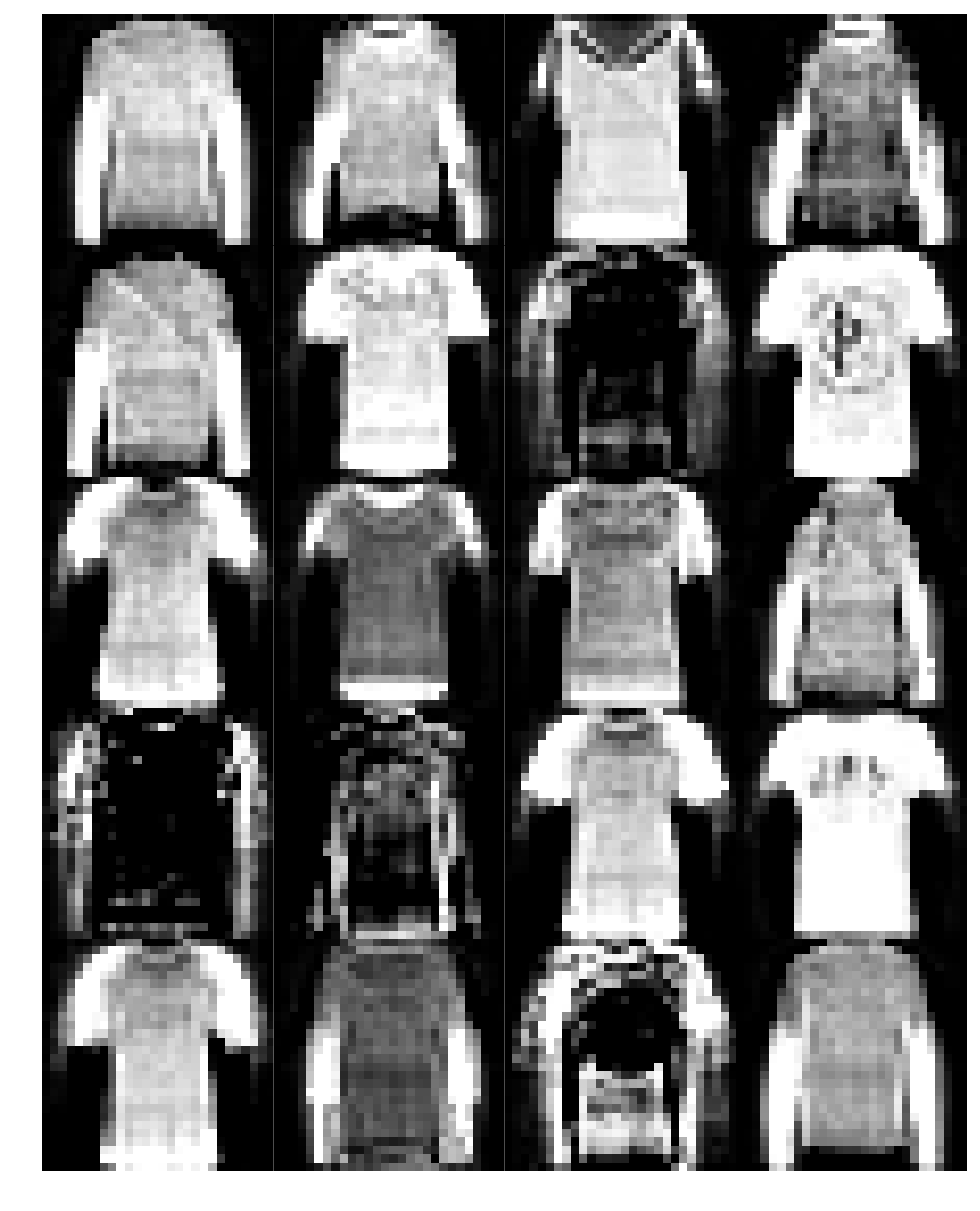}
		\caption{}
	\end{subfigure}

	\caption{Examples of poisoning points computed with RMD for the LR classifier on FMNIST: (a)~initial features, (b) features optimised for $\lambda=-8$, and (c) features optimised for $\lambda_\text{RMD}$.}
	\label{fig:plotfmnist}
\end{figure*}

Here we depict the image representation of poisoning points randomly sampled across different fractions of poisoning. These points are computed with RMD for the LR classifier on MNIST (Fig. \ref{fig:plotmnist}) and FMNIST (Fig. \ref{fig:plotfmnist}), respectively. For both datasets we represent the initial values of the features of the poisoning points and their optimised values for the cases where there is no regularisation ($\lambda=-8$), and when $\lambda$ is learned ($\lambda_\text{RMD}$). We do not include examples for ImageNet as we use the same Inception-v3 features as in \cite{koh2017understanding}, so
that each of the images is represented by a $2,048$-dimensional vector that does not have an obvious representation in the image domain.

For MNIST, we can observe that the optimised points present a similar pattern for the cases where $\lambda=-8$ and when $\lambda$ is learned. For FMNIST, they also present a similar representation in most of the cases, although when the attack is strong---i.e. the ratio of poisoning points is large---the solutions obtained when $\lambda$ is learned present more saturated features, which could be more detectable for the human perception. In other words, in this case the poisoning attack against the classifier learning the regularisation term requires increasing the level of perturbation in the malicious examples, which can make the attack more detectable.

\end{document}